\documentclass[]{fairmeta}
\usepackage{times}

\usepackage{amsmath,amsfonts,bm}

\def\eqref#1{equation~\ref{#1}}

\def\1{\bm{1}}

\DeclareMathAlphabet{\mathsfit}{\encodingdefault}{\sfdefault}{m}{sl}
\SetMathAlphabet{\mathsfit}{bold}{\encodingdefault}{\sfdefault}{bx}{n}

\usepackage{color-edits}
\addauthor{ye}{cyan}
\usepackage{ulem}

\usepackage{hyperref}
\usepackage{url}
\usepackage{algorithm}
\usepackage{algpseudocode}
\usepackage{algpseudocode}
\usepackage{booktabs,multirow,threeparttable}
\usepackage{graphicx}
\usepackage{subcaption}
\usepackage{pgfplots}
\usepackage{booktabs}
\usepackage{siunitx}
\usepackage{xcolor}
\usepackage{wrapfig}
\usepackage{svg}
\usepackage[dvipsnames]{xcolor}
\usepackage{framed}
\usepackage{adjustbox}
\usepackage[most]{tcolorbox}
\tcbuselibrary{listingsutf8}
\usepackage{listings}
\usepackage{siunitx}
\usepackage{makecell}
\newcommand{\DeltaUp}[1]{\ensuremath{\color{ForestGreen}{\uparrow}\,#1}}
\newcommand{\DeltaUpGreen}[1]{\textcolor{ForestGreen}{\uparrow}#1}
\newcommand{\DeltaDownRed}[1]{\textcolor{Red}
{\downarrow}#1}

\usepackage{dsfont}
\usepackage{amsmath} 

\newcommand{\MethodName}{MACA}
\title{Self-Improvement of Language Models by Post-Training on Multi-Agent Debate}

\author[1,2,*]{Ankur Samanta}
\author[1]{Akshayaa Magesh}
\author[1,3,*]{Runzhe Wu}
\author[1]{Ayush Jain}
\author[1]{Youliang Yu}
\author[1]{Daniel Jiang}
\author[1]{Boris Vidolov}
\author[2]{Paul Sajda}
\author[1,\dagger]{Yonathan Efroni}
\author[4,\dagger]{Kaveh Hassani}

\affiliation[1]{Meta AI}
\affiliation[2]{Columbia University}
\affiliation[3]{Cornell Tech}
\affiliation[4]{Meta Superintelligence Labs}

\contribution[*]{Work done at Meta}
\contribution[\dagger]{Joint last author, alphabetical order}

\abstract{Self-improvement, where models improve beyond their current performance without external supervision, remains a challenge. The core difficulty is sourcing a training signal stronger than what the model itself can currently produce. Majority voting has been shown to provide such a signal by aggregating over multiple samples, helping mitigate some of the inconsistencies in LM reasoning. In this work, we show that multi-agent debate---where models collaborate and exchange reasoning over multiple rounds---provides an even richer signal than single-round majority voting. We introduce Multi-Agent Consensus Alignment (\MethodName), which uses reinforcement learning (RL) to post-train models to effectively utilize multi-agent debate. We find that preference learning over full reasoning traces, learning to differentiate between majority and minority reasoning, is more effective than binary consensus rewards or SFT-based approaches for leveraging these debate signals. This produces three key improvements: models are (1) better at utilizing the multi-agent debate setting (+26.87\% on MATH), (2) individually more accurate (+21.51\% on MathQA), and (3) more self-consistent (+27.6\% on GSM8K). We also see strong generalization to unseen benchmarks (+16.3\% on GPQA, +11.6\% on CommonsenseQA).}

\date{\today}
\correspondence{Ankur Samanta at \email{as7416@columbia.edu}}

\metadata[Code]{\url{https://github.com/facebookresearch/maca}}
\begin{document}

\maketitle

\section{Introduction}
Self-improvement in language models, where models train on self-generated signals without external supervision, is challenging. LMs suffer from inconsistent reasoning, often producing contradictory solutions to the same problem under exploratory sampling~\citep{Holtzman2020Curious,Wang2022}. While inference-time aggregation methods such as majority voting and multi-agent debate help improve the reliability of the final response~\citep{Wang2022,li2024self,irving2018debate}, they do not address the core problem: the internal reasoning instability that produces these inconsistencies in the first place. We argue that self-consistency, the ability to generate consistent outputs across diverse reasoning paths~\citep{Elazar2021Consistency,Wang2022}, is a property of well-aligned reasoning systems. In the brain, consistency emerges through dedicated control mechanisms in prefrontal and anterior cingulate regions that resolve conflicts between competing neural activations, directing cognition toward coherent solutions~\citep{miller2001integrative,friston2010free,botvinick2004conflict,shenhav2013expected,zhang2025neural}. Language models, by contrast, lack mechanisms to align their diverse reasoning paths toward consensus.

Prior work shows that majority-vote signals can be an effective self-improvement signal~\citep{Wang2022,Zuo2025TTRL,prasad2024self}. However, the models' underlying poor self-consistency can reduce the efficacy of such reinforcement. When models sample low-quality, inconsistent reasoning traces, aggregation can be counterproductive: noisy arguments compound rather than cancel out~\citep{Radharapu2025Arbiters}. We show that this problem is particularly acute in small language models, where simple statistical aggregation can compound low-quality reasoning and collaboration can result in weak consensus. 

\begin{figure*}[t]
    \centering
    \includegraphics[width=1.0\linewidth]{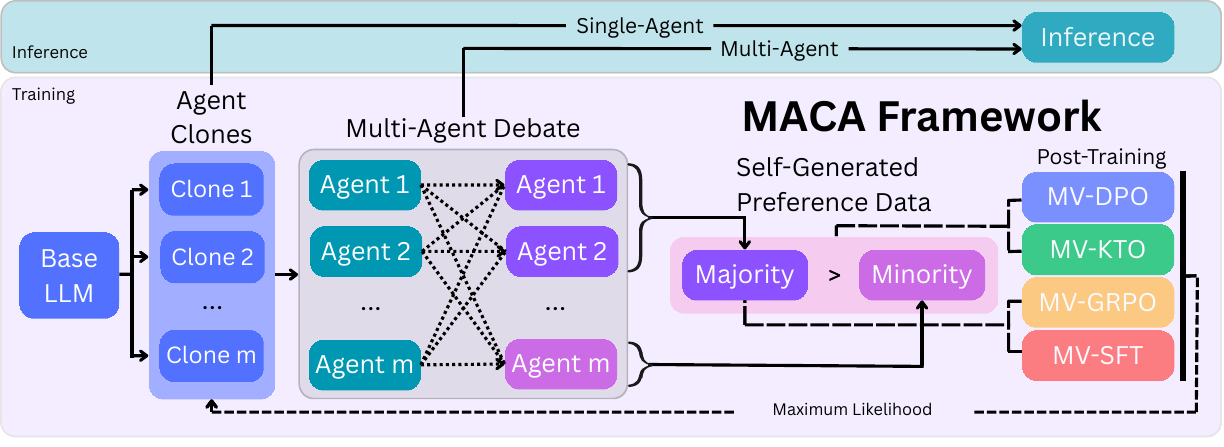}
    \caption{Multi-Agent Consensus Alignment framework: Multiple clones of a base LM engage in debate to generate majority and minority reasoning trajectories through multi-agent debate (collaborative exchange of reasoning). The framework splits responses based on alignment with majority consensus to create preference pairs. MV-GRPO compares online samples against majority signals, while MV-SFT imitates majority traces directly. In contrast, MV-DPO and MV-KTO utilize both positive (majority) and negative (minority) examples to learn to differentiate between these preference pairs. Updated agents can then be used for single-agent or multi-agent inference, or continue iterative training.}
    \label{fig:maca}
\end{figure*}

To achieve effective self-improvement in language models, we introduce Multi-Agent Consensus Alignment (MACA), a reinforcement learning framework that post-trains models using debate-derived signals to strengthen self-consistency and collaborative reasoning. Through multi-agent debate, models observe which reasoning patterns converge across peers and learn to favor those patterns through reinforcement on debate-derived consensus signals. This creates a positive feedback loop: improved self-consistency and ability to utilize peer context enables more effective debate, which generates higher-quality training signals for the next iteration (Fig.~\ref{fig:maca}).

A key design question in MACA is how to translate the multi-agent debate into a training signal. We explore the use of majority/minority labels as rewards through majority-vote GRPO~\citep{saa2024grpo}, as well as using them as preference labels for DPO~\citep{rafailov2023direct} and KTO~\citep{ethayarajh2024kto}. We find that preference learning is more effective: while both use majority/minority labels, preference learning over full reasoning traces allows models to learn to differentiate between reasoning traces that survived through deliberation and those that did not, capturing patterns where agents successfully reference and build on each other's reasoning. Crucially, this requires no ground-truth labels. Preference learning through relative comparison naturally grounds reasoning in the absence of external verification: when truth is ambiguous, judgments become robust through comparative assessment where majority reasoning provides a normative anchor while minority views introduce necessary variation~\citep{Festinger1954Comparison,moscovici1976social,nemeth1986minority,condorcet1785essai}. We teach models to ground their reasoning through this comparison of debate-derived trajectories, learning stable reasoning through peer deliberation alone.

\textbf{Key contributions.}
\begin{itemize}
\item \textbf{Method.} We introduce \MethodName, a reinforcement learning framework that post-trains models on consensus signals derived from multi-agent debate, teaching them to internalize collaborative reasoning patterns.
\item \textbf{Insight.} We demonstrate that consensus signals from multi-agent debate, grounded in peer interaction, provide richer and more robust training signals for self-improvement than single-round majority voting. This creates a positive feedback loop where improved reasoning enables more effective debate, which generates higher-quality training signals.
\item \textbf{Empirical validation.} We show that \MethodName\ improves multi-agent collaborative performance (+26.87\% on MATH), individual reasoning (+21.51\% on MathQA), and self-consistency (+27.6\% on GSM8K), with strong generalization to unseen reasoning domains (+16.3\% on GPQA, +11.6\% on CommonsenseQA).
\end{itemize}

\section{Related Work}
Existing approaches address sampling inconsistency primarily through inference-time techniques. Self-consistency prompting~\citep{Wang2022, li2024self} samples multiple reasoning paths and selects the majority-voted answer, with extensions for non-verifiable outputs~\citep{chen2023usc} and path pruning~\citep{zhu2024pathconsistency}. Multi-agent debate frameworks~\citep{du2024improving, irving2018debate} similarly utilize consensus across models to improve reliability, with recent work exploring applications in scientific discovery~\citep{gottweis2025ai}. However, these methods require additional inference compute and do not internalize self-consistency into the model. To improve self-consistency and single rollout performance through post-training, current training-time alternatives have limitations: relative log-probability rankings as a proxy for relative confidence correlate less strongly with accuracy than majority vote signals, as we show (App.~\ref{app:dpo_pair_selection}), while LLM-as-a-Judge approaches~\citep{jiao2025preference} suffer from preference leakage~\citep{li2025preference} and bias under ambiguity~\citep{Radharapu2025Arbiters}. We improve self-consistency through post-training that optimizes consensus signals via multi-agent RL~\citep{yang2021believe,jiang2021offline,zhu2024madiff,zhan2025exploiting}. Majority vote RL methods such as TTRL~\citep{Zuo2025TTRL} and ScPO~\citep{prasad2024self} use GRPO and DPO, respectively, to reinforce single-round majority vote from independently sampled responses. By contrast, MACA introduces multi-agent, multi-round debate, including deliberation traces as conditioning contexts within the prompts during RL. This exposes models to concrete examples of relative grounding where consensus emerges through explicit engagement with peer arguments, allowing them to learn not just to aggregate outputs but to internalize the deliberative process by which consensus is reached. Both TTRL and ScPO are specialized cases of the framework developed in this work, when reducing the multi-agent debate parameters to a single-round majority vote. For post-training the multi-agent debate setting specifically,~\citet{subramaniam2025multiagent} uses Supervised Fine-Tuning (SFT) for debate optimization, whereas we demonstrate that RL-based alternatives achieve superior performance. 

\section{Formalizing Self-Consistency}
\label{sec:formalsc}
Given a prompt $x$, an LM with parameters $\theta$ defines a distribution $\pi_\theta(y \,|\, x) = \prod_{t=1}^{|y|} \pi_\theta(y_t \,|\, x, y_{<t})$ over reasoning trajectories $y$, from which answers $a = \mathrm{A}(y)$ are extracted. Under temperature sampling, the model samples from a modified distribution $\pi_{\theta,\tau}(y\,|\,x)$ where token probabilities are adjusted by temperature $\tau > 0$. This induces an answer distribution $P_{\theta,\tau}(a\,|\,x) = \sum_{y: \mathrm{A}(y) = a} \pi_{\theta,\tau}(y\,|\,x)$, which gives each answer's probability by summing over all reasoning paths that lead to it. We denote the majority answer as $a^\star_{\theta,\tau}(x) = \arg\max_a P_{\theta,\tau}(a\,|\,x)$ with majority probability $S^+_{\theta,\tau}(x) = P_{\theta,\tau}(a^\star_{\theta,\tau}(x)\,|\,x)$. This represents the total probability mass concentrated on the most likely answer, or the model's internal consensus.

\looseness-1 Temperature sampling enables exploration of diverse reasoning paths, but reduces the consistency of the final answer. While greedy decoding ($\tau=0$) trivially approaches perfect consistency, it eliminates exploration and often produces suboptimal solutions~\citep{Holtzman2020Curious}. Lower temperatures increase consistency but restrict reasoning diversity. A self-consistent model should maintain high $S^+_{\theta,\tau}(x)$ even at high temperatures, allowing the model to access diverse reasoning trajectories while reliably converging on consistent answers. We measure self-consistency in two ways. 

\textbf{Single-agent sampling consistency.}
Computing $S^+_{\theta,\tau}(x)$ directly requires summing probabilities over all trajectories that lead to the majority answer, which is untractable. Instead, we estimate it by sampling $t$ independent trajectories with answers $a_1, \ldots, a_t$ and computing:
\begin{equation}
s_t^{\theta,\tau}(x) = \frac{1}{t} \sum_{i=1}^t \mathbf{1}[a_i(x) = \hat{a}(x)], \quad \text{where } \hat{a}(x) = \text{Majority}\{a_1(x), \ldots, a_t(x)\}.
\end{equation}

This measures the fraction of sampled trajectories that agree with the majority answer. As $t \to \infty$, $s_t^{\theta,\tau}(x) \to S^+_{\theta,\tau}(x)$, providing a consistent estimate of the true majority probability.

\textbf{Multi-agent debate agreement.}
When $M$ agents produce answers $a_1(x), \ldots, a_M(x)$ through deliberation, we measure the fraction of agents converging on the majority answer:
\begin{equation}
d_M^{\theta,\tau}(x) = \frac{1}{M} \sum_{m=1}^M \mathbf{1}[a_m(x) = \hat{a}(x)], \quad \text{where } \hat{a}(x) = \text{Majority}\{a_1(x), \ldots, a_M(x)\}.
\end{equation}

Higher agreement indicates a stronger consensus. In other words, models with higher $S^+_{\theta,\tau}(x)$ reach the same conclusion more frequently.

\section{\MethodName: Multi-Agent Consensus Alignment}
Having formalized self-consistency, we now present a framework to improve it through post-training with self-generated signals from debate. In multi-agent debate, $M$ copies of the same model engage in iterative discussion: each agent generates an initial response, then all agents see each other's reasoning and update their answers over $R-1$ subsequent rounds of deliberation. Answers that persist indicate stronger reasoning. The framework requires no external supervision: agents supervise themselves by learning from their own debate dynamics. Specifically, for each prompt $x$, the debate produces final responses $\mathcal{Y}(x) = \{y_1, \ldots, y_M\}$ with extracted answers $a_m = \mathcal{A}(y_m)$. The majority consensus $\hat{a}(x) = \text{Majority}\{a_1, \ldots, a_M\}$ partitions $\mathcal{Y}(x)$ into consensus-supporting $\mathcal{G}^+(x) = \{y \in \mathcal{Y}(x) : \mathcal{A}(y) = \hat{a}(x)\}$ and dissenting $\mathcal{G}^-(x) = \{y \in \mathcal{Y}(x) : \mathcal{A}(y) \neq \hat{a}(x)\}$ trajectories. This creates a fixed post-training dataset $\mathcal{D}_{\text{post}} = \{(x, \hat{a}(x), \mathcal{G}^+(x), \mathcal{G}^-(x))\}_{x \in \mathcal{D}}$ where $\mathcal{D}$ is the original set of prompts. Debate consensus, arising through deliberative exchange rather than statistical sampling, provides rich training signals. We adapt four post-training objectives to this self-generated data, treating consensus-supporting trajectories ($\mathcal{G}^+$) as preferred and dissenting trajectories ($\mathcal{G}^-$) as not preferred. By learning to separate these groups, the model internalizes the nuanced differences between stable consensus and dissenting reasoning. See Alg.~\ref{alg:maca} in App.~\ref{app:maca} for the complete iterative debate and post-training loop.

\textbf{Majority-Vote SFT (MV-SFT)} trains the model to mimic consensus-supporting trajectories:
\begin{equation}
\mathcal{L}_{\text{SFT}}(\theta) = -\mathbb{E}_{x \sim \mathcal{D}} \mathbb{E}_{y^+ \in \mathcal{G}^+(x)} [\log \pi_\theta(y^+|x)].
\end{equation}

\textbf{Majority-Vote GRPO (MV-GRPO)} uses online sampling with consensus-based rewards. For each prompt $x$, we sample multiple trajectories from the current policy and assign reward $r_x(y) = \mathbf{1}[\mathcal{A}(y) = \hat{a}(x)]$ based on whether each sample's answer matches the pre-computed consensus:
\begin{equation}
\mathcal{L}_{\text{GRPO}}(\theta) = -\mathbb{E}_{x \sim \mathcal{D}} \mathbb{E}_{y \sim \pi_\theta} \left[\tilde{A}_x(y) \sum_t \log \pi_\theta(y_t|x,y_{<t})\right] + \lambda \, \text{KL}(\pi_\theta \| \pi_{\text{ref}}),
\end{equation}

where $\tilde{A}_x(y) = r_x(y) - \bar{r}_x$ is the group-normalized advantage. We find that model inconsistency naturally yields both consensus and dissenting trajectories, allowing GRPO's group normalization to contrast majority/minority outcomes within batches.

\textbf{Majority-Vote DPO (MV-DPO)} follows the standard DPO formulation with preference pairs constructed from our pre-generated debate outcomes:
\begin{equation}
\mathcal{L}_{\text{DPO}}(\theta) = -\mathbb{E}_{x \sim \mathcal{D}} \mathbb{E}_{(y^+,y^-) \in \mathcal{G}^+(x) \times \mathcal{G}^-(x)} \left[\log \sigma\left(\beta \left[\log \frac{\pi_\theta(y^+|x)}{\pi_{\text{ref}}(y^+|x)} - \log \frac{\pi_\theta(y^-|x)}{\pi_{\text{ref}}(y^-|x)}\right]\right)\right].
\end{equation}

By contrasting the model's own consensus and dissenting trajectories, DPO's log probability ratios capture differences across entire reasoning chains, not just final answers, allowing each token to contribute to the preference signal.

\textbf{Majority-Vote KTO (MV-KTO)} applies KTO's unpaired formulation with debate-derived labels from our fixed dataset, with class-balancing weights $\lambda_+$ and $\lambda_-$:
\begin{equation}
\begin{aligned}
\mathcal{L}_{\text{KTO}}(\theta) = &-\lambda_+ \mathbb{E}_{x \sim \mathcal{D}} \mathbb{E}_{y^+ \in \mathcal{G}^+(x)} \left[\log \sigma\left(\beta \log \frac{\pi_\theta(y^+|x)}{\pi_{\text{ref}}(y^+|x)}\right)\right] \\
&-\lambda_- \mathbb{E}_{x \sim \mathcal{D}} \mathbb{E}_{y^- \in \mathcal{G}^-(x)} \left[\log \sigma\left(-\beta \log \frac{\pi_\theta(y^-|x)}{\pi_{\text{ref}}(y^-|x)}\right)\right].
\end{aligned}
\end{equation}
KTO's unpaired structure handles imbalanced 
outcomes where majority trajectories dominate.

\section{Results and Discussion}
\label{results}
We evaluate \MethodName\ by post-training four instruction-tuned small LMs (Qwen-2B~\citep{yang2024qwen2}, Llama-3B~\citep{LlaMA3team2024LlaMA3}, Phi-4B~\citep{abdin2024phi4}, and Llama-8B~\citep{LlaMA3team2024LlaMA3}) on six reasoning benchmarks (MATH~\citep{hendrycks2021measuring}, GSM8K~\citep{cobbe2021training}, MathQA~\citep{amini2019mathqa}, SVAMP~\citep{patel2021nlp}, GPQA~\citep{rein2023GPQA}, and Commonsense-QA~\citep{talmor-etal-2019-commonsenseqa}). We use 4-bit quantization with QLoRA~\citep{dettmers2023qlora}, limit responses to 256 tokens, and instantiate multi-agent debate with $M=3$ clones and $R=2$ rounds—all design choices motivated by the computational constraints of multi-agent debate RL training on a single-node cluster. With temperature $\tau=1.0$, this setup tests exploratory sampling under a budget adequate for efficient solvers on these benchmarks while remaining trainable with multiple concurrent agents. Detailed analysis of training efficiency and throughput is provided in App.~\ref{app:efficiency}. Critically, improvements persist when models are tested without token limits and without quantization (App.~\ref{app:ntcsc}, App.~\ref{app:boundscurves}), indicating that gains generalize beyond these specific settings. We compare (1) pre-trained models, (2) SFT baselines, and (3) \MethodName\ variants of GRPO, DPO, and KTO. We train and evaluate models on 1500/500 train/test splits for each dataset independently, unless otherwise specified, isolating task-specific performance improvements. We report mean agent accuracy with standard deviation across three seeds. Debate prompts, training parameters, multi-processing design, and other experimental details can be found in App. \ref{app:expsetup}.

\subsection{Post-Training on MAD Improves Debate Performance}
\label{accuracy_results}
We evaluate \MethodName\ in the multi-agent debate setting, examining how post-training improves both the final consensus accuracy and individual agent performance across debate rounds.

\begin{table*}[t]
\centering
\small
\setlength{\tabcolsep}{4pt} 
\begin{threeparttable}
\caption{Post-training multi-agent debate yields consistent accuracy improvements.}
\begin{tabular}{ll p{9mm} c|ccc c}
\toprule
\multirow{2}{*}{Model} & \multirow{2}{*}{Dataset}
& \multicolumn{2}{c|}{Pre / SFT (Baseline)} & \multicolumn{3}{c}{RL (Our Methods)} & \multirow{2}{*}{Best $\Delta$} \\
\cmidrule(lr){3-4}\cmidrule(lr){5-7}
& & Debate & MV-SFT & MV-GRPO & MV-KTO & MV-DPO & \\
\midrule
\multirow{3}{*}{Qwen2B}
  & MATH   & 32.40 & 37.07 $\pm$ 3.07 & 39.00 $\pm$ 1.74 & \textbf{46.47 $\pm$ 3.01} & 42.60 $\pm$ 1.78 & $\DeltaUp{14.07}$ \\
  & GSM8K  & 49.60 & 50.53 $\pm$ 1.36 & 54.13 $\pm$ 2.02 & \textbf{63.07 $\pm$ 0.64} & 58.47 $\pm$ 1.62 & $\DeltaUp{13.47}$ \\
  & MathQA & 24.20 & 26.27 $\pm$ 0.58 & 29.93 $\pm$ 4.99 & \textbf{32.60 $\pm$ 0.72} & 28.33 $\pm$ 0.31 & $\DeltaUp{9.13}$ \\
\addlinespace
\multirow{3}{*}{Llama3B}
  & MATH   & 37.80 & 35.33 $\pm$ 1.62 & 48.33 $\pm$ 2.19 & \textbf{52.93 $\pm$ 0.99} & 51.93 $\pm$ 1.67 & $\DeltaUp{15.27}$ \\
  & GSM8K  & 65.60 & 64.60 $\pm$ 1.59 & 68.60 $\pm$ 1.00 & \textbf{73.13 $\pm$ 0.83} & 71.67 $\pm$ 3.03 & $\DeltaUp{7.80}$ \\
  & MathQA & 21.60 & 40.07 $\pm$ 2.00 & 48.73 $\pm$ 1.60 & \textbf{64.00 $\pm$ 0.53} & 63.13 $\pm$ 1.89 & $\DeltaUp{42.73}$ \\
\addlinespace
\multirow{3}{*}{Phi4B}
  & MATH   & 44.40 & 45.53 $\pm$ 2.53 & \textbf{49.93 $\pm$ 1.33} & 45.27 $\pm$ 0.70 & 46.73 $\pm$ 1.67 & $\DeltaUp{5.53}$ \\
  & GSM8K  & 79.60 & 78.93 $\pm$ 0.61 & 82.67 $\pm$ 1.81 & 82.47 $\pm$ 1.14 & \textbf{84.73 $\pm$ 0.31} & $\DeltaUp{5.13}$ \\
  & MathQA & 49.60 & 50.87 $\pm$ 1.62 & \textbf{63.07 $\pm$ 1.21} & 51.53 $\pm$ 1.14 & 51.40 $\pm$ 1.44 & $\DeltaUp{13.47}$ \\
\addlinespace
\multirow{3}{*}{Llama8B}
  & MATH   & 32.80 & 34.13 $\pm$ 0.70 & 45.93 $\pm$ 1.03 & 53.93 $\pm$ 1.80 & \textbf{59.67 $\pm$ 1.33} & $\DeltaUp{26.87}$ \\
  & GSM8K  & 74.00 & 66.27 $\pm$ 1.01 & 81.53 $\pm$ 2.81 & 81.00 $\pm$ 1.97 & \textbf{81.93 $\pm$ 1.51} & $\DeltaUp{8.60}$ \\
  & MathQA & 44.60 & 44.13 $\pm$ 1.10 & 57.27 $\pm$ 0.61 & 62.00 $\pm$ 2.03 & \textbf{69.27 $\pm$ 1.55} & $\DeltaUp{24.67}$ \\
\bottomrule
\end{tabular}
\label{tab:r1mv_multiagent}
\end{threeparttable}
\end{table*}

\begin{figure*}[t]
  \centering
  \includegraphics[trim={20pt 350pt 20pt 10pt},clip,width=\linewidth]{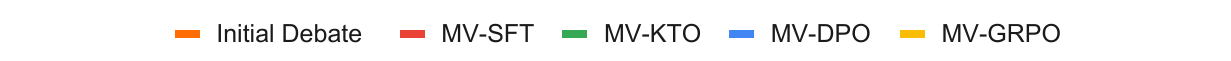}

  \begin{subfigure}{0.33\textwidth}
    \includegraphics[trim={20pt 0pt 20pt 50pt},clip,width=\linewidth]{Llama3Bmathqa.pdf}
    \caption{Llama-3B on MATHQA}
    \label{fig:sub1}
  \end{subfigure}%
  \hfill
  \begin{subfigure}{0.33\textwidth}
    \includegraphics[trim={20pt 0pt 20pt 50pt},clip,width=\linewidth]{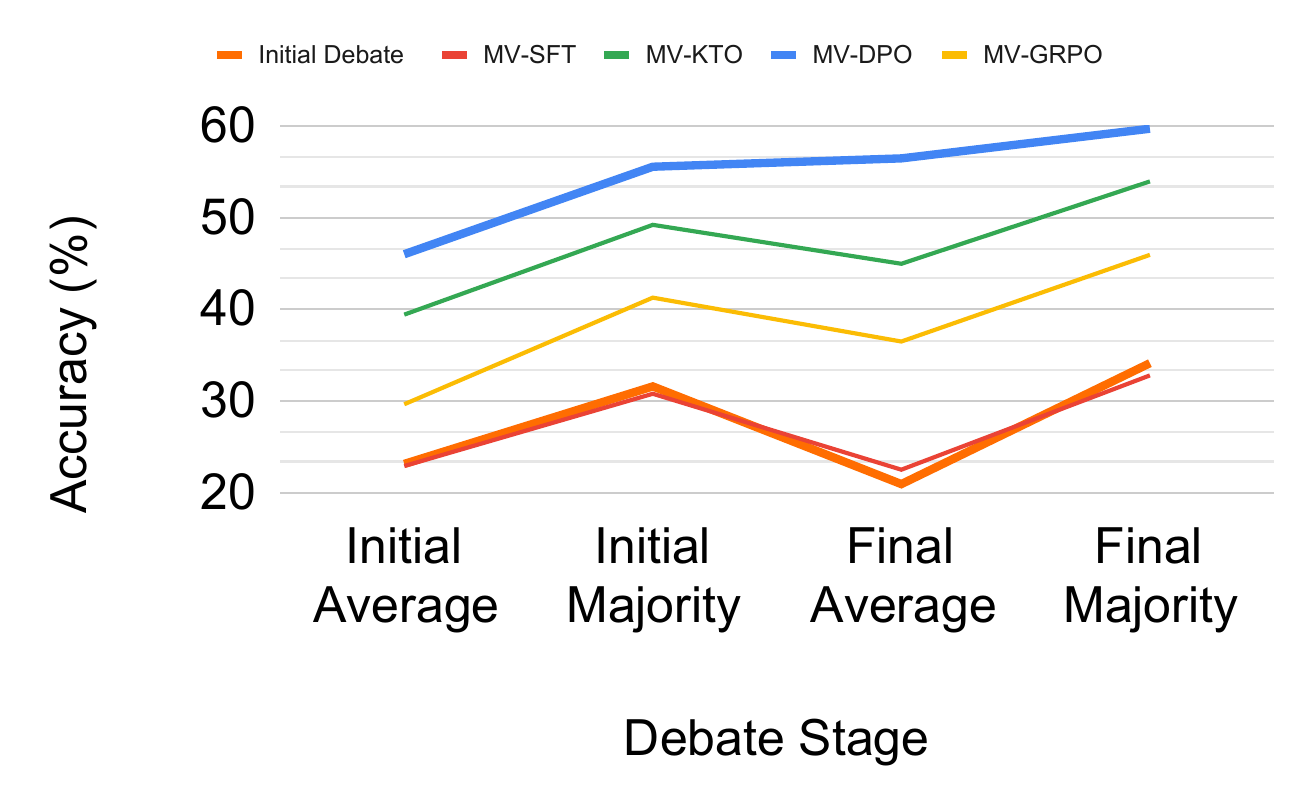}
    \caption{Llama-8B on MATH}
    \label{fig:sub2}
  \end{subfigure}%
  \hfill
  \begin{subfigure}{0.33\textwidth}
    \includegraphics[trim={20pt 0pt 20pt 50pt},clip,width=\linewidth]{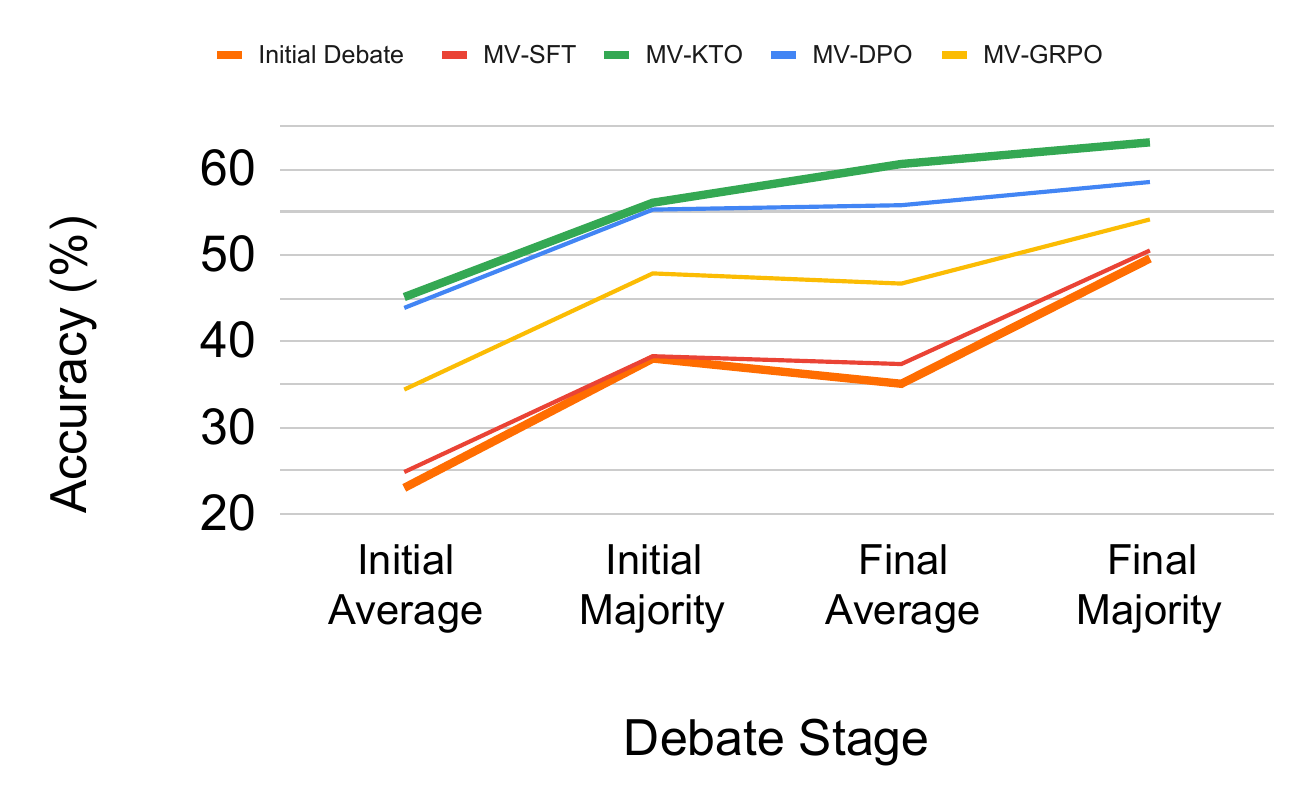}
    \caption{Qwen-2B on GSM8K}
    \label{fig:sub3}
  \end{subfigure}

  \caption{Debate-aware RL improves all stages of multi-agent debate. Incorporating debate context in RL teaches agents to leverage prior arguments, improving final-round consensus. Stages: initial round average, initial round majority vote, final round average, final round majority vote.}
  \label{fig:debatefrontiers}
\end{figure*}

\textbf{Impact on multi-agent debate setting.}
We evaluate \MethodName\ in the multi-agent setting using final-round majority-vote accuracy (the consensus after debate). Baselines are (1) the base model's debate outcome~\citep{du2024improving} and (2) MV-SFT on majority traces~\citep{subramaniam2025multiagent}. As shown in Table~\ref{tab:r1mv_multiagent}, post-training on debate-derived signals improves ensemble accuracy across all models and datasets. Preference learning variants (MV-DPO and MV-KTO) provide the largest and most consistent gains, up to +42.73 percentage points, by directly optimizing the log-probability gap between trajectories, with MV-KTO performing best on smaller models ($\le3B$) and MV-DPO on larger ones ($4-8B$).
 Learning to differentiate between full reasoning trajectory pairs appears to better address credit assignment challenges in sparse final-answer supervision compared to both MV-GRPO's scalar rewards and MV-SFT's imitation learning, simultaneously improving answer accuracy while reinforcing higher-quality and more concise intermediate reasoning steps (App.~\ref{app:formatreward}).

\textbf{Impact on debate dynamics.} Fig.~\ref{fig:debatefrontiers} summarizes four metrics: initial round average single-agent accuracy, initial round ensemble majority vote (not used in the debate), final round average single-agent accuracy (conditioned on debate context/peer chains-of-thought), and final round ensemble majority vote (consensus). Post-training produces the largest gains where agents leverage peer feedback, since it directly teaches effective peer context utilization. We also observe that post-training mitigates debate-driven performance degradation observed in MATHQA for Llama-3B. The contrastive signal allows models to learn from consensus patterns even when exposed to flawed reasoning trajectories, correcting rather than amplifying poor grounding behaviors.

\textbf{Impact on a single agent in zero-shot setting.}
 Post-training on debate also improves the performance of the individual agents in the inital round (see Initial Average in Fig.~\ref{fig:debatefrontiers}): across 12 model-dataset pairs, MV-RL methods consistently outperform the Base and MV-SFT baselines (Table~\ref{tab:r0avg_pretty_split}) in single-agent zero-shot (single trajectory) settings. Self-guided preference learning (MV-DPO and MV-KTO) outperforms scalar rewards via MV-GRPO for all models except Phi-4B. MV-DPO is best in 7/12 cases, while MV-KTO is better in some cases for smaller LMs.

\begin{table*}[t]
\centering
\small
\setlength{\tabcolsep}{4pt} 
\begin{threeparttable}
\caption{Accuracy impact of \MethodName\ on single agent performance in zero-shot setting.}
\begin{tabular}{ll p{9mm} c|ccc c}
\toprule
\multirow{2}{*}{Model} & \multirow{2}{*}{Dataset}
& \multicolumn{2}{c|}{Base / SFT (Baseline)} & \multicolumn{3}{c}{RL (Our Methods)} & \multirow{2}{*}{Best $\Delta$} \\
\cmidrule(lr){3-4}\cmidrule(lr){5-7}
& & Base & MV-SFT & MV-GRPO & MV-KTO & MV-DPO & \\
\midrule
\multirow{3}{*}{Qwen2B}
  & MATH   & 7.67  & 11.51 $\pm$ 0.60 & 18.09 $\pm$ 0.71 & 20.18 $\pm$ 0.67 & \textbf{23.49 $\pm$ 2.30} & $\DeltaUp{15.82}$ \\
  & GSM8K  & 23.00 & 24.84 $\pm$ 0.87 & 34.40 $\pm$ 2.08 & \textbf{45.13 $\pm$ 1.80} & 43.87 $\pm$ 1.92 & $\DeltaUp{22.71}$ \\
  & MathQA & 5.00 & 5.25 $\pm$ 0.33 & 17.27 $\pm$ 2.88 & \textbf{22.16 $\pm$ 1.14} & 20.91 $\pm$ 0.47  & $\DeltaUp{17.27}$ \\
\addlinespace
\multirow{3}{*}{Llama3B}
  & MATH   & 27.87 & 25.89 $\pm$ 0.56 & 35.22 $\pm$ 0.44 & 40.64 $\pm$ 1.25 & \textbf{40.71 $\pm$ 0.08} & $\DeltaUp{13.26}$ \\
  & GSM8K  & 57.33 & 55.98 $\pm$ 0.68 & 52.40 $\pm$ 2.84 & \textbf{65.76 $\pm$ 1.44} & 64.98 $\pm$ 1.67 & $\DeltaUp{8.80}$ \\
  & MathQA & 23.87 & 23.44 $\pm$ 0.73 & 30.09 $\pm$ 1.98 & 42.84 $\pm$ 0.67 & \textbf{45.00 $\pm$ 2.23} & $\DeltaUp{21.13}$ \\
\addlinespace
\multirow{3}{*}{Phi4B}
  & MATH   & 34.60 & 34.60 $\pm$ 0.82 & \textbf{37.42 $\pm$ 0.16} & 33.84 $\pm$ 0.78 & 34.62 $\pm$ 1.48 & $\DeltaUp{2.82}$ \\
  & GSM8K  & 67.27 & 69.58 $\pm$ 0.76 & 67.13 $\pm$ 3.60 & 75.60 $\pm$ 1.80 & \textbf{76.87 $\pm$ 0.36} & $\DeltaUp{9.84}$ \\
  & MathQA & 34.87 & 34.04 $\pm$ 0.58 & \textbf{45.52 $\pm$ 2.19} & 33.91 $\pm$ 0.16 & 33.91 $\pm$ 0.50 & $\DeltaUp{10.65}$ \\
\addlinespace
\multirow{3}{*}{Llama8B}
  & MATH   & 22.93 & 23.16 $\pm$ 0.14 & 29.66 $\pm$ 1.27 & 39.42 $\pm$ 0.44 & \textbf{46.00 $\pm$ 0.35} & $\DeltaUp{23.07}$ \\
  & GSM8K  & 57.93 & 42.09 $\pm$ 1.28 & 62.45 $\pm$ 6.01 & 72.36 $\pm$ 1.34 & \textbf{77.36 $\pm$ 0.27} & $\DeltaUp{19.43}$ \\
  & MathQA & 29.67 & 30.84 $\pm$ 0.60 & 33.07 $\pm$ 1.11 & 38.42 $\pm$ 1.22 & \textbf{51.18 $\pm$ 0.24} & $\DeltaUp{21.51}$ \\
\bottomrule
\end{tabular}
\label{tab:r0avg_pretty_split}
\end{threeparttable}
\end{table*}

\begin{figure*}[t]
  \centering
  \vspace{0.5em}
  \includegraphics[trim={0pt 25pt 0pt 310pt},clip,width=\linewidth]{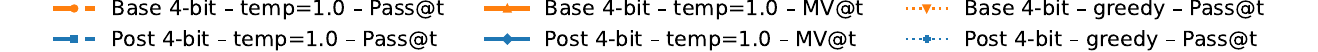}

  \begin{subfigure}{0.33\textwidth}
    \includegraphics[trim={110pt 50pt 110pt 0pt},clip,width=\linewidth]{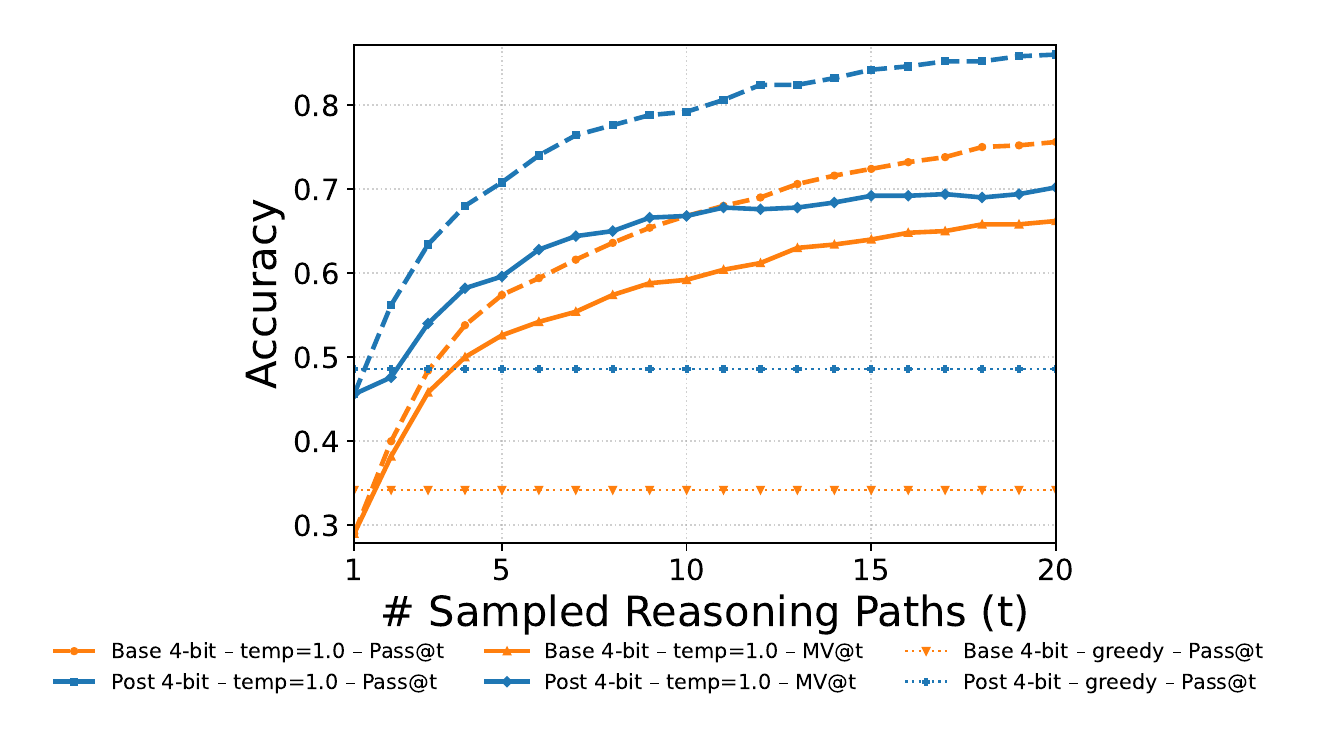}
    \caption{Llama-3B on MATHQA}
    \label{fig:sc_sub1}
  \end{subfigure}%
  \begin{subfigure}{0.33\textwidth}
    \includegraphics[trim={110pt 50pt 110pt 0pt},clip,width=\linewidth]{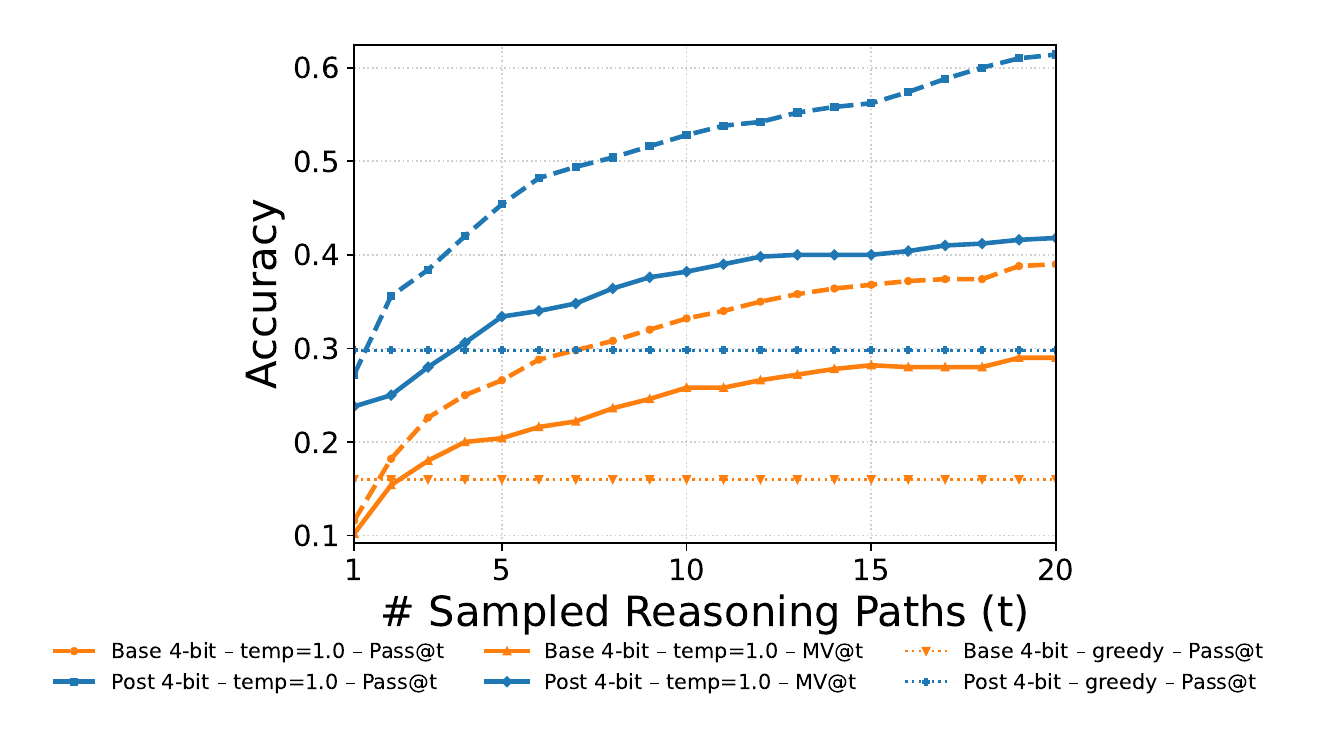}
    \caption{Qwen-2B on MATH}
    \label{fig:sc_sub2}
  \end{subfigure}
  \begin{subfigure}{0.33\textwidth}
    \includegraphics[trim={110pt 50pt 110pt 0pt},clip,width=\linewidth]{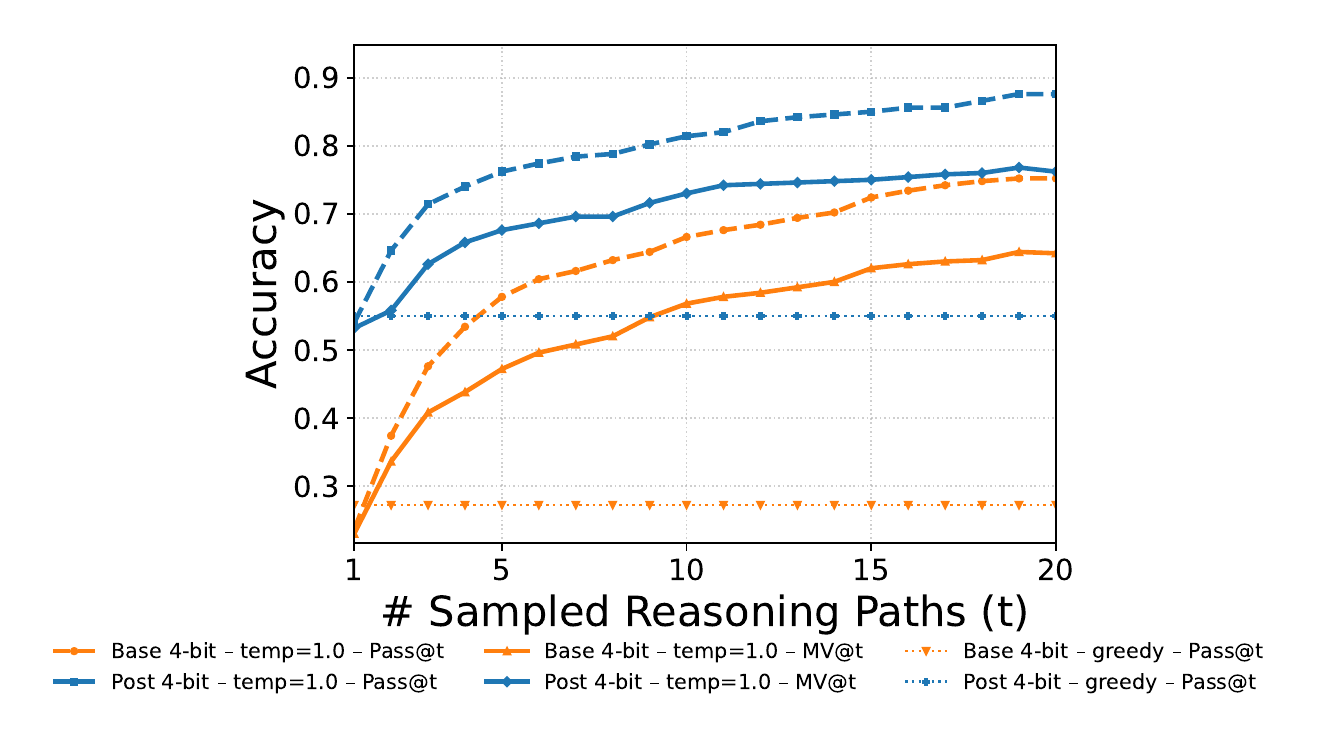}
    \caption{Qwen-2B on GSM8K}
    \label{fig:sc_sub3}
  \end{subfigure}

  \caption{Post-training self-consistency improves sampling accuracy.
  Dashed: Pass@t (oracle upper bound), solid: MV@t (majority over $t$ samples), dotted: greedy ($\tau=0$) accuracy.
  \textcolor{blue}{(Blue)}: post-trained model. \textcolor{orange}{(Orange)}: base model. Curves computed over 500 prompts.}
  \label{fig:pass}
\end{figure*}

\subsection{Post-Training on MAD Improves Self-Consistency and Accuracy}\label{consistency_results}
We measure the effect of post-training on sampling consistency $s_t^{\theta,\tau}(x)$, the fraction of sampled trajectories that match the majority answer. As formalized in Sec.~\ref{sec:formalsc}, we track the sampling consistency where $s_t^{\theta,\tau}(x)$ converges to the modal probability $S^+_{\theta,\tau}(x)$ as $t \to \infty$. For each model, we sample 20 trajectories on 500 held-out prompts and evaluate $s_t^{\theta,\tau}(x)$ for $t=1$ to $20$ (App.~\ref{app:consistency_curves}). At \(t=1\), this metric primarily captures answer completeness, i.e., the percentage of responses that produce parseable answers within the token window, which post-training substantially improves. MV-DPO and MV-KTO achieve these gains through self-supervised preference learning alone, demonstrating that models can teach themselves more efficient reasoning without format rewards (App.~\ref{app:formatreward}), which can otherwise be spurious~\citep{huang2023cannot, srivastava2025robust}. As sample size increases, the metric transitions to measuring true cross-sample agreement. The curves stabilize up to 27.6 percentage points above baseline, demonstrating that post-training increases answer concentration: models more consistently sample trajectories that converge despite high-temperature exploration. These improvements persist when tested without max token constraints (App.~\ref{app:ntcsc}).

In multi-agent debate, we evaluate the agreement metric $d_M^{\theta,\tau}(x)$ defined in Sec.~\ref{sec:formalsc}. Base small LMs initially struggle to reach meaningful consensus: for Qwen-2B on GSM8K, most of the ``consensus'' comes from random tie-breaking (1/3 agreement) or weak majorities (2/3), with only 13.4\% reaching unanimity (Fig.~\ref{fig:agreement_rates}, App.~\ref{app:agreement_rates}). Post-training with \MethodName\ thoroughly addresses this: non-parseable responses drop from 13.8\% to 0.6\%, no-agreement cases from 45.6\% to 19.8\%, while unanimous agreement triples from 13.4\% to 43.4\%. This confirms that \MethodName\ improves both individual reasoning quality and collaborative grounding, enabling genuine consensus rather than noisy aggregation (agreement distributions in App.~\ref{app:agreement_rates}).

\textbf{Impact on inference-time sampling.}
We next examine how gains in self-consistency translate to inference-time performance under various sampling regimes. For each prompt, we draw \(t\) trajectories and report two metrics: Pass@t, the fraction of prompts for which at least one of the first \(t\) samples is correct~\citep{chen2021evaluating}, and MV@t, the fraction for which a majority vote over those \(t\) samples matches the ground truth (ties counted as incorrect)~\citep{li2024self}. As shown in Fig.~\ref{fig:pass}, post-training (blue) (i) lifts greedy (\(\tau{=}0\)) accuracy, (ii) increases MV@t at each fixed \(t\), and (iii) raises Pass@t, the empirical sampling upper bound, indicating higher achievable accuracy at any given sampling budget. When additional inference compute is available (i.e., \(t>1\)), sampling techniques continue to see gains on top of the post-training improvements, showing \MethodName\ is complementary to inference-time sampling. The same pattern holds with full-precision, with improvements comparable to their 4-bit counterparts (App.~\ref{app:boundscurves}). Finally, self-consistency gains strongly correlate with accuracy improvements ($r > 0.86$ across all tested inference conditions; see App.~\ref{app:correlation}).

\begin{table*}[t!]
\caption{Post-training self-consistency improves performance across general reasoning benchmarks. Models trained on datasets (columns) tested on benchmarks (rows). Bottom rows show generalization to unseen benchmarks: SVAMP (math), GPQA (science), CSQA (commonsense). All = joint training on combined datasets. Arrows show absolute gains over instruction-tuned model.}
\centering
\begin{tabular}{l|r|rrrr|r|rrrr}
\toprule
\multirow{2}{*}{Test} & \multicolumn{5}{c|}{Qwen2B (post-trained on)} & \multicolumn{5}{c}{Llama3B (post-trained on)} \\
\cmidrule(lr){2-6} \cmidrule(l){7-11}
 & Base & MATH & GSM & MQA & All & Base & MATH & GSM & MQA & All \\
\midrule
MATH    & 10.4 & \textbf{$\DeltaUpGreen{10.0}$} & $\DeltaUpGreen{3.8}$ & $\DeltaUpGreen{10.8}$ & $\DeltaUpGreen{12.2}$ & 32.0 & $\DeltaUpGreen{19.4}$ & $\DeltaUpGreen{16.4}$ & $\DeltaUpGreen{18.2}$ & \textbf{$\DeltaUpGreen{21.2}$} \\
GSM   & 27.0 & $\DeltaUpGreen{20.0}$ & \textbf{$\DeltaUpGreen{25.6}$} & $\DeltaUpGreen{22.6}$ & $\DeltaUpGreen{27.8}$ & 69.6 & $\DeltaUpGreen{6.0}$ & $\DeltaUpGreen{6.8}$ & $\DeltaUpGreen{8.4}$ & \textbf{$\DeltaUpGreen{10.8}$} \\
MQA  & 7.4  & $\DeltaUpGreen{12.6}$ & $\DeltaUpGreen{17.0}$ & $\DeltaUpGreen{15.4}$ & \textbf{$\DeltaUpGreen{21.4}$} & 24.6 & $\DeltaUpGreen{14.0}$ & $\DeltaUpGreen{13.4}$ & $\DeltaUpGreen{21.2}$ & \textbf{$\DeltaUpGreen{21.6}$} \\
\midrule
SVAMP   & 48.3 & \textbf{$\DeltaUpGreen{19.0}$} & $\DeltaUpGreen{18.0}$ & $\DeltaUpGreen{17.0}$ & $\DeltaUpGreen{27.7}$ & 71.3 & $\DeltaUpGreen{6.0}$ & $\DeltaUpGreen{6.4}$ & $\DeltaUpGreen{9.7}$ & \textbf{$\DeltaUpGreen{7.1}$} \\
GPQA    & 0.5  & $\DeltaUpGreen{6.0}$  & $\DeltaUpGreen{5.3}$  & \textbf{$\DeltaUpGreen{12.8}$} & $\DeltaUpGreen{16.3}$ & 0.7  & $\DeltaUpGreen{5.4}$  & $\DeltaUpGreen{6.3}$ & $\DeltaUpGreen{9.8}$ & \textbf{$\DeltaUpGreen{10.7}$} \\
CSQA    & 3.8  & $\DeltaUpGreen{19.8}$ & $\DeltaUpGreen{43.0}$ & $\DeltaUpGreen{54.0}$ & \textbf{$\DeltaUpGreen{59.6}$} & 53.0 & $\DeltaUpGreen{7.4}$ & $\DeltaUpGreen{10.6}$ & \textbf{$\DeltaUpGreen{11.6}$} & $\DeltaUpGreen{11.0}$ \\
\bottomrule
\end{tabular}
\label{tab:robustness}
\end{table*}

\begin{table*}[t!]
\caption{Multi-agent: Post-training with debate (MV) is comparable to ground-truth (GT).}
\centering
\footnotesize
\setlength{\tabcolsep}{3.5pt}
\begin{tabular}{ll
                c
                c c
                c c
                c c
                c c}
\toprule
Llama-8B & Dataset
& Debate
& \multicolumn{2}{c}{SFT}
& \multicolumn{2}{c}{KTO}
& \multicolumn{2}{c}{DPO}
& \multicolumn{2}{c}{GRPO} \\
\cmidrule(lr){4-5}\cmidrule(lr){6-7}\cmidrule(lr){8-9}\cmidrule(lr){10-11}
& &
  & \textit{GT} & \textit{MV}
  & \textit{GT} & \textit{MV}
  & \textit{GT} & \textit{MV}
  & \textit{GT} & \textit{MV} \\
\midrule
\multirow{2}{*}{Single-Agent}
  & MATH  & 22.93
           & \bfseries 23.73 & 23.00
           & \bfseries 41.20 & 39.40
           & 45.13 & \bfseries 46.40
           & 29.07 & \bfseries 31.13 \\
  & GSM8K & 57.93
           & 40.20 & \bfseries 41.67
           & \bfseries 72.60 & 70.87
           & 76.33 & \bfseries 77.67
           & 61.27 & \bfseries 66.87 \\
\multirow{2}{*}{Multi-Agent}
  & MATH  & 32.80
           & \bfseries 34.00 & 33.40
           & \bfseries 56.00 & 55.80
           & \bfseries 61.80 & 60.80
           & \bfseries 48.60 & 44.80 \\
  & GSM8K & 74.00
           & \bfseries 65.20 & \bfseries 65.20
           & \bfseries 81.20 & 79.40
           & 81.60 & \bfseries 83.0
           & 83.20 & \bfseries 84.20 \\
\bottomrule
\end{tabular}
\label{tab:Llama8B_mv_gt_ablation}
\end{table*}

\textbf{Improving self-consistency on math datasets improves generalization to unseen benchmarks.}
These consistency and accuracy gains extend beyond the training distribution: training to improve internal self-consistency on any mathematical dataset enhances performance across diverse reasoning tasks, including previously unseen domains. Building on recent work showing math training enhances general reasoning~\citep{deepseek2025, nemotron2025}, we demonstrate that improving self-consistency---internal consensus strength---on math is essential for reliably unlocking these capabilities. Table~\ref{tab:robustness} reports results for models trained on MATH, GSM8K, or MathQA individually, and on all three combined (All), using MV-DPO. Training on any single dataset improves performance across all reasoning tasks, including unseen math (SVAMP), science (GPQA), and commonsense reasoning (CSQA). Joint training achieves further improvements across nearly every benchmark, demonstrating that diverse training data amplifies self-consistency gains.

\subsection{Ablation Study}
To understand the sources of performance gains from \MethodName, we conduct ablation studies examining key components. We show that self-generated consensus signals outperform ground-truth supervision, including peer context during training improves relative grounding and debate utilization, and multi-round debate provides stronger signals than single-round majority vote. We also demonstrate in App.~\ref{app:formatreward} that self-supervised preference learning serves as an effective implicit format reward by reinforcing more efficient and concise chain-of-thought construction, though most gains stem from fundamental reasoning improvements rather than formatting alone.

\textbf{RL with debate is comparable to RL with ground-truth.} Table~\ref{tab:Llama8B_mv_gt_ablation} compares post-training with debate majority-vote labels, derived from the model’s own consensus without external supervision, to ground-truth labels on Llama-3B.
Across post-training methods and in both single- and multi-agent settings, using Debate-MV is consistently comparable to ground-truth supervision, a trend that holds across other models as well (App. \ref{app:gtmvcomp}). We additionally show in App. \ref{app:macageneralization} that this comparable performance between Debate-MV and ground-truth supervision extends to general reasoning improvements under \MethodName\ across all proposed post-training methods and reasoning benchmarks tested, as observed in Table~\ref{tab:robustness}. Because Debate-MV scales naturally with sample size, its parity with ground-truth highlights self-supervised alignment as a promising direction.

\begin{table*}[t!]
\caption{Single-agent: RL on debate majority-vote with peer context (MV-DPO) outperforms RL on single-round majority vote (TTRL, ScPO). Arrows show gains over instruction-tuned model. Statistics gathered over 3 seeds.}
\centering
\small
\setlength{\tabcolsep}{4pt}
\begin{tabular}{lrrrr|rrrr}
\toprule
& \multicolumn{4}{c|}{MATH} & \multicolumn{4}{c}{GSM8K} \\
\cmidrule(lr){2-5}\cmidrule(lr){6-9}
& Qwen-2B & Llama-3B & Phi-4B & Llama-8B & Qwen-2B & Llama-3B & Phi-4B & Llama-8B \\
\midrule
TTRL      & $\DeltaUpGreen{\textbf{18.0}}_{\pm 2.9}$ & $\DeltaUpGreen{5.3}_{\pm 5.7}$ & $\DeltaUpGreen{6.1}_{\pm 2.1}$ & $\DeltaUpGreen{7.5}_{\pm 0.2}$ & $\DeltaUpGreen{\textbf{24.5}}_{\pm 1.6}$ & $\DeltaDownRed{22.3}_{\pm 8.3}$ & $\DeltaUpGreen{4.8}_{\pm 1.6}$ & $\DeltaDownRed{13.5}_{\pm 5.7}$ \\
ScPO       & $\DeltaUpGreen{2.3}_{\pm 1.1}$ & $\DeltaUpGreen{3.4}_{\pm 0.2}$ & $\DeltaUpGreen{0.1}_{\pm 0.5}$ & $\DeltaUpGreen{3.7}_{\pm 0.6}$ & $\DeltaUpGreen{15.9}_{\pm 3.4}$ & $\DeltaUpGreen{1.6}_{\pm 0.8}$ & $\DeltaUpGreen{5.9}_{\pm 0.5}$ & $\DeltaUpGreen{12.9}_{\pm 1.2}$ \\
MV-DPO       & $\DeltaUpGreen{16.7}_{\pm 0.4}$ & $\DeltaUpGreen{\textbf{12.5}}_{\pm 0.7}$ & $\DeltaUpGreen{\textbf{6.9}}_{\pm 0.2}$ & $\DeltaUpGreen{\textbf{17.1}}_{\pm 0.8}$ & $\DeltaUpGreen{23.8}_{\pm 0.9}$ & $\DeltaUpGreen{\textbf{3.7}}_{\pm 0.3}$ & $\DeltaUpGreen{\textbf{7.3}}_{\pm 0.3}$ & $\DeltaUpGreen{\textbf{18.2}}_{\pm 0.2}$ \\
\bottomrule
\end{tabular}
\label{tab:debate_vs_mv_improvements}
\end{table*}


\textbf{RL with debate outperforms RL with single-round majority vote.} We evaluate three training conditions in the single-agent setting to understand the impact of training on the debate versus initial round majority vote on the individual agents themselves, as shown in Table \ref{tab:debate_vs_mv_improvements}: (1) initial round majority vote with GRPO (akin to TTRL~\citep{Zuo2025TTRL}), (2) initial round majority vote with DPO (akin to ScPO without the weighted loss term ~\citep{prasad2024self}), and (3) debate majority vote with DPO including peer context (MV-DPO from MACA). Training on the debate signal improves single-agent performance more than reinforcing the initial round majority vote. Additionally, Table~\ref{tab:context_ablation} shows that including peer context during preference learning (MV-DPO) improves multi-agent debate performance (more models/datasets in App. \ref{app:context}), with the largest gains in final-round individual accuracy, reflecting better use of peer chains-of-thought during deliberation (Fig. \ref{fig:debatefrontiers}).

\begin{table}[t]
\caption{Multi-agent: Post-training with debate context (peer CoTs) outperforms no context. Models: 3B = Llama-3B; 4B = Phi-4B; 8B = Llama-8B. Dataset: GSM8K}
\vspace{0.5em}
\centering
\footnotesize
\setlength{\tabcolsep}{4pt}
\renewcommand{\arraystretch}{0.95}
\begin{tabular}{l c c c c c c}
\toprule
& \multicolumn{2}{c}{Initial Round Avg} &
  \multicolumn{2}{c}{Final Round Avg} &
  \multicolumn{2}{c}{Final Round MV} \\
\cmidrule(lr){2-3}\cmidrule(lr){4-5}\cmidrule(lr){6-7}
 & NoCtx & Context & NoCtx & Context & NoCtx & Context \\
\midrule
3B  & 65.87 & \bfseries 66.73 & 67.87 & \bfseries 70.60 & 70.60 & \bfseries 73.80 \\
4B  & 74.67 & \bfseries 76.60 & 74.60 & \bfseries 77.80 & 81.80 & \bfseries 84.40 \\
8B  & 75.80 & \bfseries 77.67 & 78.00 & \bfseries 81.20 & 81.60 & \bfseries 83.00 \\
\bottomrule
\end{tabular}
\label{tab:context_ablation}
\end{table}

\textbf{Debate improves consensus quality.} Finally, the debate process itself improves consensus signals. Base small LMs produce mostly random or weak majorities initially, but post-training increases unanimous agreement from 27.2\% to 43.4\% between initial and final rounds (Qwen-2B on GSM8K, App. \ref{app:agreement_rates}). This iterative refinement provides more reliable training signals than independent sampling. While majority vote correlates strongly with accuracy (alternative ranking signals like log-probability are explored in App. \ref{app:dpo_pair_selection}), post-training debate creates a self-reinforcing cycle: better debate generates higher-quality signals, producing models that debate more effectively. Iterative training yields continued improvements with diminishing returns, as shown in App.~\ref{app:iter}.

\section{Conclusion, Limitations, and Future Work}
We introduce self-consistency as an intrinsic property of well-aligned reasoning and present \MethodName, a self-supervised framework that teaches models to reliably sample coherent reasoning through reinforcement of internal consensus signals. Without external supervision, \MethodName\ drives substantial improvements: +27.6\% self-consistency on GSM8K, +22.4\% Pass@20 on MATH, +23.7\% zero-shot accuracy on MATH, strong generalization to unseen benchmarks (+16.3\% GPQA, +11.6\% CommonsenseQA), and enhanced multi-agent performance (+42.7\% on MathQA). Through debate, agents attempt to ground their reasoning in peer context, and we select trajectories where such deliberation aligns with internal consensus. These consensus-aligned examples provide natural supervision for stable reasoning patterns, enabling models to self-improve their reasoning consistency, efficiency, and accuracy without explicit chain-of-thought supervision or external answer verification. Through preference learning on debate signals, models produce more concise chains-of-thought without format rewards and correct cases where aggregation previously led to degeneration, significantly improving the robustness of multi-agent debate. While \MethodName\ has limitations, including requiring sufficient base model competence to generate meaningful consensus signals, potentially amplifying existing biases in model outputs, and lacking direct supervision of intermediate reasoning correctness, it demonstrates that consensus reinforcement has robust benefits. Future work could explore alternative consensus methods, confidence-weighted voting, heterogeneous agents, and better leveraging of minority traces. Sizeable gains on difficult unseen tasks (+10.0\% on AMC, App.~\ref{app:formatreward}) suggest consensus alignment unlocks latent capabilities beyond consistency itself. \MethodName\ demonstrates that language models can effectively harness internal deliberation for self-alignment, elevating their reasoning capabilities autonomously.

\clearpage
\newpage
\bibliographystyle{assets/plainnat}
\bibliography{iclr2026}

\clearpage
\newpage
\beginappendix

\section{Examples of Post-Trained Consistent Reasoning in Single- and Multi- Agent Settings}
\label{app:debate-analysis}
This appendix presents examples from the debates, providing a qualitative signal of how post-training affects both single-agent reasoning and multi-agent collaboration. The evidence highlights visible improvements in mathematical reasoning, consensus formation, and systematic problem-solving.

\subsection{Overview of Key Findings}
Analysis of the transcripts reveals that post-training leads to more systematic and efficient reasoning in single-agent settings. Models demonstrate clearer step-by-step problem decomposition, make fewer calculation errors, and show improved conceptual understanding. Notably, responses become more concise and are less likely to be truncated due to inefficient reasoning.

In multi-agent debates, post-training enhances collaborative effects. Agents more frequently reference and build upon each other's solutions, leading to more robust cross-verification and consensus formation. We observe that post-trained agents are more likely to resolve mistakes through discussion rather than propagate errors.

Our analysis shows that preference learning teaches models higher-quality and more concise reasoning without requiring explicit format rewards. The consensus mechanism itself becomes more effective at aggregating individual outputs, supporting both in-distribution and out-of-distribution transfer.

\subsection{Case Study 1: Single-Agent Mathematical Reasoning}
\label{app:single-agent-case}

\subsubsection{Problem: Cities A and B Relative Speed}

\begin{tcolorbox}[colback=gray!10,colframe=gray!50,title=\textbf{AMC Problem Statement}]
Cities $A$ and $B$ are $45$ miles apart. Alicia lives in $A$ and Beth lives in $B$. Alicia bikes towards $B$ at 18 miles per hour. Leaving at the same time, Beth bikes toward $A$ at 12 miles per hour. How many miles from City $A$ will they be when they meet?

\textbf{Correct Solution:} Combined speed = 18 + 12 = 30 mph; Time to meet = 45/30 = 1.5 hours; Distance from A = 18 × 1.5 = 27 miles

\textbf{Ground Truth:} 27
\end{tcolorbox}

\subsubsection{Pre-Training Response: Fundamental Conceptual Confusion}

\begin{tcolorbox}[enhanced,attach boxed title to top center={yshift=-3mm,yshifttext=-1mm},
colback=red!5!white,colframe=red!75!black,colbacktitle=red!80!black,
coltitle=white,title=\textbf{Base Model: Conceptual Breakdown}]

To solve this problem, we can use the concept of relative motion. Here's a step-by-step breakdown:

\begin{itemize}
\item Let's denote the time that Alicia travels as $t$ hours. Since Beth travels in the opposite direction, we can say that the time she travels is also $t$ hours.

\item Since they are traveling towards each other, the distance between them will decrease at a rate of $18 + 12 = 30$ miles per hour.

\item The initial distance between them is 45 miles. After $t$ hours, the total distance traveled by both Alicia and Beth will be the initial distance plus the distance traveled together, which is $45 + 30t$ miles.

\item We can set up an equation using the fact that their combined distance traveled equals the initial distance between them: $45 + 30t = 45$. This is because they meet when the sum of their distances is equal to the initial distance.

\item Solving for $t$, we get $t = 0$ hours. However, this doesn't make sense because it means they meet immediately after starting. Let's try again.

\item Since we can't set up a simple equation with this method, let's think about it in terms of their relative speed: 12
\end{itemize}

[Response cuts off mid-sentence]

\textbf{Parsed Answer:} null
\end{tcolorbox}

\textbf{Analysis: Multiple failure modes.}
The pre-training response reveals several critical reasoning failures: incorrect problem setup, misapplication of distance-time relationships, and ultimately an illogical result. The model exhausts its token budget exploring unproductive solution paths before reaching a conclusion.

\subsubsection{Post-Training Response: Correct Systematic Solution}

\begin{tcolorbox}[enhanced,attach boxed title to top center={yshift=-3mm,yshifttext=-1mm},
colback=green!5!white,colframe=green!75!black,colbacktitle=green!80!black,
coltitle=white,title=\textbf{Post-Training: Correct and Concise}]

To solve this problem, we can follow these steps:

\begin{itemize}
\item We know the distance between Cities A and B is 45 miles.

\item Alicia and Beth are biking towards each other at a combined speed of 18 + 12 = 30 miles per hour.

\item Since they are biking towards each other, we can use the concept of relative speed to find the time it takes for them to meet.

\item We can calculate the time using the formula: time = distance / speed.
  Using this formula, the time they take to meet is 45 miles / 30 miles per hour, which equals 1.5 hours.

\item Now that we have the time, we can find the distance covered by Alicia, since we know her speed is 18 miles per hour.
  Distance covered by Alicia is 18 × 1.5, which equals 27 miles.
\end{itemize}

Therefore, Alicia and Beth will meet 27 miles from City A.

$\boxed{27}$
\end{tcolorbox}

\textbf{Analysis: Systematic improvement.}
Post-training produces a clear, systematic approach: correct conceptual application, logical step progression, and efficient completion within the token limit. This demonstrates improved reasoning quality and token efficiency achieved through preference learning without explicit format rewards.

\subsection{Case Study 2: Multi-Agent Mathematical Reasoning}
\label{app:multi-agent-case}

\subsubsection{Problem: Road Trip Gas Mileage Calculation}

\begin{tcolorbox}[colback=gray!10,colframe=gray!50,title=\textbf{Problem Statement}]
Sophia is thinking of taking a road trip in her car, and would like to know how far she can drive on a single tank of gas. She has traveled 100 miles since last filling her tank, and she needed to put in 4 gallons of gas to fill it up again. The owner's manual for her car says that her tank holds 12 gallons of gas. How many miles can Sophia drive on a single tank of gas?

\textbf{Correct Solution:} 100 ÷ 4 = 25 mpg; 25 × 12 = 300 miles

\textbf{Ground Truth:} 300
\end{tcolorbox}

\subsection{Complete Multi-Agent Transcripts: Pre-Training Failure (Iteration 0)}
\label{app:multi-agent-pre-training}

\subsubsection{Round 1: Initial Agent Responses}

\begin{tcolorbox}[enhanced,attach boxed title to top center={yshift=-3mm,yshifttext=-1mm},
colback=blue!5!white,colframe=blue!75!black,colbacktitle=blue!80!black,
coltitle=white,title=\textbf{User Prompt}]

Solve the following math problem. Sophia is thinking of taking a road trip in her car, and would like to know how far she can drive on a single tank of gas. She has traveled $100$ miles since last filling her tank, and she needed to put in $4$ gallons of gas to fill it up again. The owner's manual for her car says that her tank holds $12$ gallons of gas. How many miles can Sophia drive on a single tank of gas?

Provide a bullet point summary of your step-by-step reasoning.
Your final answer should be a single numerical number, in the form \textbackslash boxed\{answer\}, at the end of your response.
\end{tcolorbox}

\begin{tcolorbox}[enhanced,attach boxed title to top center={yshift=-3mm,yshifttext=-1mm},
colback=red!5!white,colframe=red!75!black,colbacktitle=red!80!black,
coltitle=white,title=\textbf{Agent 1: Mathematical Error}]

To solve this problem, I need to determine Sophia's car's fuel efficiency and then calculate how far she can drive on a full tank.

Step 1: Determine fuel efficiency from the given information.
\begin{itemize}
\item Sophia traveled 100 miles and used 4 gallons of gas
\item Fuel efficiency = 100 miles ÷ 4 gallons = 25 miles per gallon
\end{itemize}

Step 5: Convert the fuel efficiency from mpg to gallons per mile (GPM):
\begin{itemize}
\item Since 1 mpg means traveling 1 mile with each gallon, Sophia will use 4 gallons × (1 mile/mpg) = 4 miles per tank.
\end{itemize}

Step 6: Calculate total distance for full tank:
\begin{itemize}
\item Total distance = 4 miles per tank × 12 gallons = 48 miles
\end{itemize}

$\boxed{48}$
\end{tcolorbox}

\begin{tcolorbox}[enhanced,attach boxed title to top center={yshift=-3mm,yshifttext=-1mm},
colback=orange!5!white,colframe=orange!75!black,colbacktitle=orange!80!black,
coltitle=white,title=\textbf{Agent 2: Incomplete Reasoning}]

If the distance covered in one tank of fuel is 100 miles, then the Miles Per Gallon (MPG) is calculated as follows...

[Response appears incomplete and truncated mid-calculation]

After reviewing the problem again, I need to be more careful with my calculations.

Let me reconsider: Sophia used 4 gallons to travel some distance, and her tank holds 12 gallons total.

I'm getting confused by the different approaches. Let me try again: 100 miles with 4 gallons means...

Final answer: $\boxed{48}$
\end{tcolorbox}

\begin{tcolorbox}[enhanced,attach boxed title to top center={yshift=-3mm,yshifttext=-1mm},
colback=green!5!white,colframe=green!75!black,colbacktitle=green!80!black,
coltitle=white,title=\textbf{Agent 3: Correct but Overruled}]

Let me solve this step by step:

Step 1: Calculate fuel efficiency from given data:
\begin{itemize}
\item Distance traveled: 100 miles
\item Gas used: 4 gallons  
\item Fuel efficiency: 100 ÷ 4 = 25 miles per gallon
\end{itemize}

Step 2: Calculate range on full tank:
\begin{itemize}
\item Full tank capacity: 12 gallons
\item Total range: 25 mpg × 12 gallons = 300 miles
\end{itemize}

Therefore, Sophia can drive 300 miles on a single tank of gas.

$\boxed{300}$
\end{tcolorbox}

\subsubsection{Round 2: Failed Collaborative Discussion}

\begin{tcolorbox}[enhanced,attach boxed title to top center={yshift=-3mm,yshifttext=-1mm},
colback=red!5!white,colframe=red!75!black,colbacktitle=red!80!black,
coltitle=white,title=\textbf{Agent 1: Doubling Down on Error}]

After reviewing all approaches, I maintain my calculation:

Based on the calculations, I'm confident that 48 miles is correct because we need to convert the efficiency properly.

$\boxed{48}$
\end{tcolorbox}

\textbf{Pre-Training Result:} Consensus selected 48 miles despite Agent 3's correct reasoning.

\subsubsection{Complete Multi-Agent Transcripts: Post-Training Success (Iteration 1)}
\label{app:multi-agent-post-training}

\subsubsection{Round 1: Perfect Agent Responses}

\begin{tcolorbox}[enhanced,attach boxed title to top center={yshift=-3mm,yshifttext=-1mm},
colback=green!5!white,colframe=green!75!black,colbacktitle=green!80!black,
coltitle=white,title=\textbf{Agent 1: Complete Correction}]

To solve this problem, I need to determine Sophia's fuel efficiency and calculate her range on a full tank.

Step 1: Calculate Sophia's car's fuel efficiency:
\begin{itemize}
\item Sophia traveled 100 miles since last fill-up
\item She used 4 gallons of gas
\item Fuel efficiency = 100 miles ÷ 4 gallons = 25 miles per gallon
\end{itemize}

Step 2: Calculate range on full tank:
\begin{itemize}
\item Tank capacity: 12 gallons
\item Range = 25 mpg × 12 gallons = 300 miles
\end{itemize}

Therefore, Sophia can drive 300 miles on a single tank of gas.

$\boxed{300}$
\end{tcolorbox}

\begin{tcolorbox}[enhanced,attach boxed title to top center={yshift=-3mm,yshifttext=-1mm},
colback=green!5!white,colframe=green!75!black,colbacktitle=green!80!black,
coltitle=white,title=\textbf{Unanimous Agreement}]

\textbf{Agent 1:} ``I see all agents used the same systematic approach. The calculation is clear: 25 mpg × 12 gallons = 300 miles.''

\textbf{Agent 2:} ``Perfect agreement on methodology. The fuel efficiency calculation is the key step, then applying it to full tank capacity.''

\textbf{Agent 3:} ``Unanimous consensus on 300 miles. All reasoning aligns perfectly.''
\end{tcolorbox}

\textbf{Post-Training Result:} Perfect consensus on 300 miles with all agents showing correct reasoning.

\subsubsection{Analysis: Training Effects on Multi-Agent Collaboration}
\label{app:analysis}
Post-training produces substantial improvements in both individual and collective reasoning. Before training, agents make calculation errors and the consensus mechanism selects incorrect answers even when correct solutions are present, with two agents converging on the same mistake and overruling the correct agent. After training, all agents apply correct methodology and reach unanimous agreement on the right answer. These results demonstrate that preference learning encourages systematic step-by-step reasoning, reducing errors and their propagation during collaboration, while multi-agent training enhances the consensus mechanism's ability to reliably select correct solutions.

\section{Grounding Behaviors in Base-Model Multi-Agent Debates}
\label{app:grounding_examples}

Having provided examples of the impact of post-training on single- and multi-agent reasoning, we now highlight representative examples from some initial debates among base models demonstrating spontaneous grounding behaviors that emerge during multi-agent reasoning. Note that each debate consists of an initial round (three agents provide independent solutions) and a final round (agents update after seeing peers' responses).

\subsection{Example 1: Airport Timing Problem}

\begin{tcolorbox}[enhanced,attach boxed title to top center={yshift=-3mm,yshifttext=-1mm},
colback=blue!5!white,colframe=blue!75!black,colbacktitle=blue!80!black,
coltitle=white,title=\textbf{Problem}]

Amber and Josh are flying to France for vacation. Their flight leaves at 8:00 pm and as an international flight, it's recommended that they check in 2 hours early. It will take 45 minutes to drive to the airport and another 15 minutes to park their vehicle and make their way to the terminal. What is the latest time they can leave their house to make it to the airport in time?
\end{tcolorbox}

The correct solution requires working backwards from the 8:00 PM flight time: they must arrive by 6:00 PM (2 hours early), and with 1 hour total travel time, they must leave by 5:00 PM.

\textbf{Initial round:}

\begin{tcolorbox}[enhanced,attach boxed title to top center={yshift=-3mm,yshifttext=-1mm},
colback=red!5!white,colframe=red!75!black,colbacktitle=red!80!black,
coltitle=white,title=\textbf{Agent 1: Directional Error}]

\begin{itemize}
\item Flight leaves at 8:00 pm, need to check in 2 hours early = 6:00 pm arrival
\item Calculate travel time: 45 + 15 = 60 minutes = 1 hour
\item Add travel time to departure time: 6:00 pm + 1 hour = 7:00 pm
\end{itemize}
\textcolor{red}{Error: Adds travel time instead of subtracting it}\\
Answer: 7:00 pm
\end{tcolorbox}

\begin{tcolorbox}[enhanced,attach boxed title to top center={yshift=-3mm,yshifttext=-1mm},
colback=red!5!white,colframe=red!75!black,colbacktitle=red!80!black,
coltitle=white,title=\textbf{Agent 2: Fundamental Misunderstanding}]

\begin{itemize}
\item Find total travel time: 45 + 15 = 60 minutes = 1 hour
\item Add travel time to flight departure time: 8:00 pm + 1 hour = 9:00 pm
\end{itemize}
\textcolor{red}{Completely ignores 2-hour check-in requirement and adds travel time instead of subtracting it}\\
Answer: 9:00 pm
\end{tcolorbox}

\begin{tcolorbox}[enhanced,attach boxed title to top center={yshift=-3mm,yshifttext=-1mm},
colback=green!5!white,colframe=green!75!black,colbacktitle=green!80!black,
coltitle=white,title=\textbf{Agent 3: Correct Solution}]

\begin{itemize}
\item Flight at 8:00 pm requires arrival by 6:00 pm (2 hours early)
\item Travel time: 45 min driving + 15 min parking = 60 minutes
\item Must leave by: 6:00 pm - 1 hour = 5:00 pm
\end{itemize}
Answer: 5:00 pm
\end{tcolorbox}

\textbf{Final round (after seeing peers):}

\begin{tcolorbox}[enhanced,attach boxed title to top center={yshift=-3mm,yshifttext=-1mm},
colback=red!5!white,colframe=red!75!black,colbacktitle=red!80!black,
coltitle=white,title=\textbf{Agent 1: Adopts Worse Error}]

``After reviewing other solutions, I update my approach:
\begin{itemize}
\item Departure time from house: 8:00 pm
\item Total travel time: 45 + 15 = 60 minutes = 1 hour  
\item Add travel time to departure: 8:00 pm + 1 hour = 9:00 pm''
\end{itemize}
\textcolor{red}{Abandons partial understanding to adopt Agent 2's incorrect reasoning as well}\\
Answer: 9:00 pm
\end{tcolorbox}

\begin{tcolorbox}[enhanced,attach boxed title to top center={yshift=-3mm,yshifttext=-1mm},
colback=red!5!white,colframe=red!75!black,colbacktitle=red!80!black,
coltitle=white,title=\textbf{Agent 2: Persists with Error}]

Maintains original incorrect answer of 9:00 pm
\end{tcolorbox}

\begin{tcolorbox}[enhanced,attach boxed title to top center={yshift=-3mm,yshifttext=-1mm},
colback=green!5!white,colframe=green!75!black,colbacktitle=green!80!black,
coltitle=white,title=\textbf{Agent 3: Explicitly Identifies Error Pattern}]

``The other agents seem to be adding time to the flight departure rather than working backwards from the required arrival time. The correct approach:
\begin{itemize}
\item Required arrival: 6:00 pm (2 hours before 8:00 pm flight)
\item Subtract travel time: 6:00 pm - 1 hour = 5:00 pm departure''
\end{itemize}
\textcolor{green}{Recognizes error in other agents' responses and upholds its own correct understanding}\\
Answer: 5:00 pm
\end{tcolorbox}

\textbf{Key observation.}
Agent 1 had partial understanding (recognized the 6:00 PM arrival requirement) but abandoned this to adopt Agent 2's more fundamental error. Agent 3 explicitly identifies the common error pattern of ``adding time to flight departure'' and maintains the correct backward calculation. This incidentally reflects exactly the case where statistical aggregation can be counterproductive: the inconsistencies in reasoning among the sampled trajectories outnumber the sensible agent, and this error gets propagated. Interestingly, we show that post-training is able to correct these cases despite the initial debate datasets containing examples of bad grounding.

\subsection{Example 2: Self-Correction in Mathematical Reasoning}

\begin{tcolorbox}[enhanced,attach boxed title to top center={yshift=-3mm,yshifttext=-1mm},
colback=blue!5!white,colframe=blue!75!black,colbacktitle=blue!80!black,
coltitle=white,title=\textbf{Problem}]

A store offers a 20\% discount on all items. If Sarah buys a jacket originally priced at \$80 and pants originally priced at \$50, how much does she pay in total after the discount?
\end{tcolorbox}

The correct solution: Total original price = \$130, discount = 20\% of \$130 = \$26, final price = \$104.

\textbf{Initial round:}

\begin{tcolorbox}[enhanced,attach boxed title to top center={yshift=-3mm,yshifttext=-1mm},
colback=orange!5!white,colframe=orange!75!black,colbacktitle=orange!80!black,
coltitle=white,title=\textbf{Agent 1: Calculation Error}]

Original total: \$80 + \$50 = \$130\\
Discount: 20\% = 0.2 × \$130 = \$26\\
Final price: \$130 + \$26 = \$156\\
\textcolor{orange}{Error: Adds discount instead of subtracting}\\
Answer: \$156
\end{tcolorbox}

\begin{tcolorbox}[enhanced,attach boxed title to top center={yshift=-3mm,yshifttext=-1mm},
colback=green!5!white,colframe=green!75!black,colbacktitle=green!80!black,
coltitle=white,title=\textbf{Agent 2: Correct Approach}]

Total: \$130\\
After 20\% discount: \$130 × 0.8 = \$104\\
Answer: \$104
\end{tcolorbox}

\begin{tcolorbox}[enhanced,attach boxed title to top center={yshift=-3mm,yshifttext=-1mm},
colback=red!5!white,colframe=red!75!black,colbacktitle=red!80!black,
coltitle=white,title=\textbf{Agent 3: Misapplies Discount}]

Jacket discount: 20\% of \$80 = \$16, price = \$64\\
Pants discount: 20\% of \$50 = \$10\\
Total: \$64 + \$10 = \$74\\
\textcolor{red}{Error: Adds discounted jacket price to pants discount amount}\\
Answer: \$74
\end{tcolorbox}

\textbf{Final round:}

\begin{tcolorbox}[enhanced,attach boxed title to top center={yshift=-3mm,yshifttext=-1mm},
colback=green!5!white,colframe=green!75!black,colbacktitle=green!80!black,
coltitle=white,title=\textbf{Agent 1: Self-Correction}]

``I realize I added the discount instead of subtracting. Looking at Agent 2's approach of multiplying by 0.8 is cleaner:
\$130 × 0.8 = \$104"\\
\textcolor{green}{Recognizes own error and adopts correct method}\\
Answer: \$104
\end{tcolorbox}

\begin{tcolorbox}[enhanced,attach boxed title to top center={yshift=-3mm,yshifttext=-1mm},
colback=green!5!white,colframe=green!75!black,colbacktitle=green!80!black,
coltitle=white,title=\textbf{Agent 3: Fixes Calculation}]

``Jacket after discount: \$80 × 0.8 = \$64\\
Pants after discount: \$50 × 0.8 = \$40\\
Total: \$64 + \$40 = \$104"\\
Answer: \$104
\end{tcolorbox}

\textbf{Key observation.} Agent 1 explicitly acknowledges their error (``I added the discount instead of subtracting") and adopts Agent 2's more efficient method. Agent 3 corrects their calculation error after seeing consistent peer answers.

\subsection{Example 3: Geometric Problem - Method Simplification}

\begin{tcolorbox}[enhanced,attach boxed title to top center={yshift=-3mm,yshifttext=-1mm},
colback=blue!5!white,colframe=blue!75!black,colbacktitle=blue!80!black,
coltitle=white,title=\textbf{Problem}]

A rectangle has a perimeter of 24 units and an area of 32 square units. What are the dimensions of the rectangle?
\end{tcolorbox}

The correct solution: If dimensions are length $l$ and width $w$, then $2(l+w) = 24$ so $l+w = 12$, and $lw = 32$. Solving this system gives dimensions 8 and 4.

\textbf{Initial round:}

\begin{tcolorbox}[enhanced,attach boxed title to top center={yshift=-3mm,yshifttext=-1mm},
colback=orange!5!white,colframe=orange!75!black,colbacktitle=orange!80!black,
coltitle=white,title=\textbf{Agent 1: Algebraic Approach}]

Let length = $l$, width = $w$\\
From perimeter: $2l + 2w = 24$, so $l + w = 12$\\
From area: $lw = 32$\\
Setting up quadratic: $w(12-w) = 32$\\
$12w - w^2 = 32$\\
$w^2 - 12w + 32 = 0$\\
\textcolor{orange}{Stops without solving the quadratic}
\end{tcolorbox}

\begin{tcolorbox}[enhanced,attach boxed title to top center={yshift=-3mm,yshifttext=-1mm},
colback=green!5!white,colframe=green!75!black,colbacktitle=green!80!black,
coltitle=white,title=\textbf{Agent 2: Direct Factoring}]

$l + w = 12$ and $lw = 32$\\
Looking for factors of 32 that sum to 12: 8 and 4\\
Dimensions: 8 × 4\\
Answer: 8 units by 4 units
\end{tcolorbox}

\begin{tcolorbox}[enhanced,attach boxed title to top center={yshift=-3mm,yshifttext=-1mm},
colback=green!5!white,colframe=green!75!black,colbacktitle=green!80!black,
coltitle=white,title=\textbf{Agent 3: Quadratic Solution}]

From $w^2 - 12w + 32 = 0$\\
Using quadratic formula or factoring: $(w-8)(w-4) = 0$\\
So $w = 8$ or $w = 4$\\
Dimensions: 8 × 4\\
Answer: 8 units by 4 units
\end{tcolorbox}

\textbf{Final round:}

\begin{tcolorbox}[enhanced,attach boxed title to top center={yshift=-3mm,yshifttext=-1mm},
colback=green!5!white,colframe=green!75!black,colbacktitle=green!80!black,
coltitle=white,title=\textbf{Agent 1: Adopts Simpler Method}]

``Agent 2's approach of finding factors of 32 that sum to 12 is much more direct than solving the quadratic. Since 8 × 4 = 32 and 8 + 4 = 12, the dimensions are 8 by 4."\\
\textcolor{green}{Recognizes and adopts more efficient solution method}\\
Answer: 8 units by 4 units
\end{tcolorbox}

\textbf{Key observation.} Agent 1 explicitly acknowledges that Agent 2's factor-finding approach is ``much more direct" than the algebraic method, demonstrating recognition of solution efficiency. Post-training reinforces these behaviors, enabling the model to learn patterns of concise and efficient reasoning.

\subsection{Summary}

These examples demonstrate observable grounding behaviors in multi-agent debates:

\textbf{Error propagation:} Agents sometimes abandon partially correct reasoning to adopt peer errors (Example 1, Agent 1).

\textbf{Explicit error recognition:} Agents identify specific errors in their own reasoning (``I added the discount instead of subtracting") and in peers' approaches (``adding time to flight departure").

\textbf{Method adoption:} Agents recognize more efficient solution methods from peers and explicitly acknowledge this (``Agent 2's approach... is much more direct").

\textbf{Self-correction through peer comparison:} Seeing consistent peer answers prompts agents to revisit and correct their calculations.

These behaviors generate training signals beyond simple majority voting. The debate format captures not just correct answers but also common error patterns, self-correction mechanisms, and explicit reasoning about solution quality, providing rich supervision for teaching models to recognize and adopt sound reasoning strategies.

\section{MACA Algorithm: Additional Implementation Details}
\label{app:maca}
\begin{algorithm}[t]
\caption{Multi-Agent Consensus Alignment Via Debate-RL}\label{alg:maca}
\begin{algorithmic}[1]\footnotesize
  \State \textbf{input:} agents $\{\pi_{\theta_m}\}_{m=1}^M$, rounds $R$, iters $L$, batch size $B$, method in MV-SFT, MV-GRPO, MV-DPO, or MV-KTO
  \For{$\ell = 1, \ldots, L$} 
    \State sample batch $\{ x^{(\ell,i)} \}_{i=1}^B$ of prompts from $\mathcal{D}$
    \For{$i = 1, \ldots, B$}  \Comment{Generate debate trajectories}
      \State $y^{(\ell,i)}_{m,1} \sim \pi_{\theta_m}(\cdot \mid x^{(\ell,i)}) \quad \forall m$
      \For{$r = 2, \ldots, R$}  \Comment{Condition on peers' previous round outputs}
        \State $x^{(\ell,i)}_{m,r} = [x^{(\ell,i)}; \{ y^{(\ell,i)}_{j,r-1} \}_{j \neq m}]$
        \State $y^{(\ell,i)}_{m,r} \sim \pi_{\theta_m}(\cdot \mid x^{(\ell,i)}_{m,r}) \quad \forall m$
      \EndFor
      \State Parse final answers $a^{(\ell,i)}_m = \mathcal{A}( y^{(\ell,i)}_{m,R} ) \quad \forall m$
      \State Aggregate consensus $\hat{a}(x^{(\ell,i)}) = \text{Majority}\{ a^{(\ell,i)}_1, \ldots, a^{(\ell,i)}_M \}$
      \State Extract final contexts $x^{(\ell,i)}_m \gets x^{(\ell,i)}_{m,R}$ for all $m$
      \State Partition responses by consensus alignment:
      \Statex \quad \quad \quad \quad $\mathcal{G}_+^{(\ell,i)} = \{ y^{(\ell,i)}_{m,R} : a^{(\ell,i)}_m = \hat{a}(x^{(\ell,i)}) \}$
      \Statex \quad \quad \quad \quad $\mathcal{G}_-^{(\ell,i)} = \{ y^{(\ell,i)}_{m,R} : a^{(\ell,i)}_m \neq \hat{a}(x^{(\ell,i)}) \}$
    \EndFor
    \State \textbf{build dataset:} $\mathcal{D}_{\text{post}} = \{(x^{(\ell,i)}, \hat{a}(x^{(\ell,i)}), \mathcal{G}_+^{(\ell,i)}, \mathcal{G}_-^{(\ell,i)})\}_{i=1}^B$
    \State \textbf{training data by method:}
    \Statex \quad \quad \quad \quad 
    \textit{MV-SFT:} for each $m$, use $\{ (x^{(\ell,i)}_m, y^{(\ell,i)}_{m,R}) : y^{(\ell,i)}_{m,R} \in \mathcal{G}_+^{(\ell,i)} \}_i$.
    \Statex \quad \quad \quad \quad 
    \textit{MV-GRPO:} for each $m$, store $\{ (x^{(\ell,i)}_m, \hat{a}(x^{(\ell,i)})) \}_i$ for reward computation on new samples.
    \Statex \quad \quad \quad \quad  \textit{MV-DPO/MV-KTO:} pool $\{ (x^{(\ell,i)}_m, y^{(\ell,i)}_{m,R}) \}_{i,m}$ partitioned by $\mathcal{G}^+, \mathcal{G}^-$ across agents.
    \State \textbf{update policies:} apply majority-vote objective to refine each $\pi_{\theta_m}$
  \EndFor
\end{algorithmic}
\end{algorithm}

\subsection{Agent Heterogeneity and Specialization}
In our current experiments, we use homogeneous agents (clones) that update independently over a single training iteration, enabling evaluation of both individual and averaged performance to measure the benefits of divergence. We also explore the impact of training over multiple such iterations of debate and post-training in App.~\ref{app:iter}.
For multi-iteration runs, agent checkpoints can be managed in two ways: agents may either remain synchronized by resetting checkpoints to the best performing one after every iteration, or diverge by updating independently from distinct trajectory data, allowing specialization and diversity through differential learning.
Additionally, the \MethodName\ framework also supports heterogeneous agents, allowing different language models to be independently optimized through this framework, where agents represent distinct models or architectures from the outset, each bringing inherently different capabilities or perspectives.

\section{Experimental Setup}
\label{app:expsetup}

\subsection{Training Parameters}
We conducted hyperparameter sweeps across learning rates from 1e-7 to 1e-5, finding that 1e-5 consistently performed best across all methods. For preference-based methods (MV-KTO, MV-DPO), we used $\beta=0.1$ throughout all experiments. LoRA ranks and alpha values were adjusted based on model size and computational constraints, with rank/alpha of 64-128 for MV-DPO and MV-KTO, 32-64 for MV-GRPO, and 128 for MV-SFT. Tables~\ref{tab:training-params}--\ref{tab:model-dataset-config} provide complete parameter specifications.

\begin{table}[h]
\caption{Training parameters.}
\vspace{0.5em}
\centering

\setlength{\tabcolsep}{6pt}
\renewcommand{\arraystretch}{1.1}
\begin{tabular}{lllll}
\toprule
\textbf{Parameter} & \textbf{SFT} & \textbf{GRPO} & \textbf{DPO} & \textbf{KTO} \\
\midrule
Learning rate & 
$1 \times 10^{-5\text{--}7}$ & 
$1 \times 10^{-5\text{--}7}$ & 
$1 \times 10^{-5\text{--}7}$ & 
$1 \times 10^{-5\text{--}7}$ \\
Weight decay & $1 \times 10^{-2}$ & $1 \times 10^{-2}$ & $1 \times 10^{-2}$ & $1 \times 10^{-2}$ \\
Batch size & 1--8 & 8 & 1--8 & 1--8 \\
Epochs & 1--3 & 1--3 & 1--3 & 1--3 \\
Gradient accumulation steps & 2--4 & 2--4 & 2--4 & 2--4 \\
Optimizer & AdamW & AdamW & AdamW & AdamW \\
LoRA rank (r) & 8--128 & 8--64 & 8--128 & 8--128 \\
LoRA alpha & 8--128 & 8--64 & 8--128 & 8--128 \\
LoRA dropout & 0.1 & 0.1 & 0.1 & 0.1 \\
LoRA target modules & \multicolumn{4}{c}{q\_proj, k\_proj, v\_proj, o\_proj} \\
Entropy coefficient & - & 0.01 & - & - \\
Beta (regularization) & - & - & 0.1--0.3 & 0.1--0.3 \\
Number of generations per step & - & 8 & - & - \\
Maximum sequence length & \multicolumn{4}{c}{Model-dependent (2048--4096)} \\
Mixed precision & \multicolumn{4}{c}{bf16} \\
Quantization & \multicolumn{4}{c}{4-bit (BitsAndBytesConfig)} \\
\bottomrule
\end{tabular}
\label{tab:training-params}
\end{table}

\begin{table}[h]
\caption{Generation parameters.}
\vspace{0.5em}
\centering

\setlength{\tabcolsep}{6pt}
\renewcommand{\arraystretch}{1.1}
\begin{tabular}{ll}
\toprule
\textbf{Parameter} & \textbf{Value} \\
\midrule
Temperature & 1.0 \\
Top-p sampling & 0.9 \\
Maximum new tokens & 256 \\
Do sample & True \\
Pad token ID & EOS token ID \\
\bottomrule
\end{tabular}
\label{tab:generation-params}
\end{table}

\begin{table}[h]
\caption{Multi-agent debate parameters.}
\vspace{0.5em}
\centering

\setlength{\tabcolsep}{6pt}
\renewcommand{\arraystretch}{1.1}
\begin{tabular}{ll}
\toprule
\textbf{Parameter} & \textbf{Value} \\
\midrule
Number of agents & 3 \\
Debate rounds & 2 \\
Batch debate size & 8--24 \\
Use majority vote & True \\
Include debate context & True \\
Use async debate & True \\
\bottomrule
\end{tabular}
\label{tab:multiagent-params}
\end{table}

\begin{table}[h]
\caption{Model and dataset configuration.}
\vspace{0.5em}
\centering

\setlength{\tabcolsep}{6pt}
\renewcommand{\arraystretch}{1.1}
\begin{tabular}{ll}
\toprule
Parameter & Value \\
\midrule
Base models & Phi-4B, Qwen-2B, Llama-3B \\
Training datasets & GSM8K, MATH, MathQA \\
Test datasets & GSM8K, MATH, MathQA, SVAMP, GPQA, CSQA, AMC \\
Training size & 1500 samples \\
Test size & 500 samples \\
Validation size & 500 samples \\
Use quantization & 4-bit BitsAndBytesConfig \\
Use adapter mode & True (LoRA) \\
Use scheduler & True \\
\bottomrule
\end{tabular}
\label{tab:model-dataset-config}
\end{table}

GSM8K, MATH, and MathQA used 1500/500/500 train/valid/test splits. SVAMP, GPQA, CSQA, and AMC were used only for testing, with 300, 448, 500, and 40 test examples, respectively.

\subsection{Datasets}
\label{app:data}
\noindent
We evaluated model mathematical reasoning using seven publicly available datasets spanning a range of difficulty and subject areas: MATH~\citep{hendrycks2021measuring}, GSM8K~\citep{cobbe2021training}, MathQA~\citep{amini2019mathqa}, SVAMP~\citep{patel2021nlp}, GPQA~\citep{rein2023GPQA}, AMC 23~\cite{knoveleng2023amc23}, and CommonsenseQA/CSQA~\citep{talmor-etal-2019-commonsenseqa}.

\textbf{MATH}~\citep{hendrycks2021measuring}: The MATH dataset contains 12,500 high school mathematics problems from algebra, geometry, combinatorics, and number theory, each requiring multi-step reasoning and precise mathematical solutions.

\textbf{GSM8K}~\citep{cobbe2021training}: GSM8K is composed of 8,500 grade-school-level word problems emphasizing arithmetic and logical reasoning, with step-by-step annotated solutions for each problem.

\textbf{MathQA}~\citep{amini2019mathqa}: MathQA features over 37,000 question-answer pairs based on quantitative reasoning, requiring models to convert natural language problems into mathematical expressions and perform multi-step computation.

\textbf{SVAMP}~\citep{patel2021nlp}: SVAMP consists of carefully reworded arithmetic word problems designed to probe model robustness and prevent reliance on annotation artifacts, focusing on genuine multi-step arithmetic reasoning.

\textbf{GPQA}~\citep{rein2023GPQA}: GPQA provides 448 expert-curated, multiple-choice questions covering graduate-level biology, physics, and chemistry, emphasizing deep, multi-stage reasoning and robust factual understanding.

\textbf{AMC 23}~\citep{knoveleng2023amc23}: This dataset includes recent problems drawn from the 2023 American Mathematics Competitions (AMC), spanning algebra, combinatorics, geometry, and number theory, and is useful for assessing model performance on expert-constructed math tasks

\textbf{CommonsenseQA}~\citep{talmor-etal-2019-commonsenseqa}: CommonsenseQA is a challenging multiple-choice question-answering dataset requiring models to apply commonsense reasoning over diverse everyday scenarios. The dataset comprises 12,247 questions, each designed to probe deeper, non-trivial conceptual knowledge beyond factual recall, making it a benchmark for evaluating commonsense understanding in language models.

These datasets collectively enable a thorough evaluation of analytical reasoning capabilities in large language models.

\subsection{Multi-Agent Debate Infrastructure}
\label{sec:infrastructure}

Our multi-agent debate system addresses the computational challenges of training and deploying multiple agents efficiently through specialized infrastructure for quantized training isolation and scalable debate inference.

\subsubsection{Quantized MARL Training}
Current post-training libraries lack support for multi-gpu distributed training with 4-bit quantization and PEFT adapters. We implement training isolation where each agent trains on a single GPU with explicit resource assignment, using gradient checkpointing and paged optimizers (paged\_adamw\_8bit) for memory efficiency. This enables stable multi-agent RL with QLoRA while preserving quantization benefits.

\subsubsection{Scalable Debate Inference Framework}

The inference system enables efficient multi-agent debates through adapter hot-swapping and dynamic resource management, designed to scale seamlessly with available compute resources. A centralized base model manager maintains a pool of quantized base models shared across agents, implementing lazy loading and caching for lightweight LoRA adapters that enable rapid agent switching without full model reloading. An adapter job scheduler coordinates debate workloads across available GPUs through adaptive load balancing, priority queues for debate tasks ordered by round index, and automatic cluster adaptation that detects available hardware and distributes workloads accordingly. This architecture enables concurrent multi-agent inference on hardware that would typically support only single full-precision models, with debate throughput scaling linearly with compute resources while maintaining memory efficiency through quantization and adapter sharing.

\subsection{Debate Prompt Templates}
\label{app:prompts}

This appendix details the prompts used in our multi-agent debate system.

\subsubsection{Initial Question Prompt}

Used in the first round of debate.

\paragraph{For GPQA, MathQA, CommonsenseQA}
\noindent
\begin{tcolorbox}[colback=gray!5!white, colframe=black, title=Initial Prompt: Multiple Choice]
\begin{lstlisting}[basicstyle=\ttfamily,breaklines=true]
Answer the following multiple choice question as accurately as possible. 
{question}

Provide a bullet point summary of your step-by-step reasoning.
Your final answer should be a single choice letter in the form 
\boxed{answer}, at the end of your response.
\end{lstlisting}
\end{tcolorbox}

\paragraph{For GSM8K, SVAMP, AMC, MATH}
\noindent
\begin{tcolorbox}[colback=gray!5!white, colframe=black, title=Initial Prompt: Math Problem]
\begin{lstlisting}[basicstyle=\ttfamily,breaklines=true]
Solve the following math problem. {question}

Provide a bullet point summary of your step-by-step reasoning.
Your final answer should be a single numerical number, in the form 
\boxed{answer}, at the end of your response.
\end{lstlisting}
\end{tcolorbox}

\subsubsection{Debate Phase Prompt}

Used in subsequent rounds of debate to incorporate peer context.

\noindent
\begin{tcolorbox}[colback=gray!5!white, colframe=black, title=Debate Phase Prompt]
\begin{lstlisting}[basicstyle=\ttfamily,breaklines=true]
Here are solutions from other agents: 
{agent_responses}

Using each response as additional advice, give an updated bullet point summary of your step-by-step reasoning to the question: {question}

Make sure your final answer is in the form \boxed{answer}, at the end of your response.
\end{lstlisting}
\end{tcolorbox}

\section{Ablation: Ground Truth vs Debate Majority Vote}
\label{app:gtmvcomp}

Tables~\ref{tab:single_agent_gt_mv} and \ref{tab:multi_agent_gt_mv} report ablations comparing post-training on oracle ground-truth (GT) labels versus (DMV) labels across MATH and GSM8K benchmarks.

In the single-agent setting (Table~\ref{tab:single_agent_gt_mv}), DMV supervision proves consistently competitive with, and often superior to, GT supervision across Qwen2B, Llama-3B, Phi-4B, and Llama-8B. Gains are most pronounced for preference-based objectives such as KTO and DPO, where DMV provides a more stable learning signal. By contrast, the oracle GT signal sometimes produces degraded performance, particularly in KTO. Notably, DMV provides a robust alternative that avoids these pitfalls, often leading to stronger outcomes without requiring external supervision.

\begin{table*}[t]
\caption{Single-agent accuracy: Unsupervised post-training using debate-majority-vote–derived supervision (DMV) performs comparably to or outperforms supervised training using ground truth (GT) labels. Bold indicates the better score in each pair.}
\centering
\footnotesize
\begin{tabular}{ll
                c
                c c
                c c
                c c
                c c}
\toprule
Model & Dataset
& Debate
& \multicolumn{2}{c}{SFT}
& \multicolumn{2}{c}{KTO}
& \multicolumn{2}{c}{DPO}
& \multicolumn{2}{c}{GRPO} \\
\cmidrule(lr){4-5}\cmidrule(lr){6-7}\cmidrule(lr){8-9}\cmidrule(lr){10-11}
& &
  & \textit{GT} & \textit{DMV}
  & \textit{GT} & \textit{DMV}
  & \textit{GT} & \textit{DMV}
  & \textit{GT} & \textit{DMV} \\
\midrule
\multirow{2}{*}{Qwen2B}
  & MATH  & 7.67
           & 12.13 & \bfseries 12.20
           & 12.67 & \bfseries 19.67
           & 17.13 & \bfseries 23.00
           & \bfseries 21.33 & 17.33 \\
  & GSM8K & 23.00
           & 24.20 & \bfseries 24.80
           & 45.40 & \bfseries 47.20
           & \bfseries 46.47 & 44.20
           & \bfseries 39.73 & 32.07 \\
\addlinespace
\multirow{2}{*}{Llama3B}
  & MATH  & 27.87
           & \bfseries 26.80 & 25.60
           & 12.53 & \bfseries 39.20
           & 38.40 & \bfseries 40.67
           & \bfseries 36.80 & 35.00 \\
  & GSM8K & 57.33
           & 54.13 & \bfseries 56.73
           & 27.27 & \bfseries 66.27
           & 56.87 & \bfseries 66.73
           & \bfseries 54.93 & 52.20 \\
\addlinespace
\multirow{2}{*}{Phi4B}
  & MATH  & 34.60
           & 33.07 & \bfseries 35.20
           & \bfseries 33.73 & 33.00
           & \bfseries 37.00 & 36.33
           & \bfseries 37.80 & 37.33 \\
  & GSM8K & 67.27
           & \bfseries 71.67 & 70.20
           & \bfseries 76.13 & 75.47
           & 75.80 & \bfseries 76.60
           & \bfseries 74.80 & 70.40 \\
\addlinespace
\multirow{2}{*}{Llama8B}
  & MATH  & 22.93
           & \bfseries 23.73 & 23.00
           & \bfseries 41.20 & 39.40
           & 45.13 & \bfseries 46.40
           & 29.07 & \bfseries 31.13 \\
  & GSM8K & 57.93
           & 40.20 & \bfseries 41.67
           & \bfseries 72.60 & 70.87
           & 76.33 & \bfseries 77.67
           & 61.27 & \bfseries 66.87 \\
\bottomrule
\end{tabular}
\label{tab:single_agent_gt_mv}
\end{table*}

In the multi-agent setting (Table~\ref{tab:multi_agent_gt_mv}), the advantages of DMV supervision become even clearer. Across nearly all models and methods, DMV either matches or exceeds GT labels. The gains are especially consistent under preference-learning formulations (KTO, DPO), where DMV supervision yields more reliable improvements to consensus-based performance. While GT labels retain competitive strength in certain cases (e.g., Phi-4B on MATH tasks), DMV repeatedly delivers higher or more stable final-round accuracies.

\begin{table*}[t]
\caption{Multi-agent accuracy: Unsupervised post-training using debate-majority-vote–derived supervision (\textit{DMV}) performs comparably to or outperforms supervised training using ground truth (\textit{GT}) labels. Bold indicates the better score in each pair.}
\centering
\footnotesize
\begin{tabular}{ll
                c
                c c
                c c
                c c
                c c}
\toprule
Model & Dataset
& Init
& \multicolumn{2}{c}{SFT}
& \multicolumn{2}{c}{KTO}
& \multicolumn{2}{c}{DPO}
& \multicolumn{2}{c}{GRPO} \\
\cmidrule(lr){4-5}\cmidrule(lr){6-7}\cmidrule(lr){8-9}\cmidrule(lr){10-11}
& &
  & \textit{GT} & \textit{DMV}
  & \textit{GT} & \textit{DMV}
  & \textit{GT} & \textit{DMV}
  & \textit{GT} & \textit{DMV} \\
\midrule
\multirow{2}{*}{Qwen2B}
  & MATH  & 32.4
           & 38.6 & \bfseries 40.6
           & 40.4 & \bfseries 46.2
           & 39.4 & \bfseries 41.2
           & \bfseries 43.6 & 40.2 \\
  & GSM8K & 49.6
           & 50.0 & \bfseries 51.6
           & \bfseries 66.2 & 63.8
           & \bfseries 61.6 & 57.0
           & \bfseries 60.4 & 51.8 \\
\addlinespace
\multirow{2}{*}{Llama3B}
  & MATH  & 37.8
           & \bfseries 35.6 & 33.6
           & 15.8 & \bfseries 51.8
           & 50.8 & \bfseries 51.4
           & 49.2 & \bfseries 50.8 \\
  & GSM8K & 65.6
           & \bfseries 65.8 & 64.0
           & 29.4 & \bfseries 73.8
           & 67.0 & \bfseries 73.8
           & 64.4 & \bfseries 69.6 \\
\addlinespace
\multirow{2}{*}{Phi4B}
  & MATH  & 44.4
           & 44.0 & \bfseries 46.0
           & \bfseries 46.8 & 44.6
           & \bfseries 50.2 & 48.6
           & \bfseries 51.4 & 50.8 \\
  & GSM8K & 79.6
           & \bfseries 79.8 & 78.4
           & \bfseries 82.6 & 81.2
           & 83.8 & \bfseries 84.4
           & 81.6 & \bfseries 83.4 \\
\addlinespace
\multirow{2}{*}{Llama8B}
  & MATH  & 32.8
           & \bfseries 34.0 & 33.4
           & \bfseries 56.0 & 55.8
           & \bfseries 61.8 & 60.8
           & \bfseries 48.6 & 44.8 \\
  & GSM8K & 74.0
           & \bfseries 65.2 & \bfseries 65.2
           & \bfseries 81.2 & 79.4
           & 81.6 & \bfseries 83.0
           & 83.2 & \bfseries 84.2 \\
\bottomrule
\end{tabular}
\label{tab:multi_agent_gt_mv}
\end{table*}

Together, these results show that debate-derived majority-vote supervision provides an effective, scalable alternative to oracle ground truth. DMV not only mitigates the instability observed when GT is used in preference-learning objectives, but also enhances both single- and multi-agent training. These findings underscore the efficacy of unsupervised alignment signals, leveraging a model’s own consensus dynamics, as a robust substitute for human-labeled supervision.

\section{Ablation: Effect of Peer Context in Multi-Agent Debate}
\label{app:context}

Tables~\ref{tab:multi_agent_context_vs_nocontext} and~\ref{tab:iter1_context_ablation_all_three} examine whether conditioning on peer responses during training improves debate performance.

\begin{table*}[t]
\caption{Multi-agent comparison of Context vs No Context runs across Initial Debate, SFT, KTO, DPO, and GRPO. Bold indicates which setting performed better in each pair.}
\centering
\footnotesize
\begin{tabular}{ll
                c
                c c
                c c
                c c
                c c}
\toprule
Model & Dataset
& Debate
& \multicolumn{2}{c}{SFT} 
& \multicolumn{2}{c}{KTO} 
& \multicolumn{2}{c}{DPO} 
& \multicolumn{2}{c}{GRPO} \\
\cmidrule(lr){4-5}\cmidrule(lr){6-7}\cmidrule(lr){8-9}\cmidrule(lr){10-11}
& &
  & \textit{Context} & \textit{NoCtx}
  & \textit{Context} & \textit{NoCtx}
  & \textit{Context} & \textit{NoCtx}
  & \textit{Context} & \textit{NoCtx} \\
\midrule
\multirow{2}{*}{Qwen2B}
  & MATH  & 32.4 
           & \bfseries 40.6 & 38.0  
           & 46.2 & \bfseries 46.6  
           & 41.2 & \bfseries 43.4  
           & \bfseries 40.2 & 39.8 \\
  & GSM8K & 49.6 
           & \bfseries 51.6 & \bfseries 51.6  
           & \bfseries 63.8 & 59.2  
           & 57.0 & \bfseries 57.8  
           & 51.8 & \bfseries 57.2 \\
\addlinespace
\multirow{2}{*}{Llama3B}
  & MATH  & 37.8
           & 33.6 & \bfseries 37.4  
           & \bfseries 51.8 & 27.0  
           & \bfseries 51.4 & 51.0  
           & \bfseries 50.8 & 47.0 \\
  & GSM8K & 65.6  
           & 64.0 & \bfseries 65.4  
           & \bfseries 73.8 & 59.0  
           & \bfseries 73.8 & 70.6  
           & \bfseries 69.6 & 50.0 \\
\addlinespace
\multirow{2}{*}{Phi4B}
  & MATH  & 44.4  
           & \bfseries 46.0 & 45.0  
           & 44.6 & \bfseries 46.0  
           & \bfseries 48.6 & 48.4  
           & 50.8 & \bfseries 52.8 \\
  & GSM8K & 79.6 
           & 78.4 & \bfseries 80.0  
           & 81.2 & \bfseries 82.0  
           & \bfseries 84.4 & 81.8  
           & 83.4 & \bfseries 82.0 \\
\addlinespace
\multirow{2}{*}{Llama8B}
  & MATH  & 32.8
           & \bfseries 33.4 & 31.4  
           & \bfseries 55.8 & 49.2  
           & \bfseries 60.8 & 57.2  
           & \bfseries 44.8 & 44.6 \\
  & GSM8K & 74.0 
           & 65.2 & \bfseries 67.4  
           & 79.4 & \bfseries 80.6  
           & \bfseries 83.0 & 81.6  
           & \bfseries 84.2 & 71.0 \\
\bottomrule
\end{tabular}
\label{tab:multi_agent_context_vs_nocontext}
\end{table*}

\begin{table}[t]
\caption{Post-training with peer context teaches agents to utilize other agents' responses in the debate format for more effective final round ensemble reasoning (MV-DPO).}
\centering
\footnotesize
\begin{tabular}{l c c c c c c}
\toprule
& \multicolumn{2}{c}{Initial Round Avg} &
  \multicolumn{2}{c}{Final Round Avg} &
  \multicolumn{2}{c}{Final Round MV} \\
\cmidrule(lr){2-3}\cmidrule(lr){4-5}\cmidrule(lr){6-7}
Model-Data & NoCtx & Context & NoCtx & Context & NoCtx & Context \\
\midrule
Qwen2B-MATH   & 22.33 & \bfseries 23.00 & 37.13 & \bfseries 38.33 & \bfseries 43.40 & 41.20 \\
Qwen2B-GSM8K  & \bfseries 46.53 & 44.20 & 52.53 & \bfseries 54.73 & \bfseries 57.80 & 57.00 \\
Llama3B-MATH   & 40.27 & \bfseries 40.67 & 41.13 & \bfseries 45.80 & 51.00 & \bfseries 51.40 \\
Llama3B-GSM8K  & 65.87 & \bfseries 66.73 & 67.87 & \bfseries 70.60 & 70.60 & \bfseries 73.80 \\
Phi4B-MATH   & \bfseries 39.47 & 36.33 & \bfseries 39.53 & 34.73 & 48.40 & \bfseries 48.60 \\
Phi4B-GSM8K  & 74.67 & \bfseries 76.60 & 74.60 & \bfseries 77.80 & 81.80 & \bfseries 84.40 \\
Llama8B-MATH   & \bfseries 49.87 & 46.40 & 51.47 & \bfseries 57.80 & 57.20 & \bfseries 60.80 \\
Llama8B-GSM8K  & 75.80 & \bfseries 77.67 & 78.00 & \bfseries 81.20 & 81.60 & \bfseries 83.00 \\
\bottomrule
\end{tabular}
\label{tab:iter1_context_ablation_all_three}
\end{table}

Training with peer context substantially improves multi-agent debate performance (Table~\ref{tab:iter1_context_ablation_all_three}). While initial round (single-agent) performance shows modest benefits from context training (5 out of 8 cases improve), the advantages become pronounced in final-round multi-agent settings. Context training improves final-round individual accuracy in 7 out of 8 cases, with particularly large gains for models like L8B-MATH (57.80 vs. 51.47) and L3B-MATH (45.80 vs. 41.13). Most importantly, final-round majority voting benefits from context in 6 out of 8 cases, with GSM8K tasks showing consistent improvements across all model sizes (1.4-3.2 percentage points) and larger models achieving substantial gains on MATH tasks (e.g., L8B-MATH improving from 57.20 to 60.80). These results demonstrate that context-aware training teaches agents to effectively leverage peer feedback during deliberation, leading to stronger consensus outcomes in multi-agent debate.

\section{Iterative Improvement}
\label{app:iter}

Table~\ref{tab:iter_all_three_c} examines whether iterative training beyond the first iteration yields continued improvements. While It-1 produces the substantial gains reported in our main results, iterations 2 and 3 demonstrate continued modest improvements: in 23 of 24 evaluation settings, either It-2 or It-3 achieves the best performance. For example, Phi-4B on MATH majority vote increases from 55.00 (It-1) to 57.40 (It-3), and Llama-8B on GSM8K improves from 82.80 to 85.60.

These gains show clear diminishing returns compared to the It-0→It-1 jump, with typical improvements of 1-3 percentage points between iterations. Performance occasionally dips between adjacent iterations (e.g., L3B-GSM8K majority vote: 72.00→74.60→70.80) but generally trends upward. This pattern suggests that iterative training continues to extract useful signal from debate-generated data, though with decreasing marginal benefit after the initial iteration.

\begin{table}[t]
\caption{Iterative alternation between debate generation and post-training across four iterations shows substantial initial gains (It-0 to It-1) followed by diminishing returns. Model abbreviations: Q2B=Qwen-2B, L3B=Llama-3B, L8B=Llama-8B, P4B=Phi-4B.}
\centering

\setlength{\tabcolsep}{3.5pt}
\begin{tabular}{l c c c c | c c c c | c c c c}
\toprule
& \multicolumn{4}{c|}{\makecell{Initial Round Avg}} & 
  \multicolumn{4}{c|}{\makecell{Final Round Avg}} &
  \multicolumn{4}{c}{\makecell{Final Round MV}} \\
\cmidrule(lr){2-5} \cmidrule(lr){6-9} \cmidrule(lr){10-13}
Model-Data & It-0 & It-1 & It-2 & It-3
    & It-0 & It-1 & It-2 & It-3
    & It-0 & It-1 & It-2 & It-3 \\
\midrule
Q2B-MATH    & 7.67 & 17.40 & 18.00 & \bfseries 19.33 & 21.47 & 43.13 & 40.73 & \bfseries 43.93 & 32.40 & 47.40 & 41.20 & \bfseries 48.60 \\
Q2B-GSM8K   & 23.00 & 44.20 & 44.67 & \bfseries 44.73 & 35.07 & 58.67 & \bfseries 59.93 & 59.47 & 49.60 & 60.80 & 62.20 & \bfseries 62.60 \\
L3B-MATH    & 27.87 & 40.93 & 39.33 & \bfseries 41.93 & 24.27 & 48.27 & 46.93 & \bfseries 48.73 & 37.80 & \bfseries 55.00 & 54.00 & 53.60 \\
L3B-GSM8K   & 57.33 & 64.60 & \bfseries 67.07 & 65.20 & 49.20 & 68.73 & \bfseries 71.13 & 69.07 & 65.60 & 72.00 & \bfseries 74.60 & 70.80 \\
P4B-MATH    & 34.60 & 43.27 & 43.67 & \bfseries 43.80 & 34.37 & 48.00 & 48.93 & \bfseries 50.20 & 44.40 & 55.00 & 55.80 & \bfseries 57.40 \\
P4B-GSM8K   & 67.27 & 75.73 & \bfseries 76.20 & 75.47 & 68.53 & 77.00 & \bfseries 79.20 & 77.60 & 79.60 & 81.40 & \bfseries 84.40 & 83.00 \\
L8B-MATH    & 22.93 & 44.53 & 44.67 & \bfseries 44.87 & 22.53 & 55.73 & \bfseries 57.60 & 56.07 & 32.80 & 58.20 & \bfseries 60.40 & 59.80 \\
L8B-GSM8K   & 57.93 & 77.00 & 77.87 & \bfseries 78.80 & 56.53 & 80.80 & 80.73 & \bfseries 82.93 & 74.00 & 82.80 & 82.60 & \bfseries 85.60 \\
\bottomrule
\end{tabular}
\label{tab:iter_all_three_c}
\end{table}

\section{DPO Pair Selection Strategy Analysis}
\label{app:dpo_pair_selection}

We compared two strategies for creating preference pairs from multi-agent debate data: majority/minority partitioning versus confidence-based selection using model log-probabilities. Analysis covered 2,226 agent responses from 742 problems across multiple models and datasets. Majority/minority partitioning substantially outperforms confidence-based selection across all metrics. Majority responses achieve 68.0\% accuracy versus 28.1\% for minority responses (39.8 percentage point gap), while high-confidence responses achieve only 51.5\% accuracy versus 33.2\% for low-confidence (18.3 percentage point gap). The majority strategy yields an effect size of Cohen's d = 1.832 compared to 0.281 for confidence-based selection, a 6.5× difference in discriminative power. Additionally, majority voting provides usable preference signals in 70.5\% of examples versus 35.2\% for confidence-based selection. These results validate using majority vote consensus for DPO pair selection, demonstrating that collective agreement provides more reliable quality signals than individual model confidence for mathematical reasoning tasks.

\section{Self-Consistency and Accuracy Correlation Analysis}
\label{app:correlation}

We analyze the correlation between self-consistency and accuracy improvements across three experimental conditions to assess the robustness of our findings under different token generation limits and quantization settings. We test three conditions: (1) Token Capped + Quantized (256 tokens, 4-bit), our standard experimental setup matching computational constraints; (2) Token Uncapped + Non-Quantized (2048 tokens with no observed truncation, full precision), representing maximum generation quality without computational constraints; and (3) Token Uncapped + Quantized (2048 tokens with no observed truncation, 4-bit), a balanced approach removing token truncation while maintaining efficiency.

\begin{table}[h]
\centering
\caption{Self-consistency vs accuracy correlation across experimental conditions. ``Capped'' refers to 256 token limit, which is used throughout the work due to computational constraints in multi-agent RL settings, and ``Uncapped'' refers to 2048 token limit where no response truncation occurred.}
\label{tab:correlation}
\begin{tabular}{lcccccc}
\toprule
Condition & Model-Dataset & \multicolumn{2}{c}{Self-Consistency (\%)} & \multicolumn{2}{c}{Accuracy (\%)} & Correlation \\
& & Base & Post & Base & Post & ($r$) \\
\midrule
\multirow{3}{*}{\parbox{1.5cm}{\centering Capped\\+Quant}} 
& Llama3B-MathQA & 31.2 & \bfseries 49.4 & 29.1 & \bfseries 45.8 & 0.954 \\
& Qwen2B-Math & 12.0 & \bfseries 27.8 & 11.6 & \bfseries 26.8 & 0.989 \\
& Qwen2B-GSM8K & 26.4 & \bfseries 54.0 & 25.9 & \bfseries 52.7 & 0.982 \\
\midrule
\multirow{3}{*}{\parbox{1.5cm}{\centering Uncapped\\+Full}}
& Llama3B-MathQA & 49.5 & \bfseries 68.5 & 45.1 & \bfseries 60.9 & 0.869 \\
& Qwen2B-Math & 50.4 & \bfseries 57.2 & 48.2 & \bfseries 54.4 & 0.944 \\
& Qwen2B-GSM8K & 51.4 & \bfseries 76.4 & 50.8 & \bfseries 75.5 & 0.933 \\
\midrule
\multirow{3}{*}{\parbox{1.5cm}{\centering Uncapped\\+Quant}}
& Llama3B-MathQA & 49.8 & \bfseries 67.3 & 45.8 & \bfseries 60.8 & 0.899 \\
& Qwen2B-Math & 42.8 & \bfseries 49.7 & 38.9 & \bfseries 45.2 & 0.935 \\
& Qwen2B-GSM8K & 50.6 & \bfseries 73.5 & 48.9 & \bfseries 71.3 & 0.927 \\
\bottomrule
\end{tabular}
\end{table}

All conditions show strong positive correlations ($r > 0.86$) between self-consistency and accuracy, validating self-consistency as a robust proxy for model performance across computational settings. Interestingly, capped conditions show slightly higher correlations (mean $r = 0.975$) compared to uncapped conditions (mean $r = 0.915$ for both quantized and full precision). This tighter coupling likely reflects how post-training teaches more efficient reasoning: models learn to better utilize limited token budgets, producing fewer truncated responses while achieving higher accuracy within constraints. In uncapped settings, post-trained models generate 22-36\% shorter responses while still being more accurate, demonstrating that this efficiency persists without token limits. Four-bit quantization minimally impacts correlation strength, with quantized uncapped (mean $r = 0.920$) closely matching full precision uncapped (mean $r = 0.915$), demonstrating that computational efficiency can be achieved without degrading the consistency-accuracy relationship. Post-training consistently improves both metrics across all conditions, with self-consistency gains ranging from 6.9 to 27.6 percentage points and accuracy gains from 6.3 to 24.7 percentage points, confirming that our approach remains effective under varying computational constraints.

\subsection{Self-Consistency Curves}
\label{app:consistency_curves}
\begin{figure}[t]
  \centering
  \begin{subfigure}{0.33\textwidth}
    \includegraphics[trim={20pt 0pt 20pt 0pt},clip,width=\linewidth]{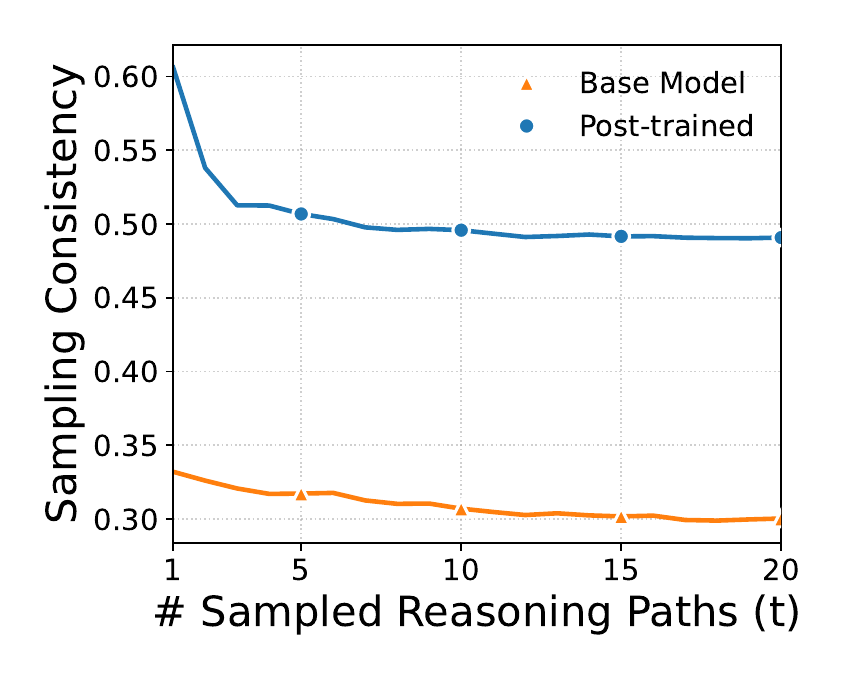}
    \caption{Llama-3B on MATHQA}
  \end{subfigure}\hfill
  \begin{subfigure}{0.33\textwidth}
    \includegraphics[trim={20pt 0pt 20pt 0pt},clip,width=\linewidth]{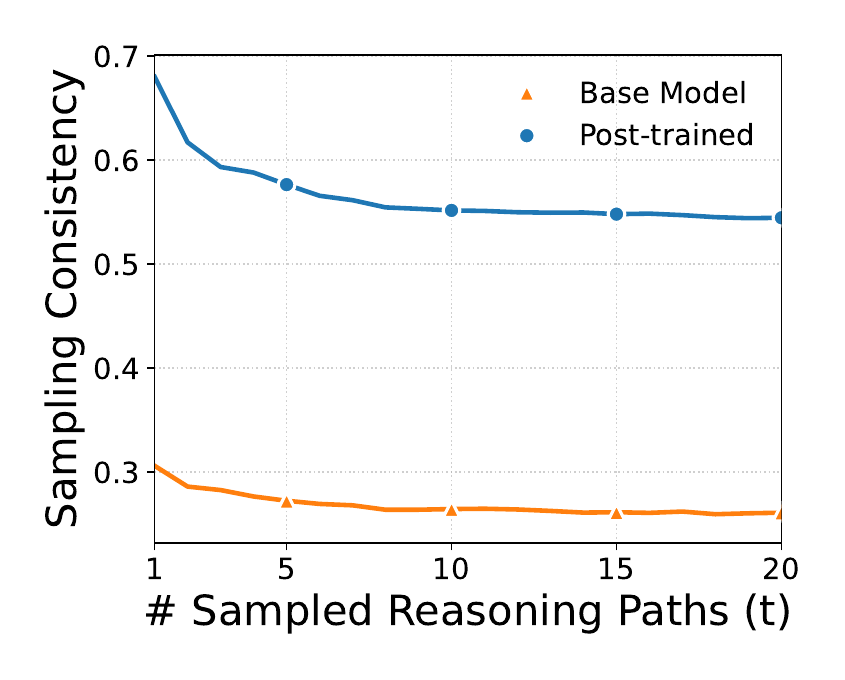}
    \caption{Qwen-2B on GSM8K}
  \end{subfigure}\hfill 
    \begin{subfigure}{0.33\textwidth}
    \includegraphics[trim={20pt 0pt 20pt 0pt},clip,width=\linewidth]{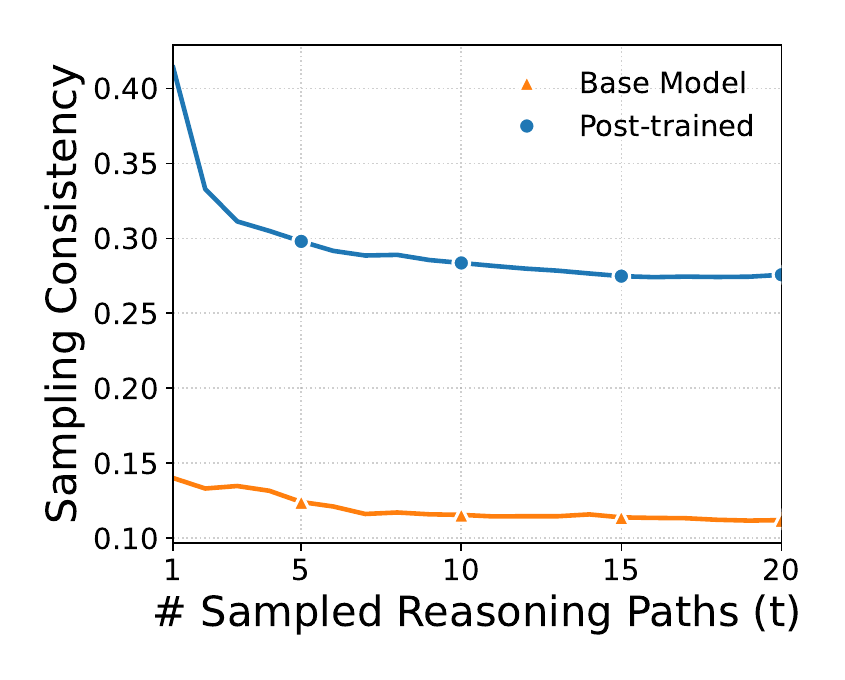}
    \caption{Qwen-2B on MATH}
  \end{subfigure}
  \caption{Consistency before and after \MethodName\ post-training with token constraints (maximum of 256 tokens).
  Pre-trained models \textcolor{orange}{(Orange)} show low sampling consistency across sampled trajectories.
  Post-training with \MethodName\ \textcolor{blue}{(Blue)} substantially improves sampling consistency.
  Averaged over 500 test prompts with 20 trajectories each.}
  \label{fig:consistencysampling}
\end{figure}

Figure~\ref{fig:consistencysampling} shows sampling consistency before and after \MethodName\ post-training. Pre-trained models show low sampling consistency across sampled trajectories, while post-training substantially improves consistency, with curves stabilizing up to 27.6 percentage points above baseline.

\subsection{Self-Consistency Improvements Without Token Constraints}
\label{app:ntcsc}
\begin{figure}[t]
  \centering
  \begin{subfigure}{0.33\textwidth}
    \includegraphics[width=\linewidth]{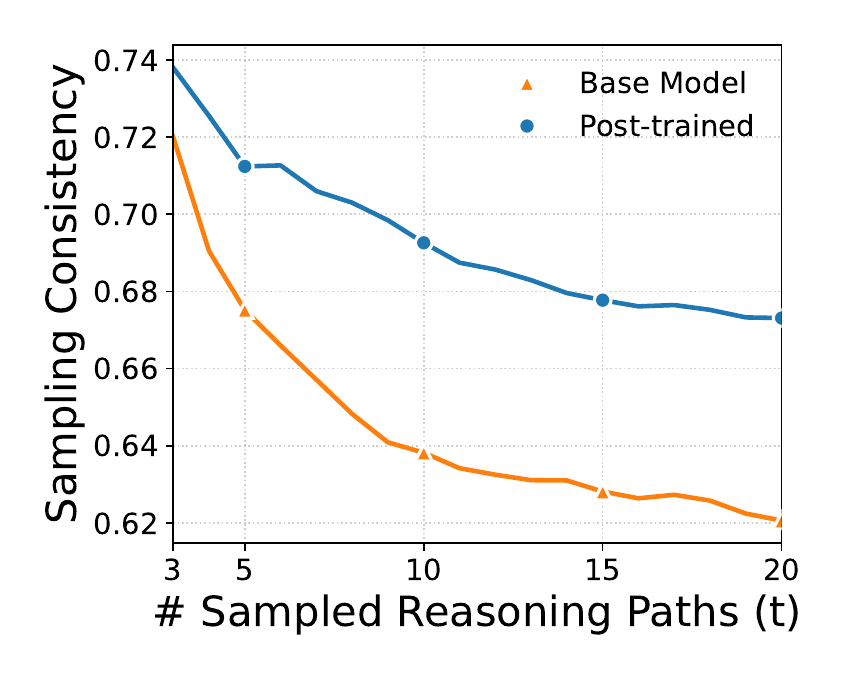}
    \caption{Llama-3B on MathQA}
  \end{subfigure}\hfill
  \begin{subfigure}{0.33\textwidth}
    \includegraphics[width=\linewidth]{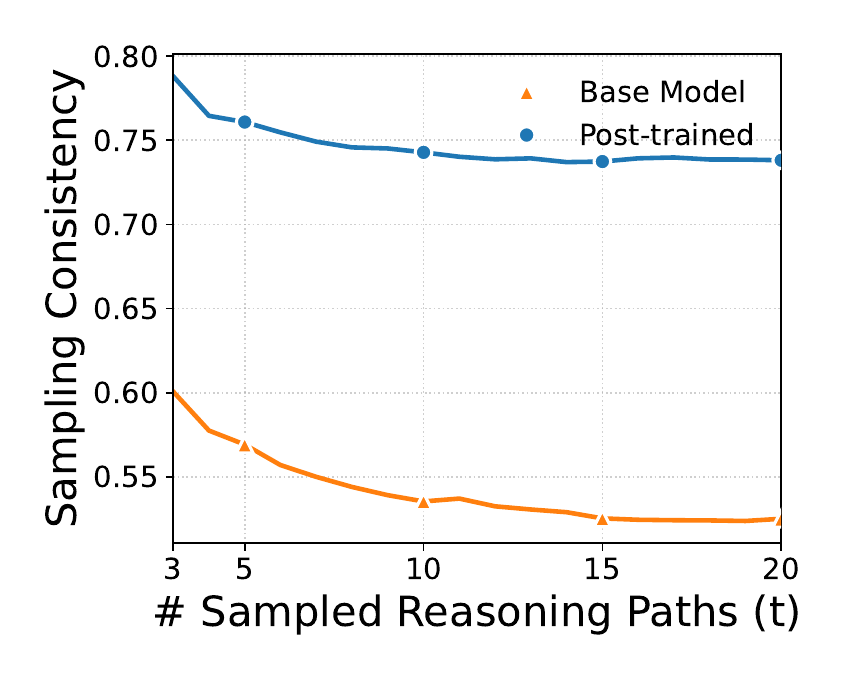}
    \caption{Qwen-2B on GSM8K}
  \end{subfigure}\hfill
  \begin{subfigure}{0.33\textwidth}
    \includegraphics[width=\linewidth]{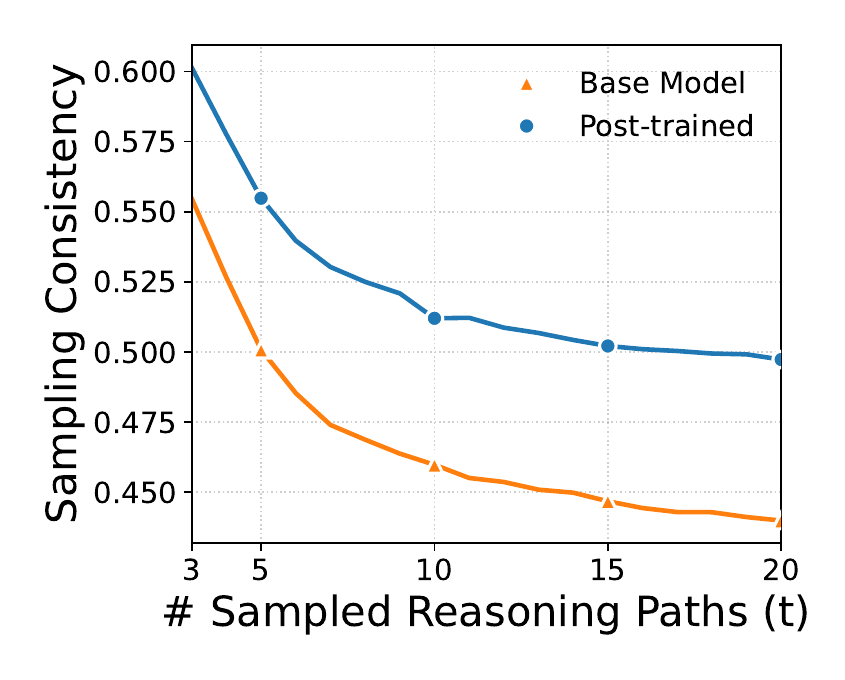}
    \caption{Qwen-2B on MATH}
  \end{subfigure}
  \caption{Consistency before and after \MethodName\ post-training without token constraints. Models trained with 256-token debates still show gains when tested with full-length responses, though with reduced effect sizes due to the weaker training signal compared to testing conditions. Colors: \textcolor{blue}{Blue}: post-trained model, \textcolor{orange}{Orange}: base model.}
  \label{fig:no_token_cap}
\end{figure}

Figure~\ref{fig:no_token_cap} shows self-consistency curves for the aforementioned model-dataset pairs without token constraints. Improvements persist across all configurations, demonstrating that \MethodName's benefits are not merely artifacts of addressing truncation. The effect sizes are slightly smaller than in our main results due to train-test mismatch: these models were trained on debate signals from 256-token responses but tested without constraints. Training on full-length debates would likely yield larger improvements, as the consensus signals would be stronger and better aligned with test conditions.

\section{Impact of \MethodName\ on General Reasoning}
\label{app:macageneralization}

We demonstrate that debate-derived majority vote supervision achieves comparable performance to ground truth supervision while enabling effective generalization to unseen reasoning domains. Tables~\ref{tab:within_distribution_comparison} and ~\ref{tab:cross_domain_generalization} present direct comparisons between these supervision approaches across mathematical training domains and out-of-distribution tasks.

Table~\ref{tab:within_distribution_comparison} shows that debate-derived supervision performs comparably to ground truth labels on training domains, with methods trading wins across model-dataset combinations. Table~\ref{tab:cross_domain_generalization} reveals that both supervision approaches generalize effectively to unseen reasoning tasks, including mathematical word problems (SVAMP), science reasoning (GPQA), and commonsense reasoning (CSQA). Both methods show substantial improvements over base performance across all domains, confirming that \MethodName\ develops transferable reasoning capabilities.

These results demonstrate that debate-generated consensus signals provide an effective unsupervised alternative to ground truth supervision, achieving comparable performance without human annotation. This approach offers significant advantages for scaling reasoning improvements to new domains or large datasets where expert labels are unavailable or prohibitively expensive.

\begin{table*}[t]
\caption{Within-distribution performance comparison: Debate-derived majority vote supervision vs. ground truth supervision on training domains. Models are trained on MATH, GSM8K, and MathQA datasets. Bold indicates the better method for each model-dataset pair, demonstrating that debate-derived supervision achieves comparable performance to ground truth labels.}
\centering
\footnotesize
\begin{tabular}{ll|cc|cc|cc}
\toprule
\multirow{2}{*}{Model} & \multirow{2}{*}{Method} & \multicolumn{2}{c|}{MATH} & \multicolumn{2}{c|}{GSM8K} & \multicolumn{2}{c}{MathQA} \\
\cmidrule(lr){3-4} \cmidrule(lr){5-6} \cmidrule(l){7-8}
& & Debate-MV & GT & Debate-MV & GT & Debate-MV & GT \\
\midrule
\multirow{5}{*}{Qwen2B}
  & Base & 10.4 & 10.4 & 27.0 & 27.0 & 7.4 & 7.4 \\
  & SFT  & \bfseries 10.8 & 10.4 & 25.6 & \bfseries 26.4 & 8.2 & \bfseries 8.8 \\
  & GRPO & 19.4 & \bfseries 21.0 & 45.2 & \bfseries 48.6 & 18.6 & \bfseries 19.6 \\
  & KTO  & 22.6 & \bfseries 23.2 & \bfseries 54.8 & 54.6 & \bfseries 28.8 & 28.6 \\
  & DPO  & \bfseries 24.8 & 24.2 & 51.4 & \bfseries 52.0 & \bfseries 24.2 & 24.0 \\
\addlinespace
\multirow{5}{*}{Llama3B}
  & Base & 32.0 & 32.0 & 69.6 & 69.6 & 24.6 & 24.6 \\
  & SFT  & \bfseries 33.2 & 32.4 & 64.2 & 64.2 & \bfseries 26.4 & 25.2 \\
  & GRPO & 45.8 & \bfseries 46.4 & \bfseries 75.8 & 74.8 & \bfseries 36.2 & 31.8 \\
  & KTO  & \bfseries 48.0 & 47.8 & 76.0 & \bfseries 76.8 & \bfseries 41.4 & 40.6 \\
  & DPO  & 53.2 & \bfseries 53.6 & \bfseries 80.4 & 77.8 & \bfseries 46.2 & 45.4 \\
\bottomrule
\end{tabular}
\label{tab:within_distribution_comparison}
\end{table*}

\begin{table*}[t]
\caption{Cross-domain generalization: Debate-derived majority vote supervision vs. ground truth supervision on unseen reasoning tasks. Models trained on mathematical datasets (MATH, GSM8K, MathQA) generalize effectively to diverse reasoning domains. Bold indicates the better method for each model-dataset pair, showing that both supervision approaches transfer well to out-of-distribution tasks.}
\centering
\footnotesize
\begin{tabular}{ll|cc|cc|cc}
\toprule
\multirow{2}{*}{Model} & \multirow{2}{*}{Method} & \multicolumn{2}{c|}{SVAMP} & \multicolumn{2}{c|}{GPQA} & \multicolumn{2}{c}{CSQA} \\
\cmidrule(lr){3-4} \cmidrule(lr){5-6} \cmidrule(l){7-8}
& & Debate-MV & GT & Debate-MV & GT & Debate-MV & GT \\
\midrule
\multirow{5}{*}{Qwen2B}
  & Base & 48.30 & 48.30 & 0.45 & 0.45 & 3.80 & 3.80 \\
  & SFT  & \bfseries 53.30 & 53.00 & \bfseries 17.90 & 0.89 & 16.80 & \bfseries 18.80 \\
  & GRPO & \bfseries 60.30 & 58.33 & \bfseries 8.70 & 7.81 & \bfseries 10.80 & 9.60 \\
  & KTO  & 76.00 & \bfseries 76.33 & 16.70 & \bfseries 16.96 & \bfseries 63.40 & 60.80 \\
  & DPO  & \bfseries 65.00 & 64.67 & 19.64 & \bfseries 20.98 & \bfseries 62.2 & 60.80 \\
\addlinespace
\multirow{5}{*}{Llama3B}
  & Base & 71.30 & 71.30 & 0.67 & 0.67 & 53.00 & 53.00 \\
  & SFT  & 68.33 & \bfseries 72.0 & \bfseries 2.23 & 2.23 & 57.40 & \bfseries 57.80 \\
  & GRPO & 75.00 & \bfseries 79.33 & \bfseries 6.92 & 5.13 & \bfseries 63.20 & 59.40 \\
  & KTO  & 76.70 & \bfseries 78.67 & 8.93 & \bfseries 9.82 & \bfseries 62.20 & 61.80 \\
  & DPO  & 78.40 & \bfseries 80.67 & \bfseries 11.40 & 11.60 & \bfseries 64.00 & 62.40 \\
\bottomrule
\end{tabular}
\label{tab:cross_domain_generalization}
\end{table*}

\section{Post-Training Method Impact on Log-Probability Distribution}
\label{app:logp}

\begin{figure}
    \centering
    \includegraphics[trim={0pt 0pt 0pt 46pt},clip,width=1.0\linewidth]{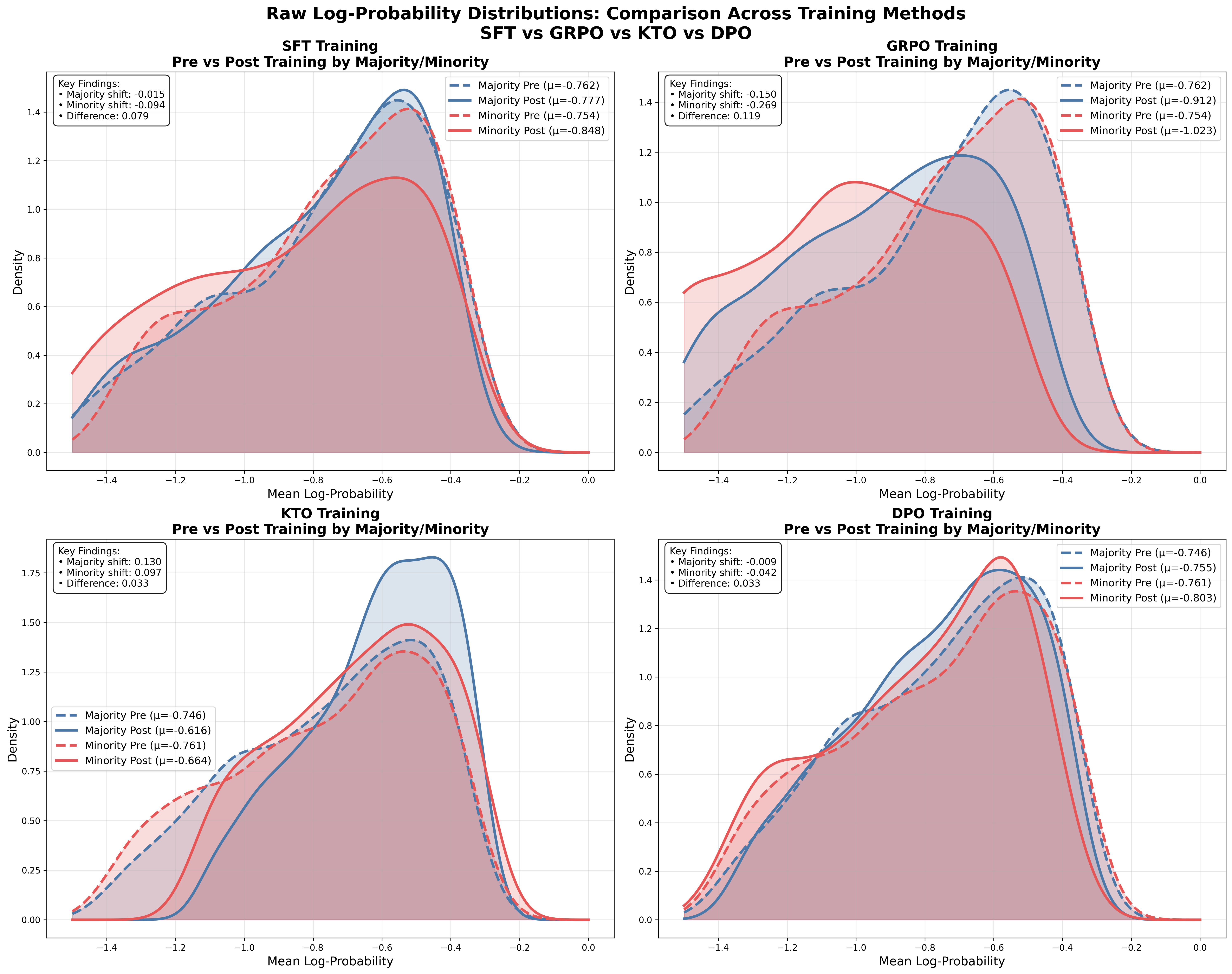}
    \caption{Log probability distributions for majority and minority answers before and after post-training (Qwen2B, Math).}
    \label{fig:pma}
\end{figure}

Figure \ref{fig:pma} shows how different post-training methods affected log-probability distributions for Qwen-2B on MATH. Preference-based methods (DPO, KTO) increased the density of majority distributions, with KTO showing particularly pronounced effects: higher peaks and tighter tails indicating more concentrated probability mass around consensus responses. SFT left the majority distribution largely unchanged while substantially reducing minority distribution density. GRPO created the most dramatic separation between distributions, shifting both leftward (lower log-probabilities overall) but with the minority distribution shifting much more substantially than the majority. While these different patterns (probability concentration for preference methods versus selective penalization for GRPO) all corresponded with performance improvements, further research is needed to better understand the relationship between these specific distributional changes and the impact on consistency.

\section{Preference Learning as an Implicit Format Reward}
\label{app:formatreward}

This section investigates how preference learning through consensus signals acts as an implicit format reward, teaching models to produce more efficient and accurate reasoning without explicit formatting supervision. We examine three key aspects: the baseline capability requirements for effective consensus formation, how token constraints interact with reasoning improvements, and the decomposition of performance gains into format improvements versus problem-solving accuracy improvements. Crucially, we demonstrate that while preference learning does function effectively as an implicit format reward, most of the performance gains observed are attributed to fundamental improvements in problem-solving accuracy rather than mere formatting compliance.

\subsection{Baseline Capability Requirements for Consensus Formation}

Our experiments reveal that \MethodName\ requires sufficient baseline model capability to generate meaningful consensus signals. When models lack foundational problem-solving ability, they fail to produce the correct responses necessary for consensus-based reinforcement. Table~\ref{tab:maxtoken} illustrates this limitation: under a strict 256-token limit, Qwen2B produces no correct responses on AMC, resulting in no useful consensus to reinforce. Increasing the token limit to 512 partially mitigates this by allowing more reasoning space, though inefficient reasoning patterns can still cause truncation.

\begin{table}[h]
\caption{Impact of token limits and baseline capability on AMC performance. Shows accuracy percentages for base models and post-trained models (``All") under different token constraints. When models lack sufficient baseline capability (Qwen2B at 256 tokens), consensus formation fails as no correct responses are generated for reinforcement.}
\centering
\begin{tabular}{lcccc}
\toprule
 & \multicolumn{2}{c}{max\_new\_tokens = 256} & \multicolumn{2}{c}{max\_new\_tokens = 512} \\
\cmidrule(lr){2-3} \cmidrule(lr){4-5}
 & Base & All & Base & All \\
\midrule
Qwen2B   & 0.0   & 0.0    & 5.0   & \bfseries 12.5 \\
Llama3B  & 7.5 & \bfseries 10.0   & 10.0  & \bfseries 20.0   \\
\bottomrule
\end{tabular}
\label{tab:maxtoken}
\end{table}
This baseline capability requirement has important implications for applying \MethodName: models must possess some initial problem-solving ability on the target domain to benefit from consensus-based training. However, once this threshold is met, we observe that improvements on easier datasets can generalize to more challenging tasks, suggesting that enhanced self-consistency helps overcome some limitations of consensus formation on difficult problems.

\subsection{Token Efficiency and Reasoning Quality Improvements}

Post-training simultaneously addresses token efficiency and reasoning quality. Models trained with debate under constrained token limits generate more parseable answers within the budget while achieving higher accuracy on those answers. To verify that improvements extend beyond mere format optimization, we evaluated models post-trained with 256-token debates using 512-token test budgets.

\begin{table}[t]
\centering
\caption{Improvements from post-training with $\textrm{max\_new\_tokens}=256$ translate when tested with larger token counts (512). ``All" represents a model post-trained on Math, MathQA, and GSM8K; its row reports improvement deltas over the base model performance.}
\label{tab:pt_generalize_512}
\setlength{\tabcolsep}{6pt}
\begin{tabular}{ll|rrrrrr}
\toprule
\multirow{2}{*}{Model} & \multirow{2}{*}{Trained on} & \multicolumn{6}{c}{Tested on ($\textrm{max\_new\_tokens}=512$)} \\
\cmidrule(l){3-8}
 &  & MATH & GSM8K & MathQA & SVAMP & AMC & GPQA \\
\midrule
\multirow{2}{*}{Qwen2B}
  & Base & 32.8 & 47.4 & 18.8 & 53.7 & 7.5 & 12.5 \\
  & All  & $\DeltaUpGreen{15.6}$ & $\DeltaUpGreen{24.6}$ & $\DeltaUpGreen{22.2}$ & $\DeltaUpGreen{20.0}$ & $\DeltaUpGreen{2.5}$ & $\DeltaUpGreen{11.2}$ \\
\midrule
\multirow{2}{*}{Llama3B}
  & Base & 60.8 & 76.8 & 43.0 & 78.0 & 10.0 & 9.2 \\
  & All  & $\DeltaUpGreen{6.4}$ & $\DeltaDownRed{0.2}$ & $\DeltaUpGreen{11.2}$ & $\DeltaUpGreen{3.0}$ & $\DeltaUpGreen{10.0}$ & $\DeltaUpGreen{14.5}$ \\
\bottomrule
\end{tabular}
\end{table}

Table~\ref{tab:pt_generalize_512} demonstrates that performance gains persist under increased token limits, indicating genuine reasoning enhancement rather than just format improvements. We also analyze the sequence lengths of the trajectories generated for the analysis in App.~\ref{app:correlation}, showing that uncapped post-trained models generate responses that are 22-36\% shorter than their base models. This evidence suggests that self-guided preference learning functions as an implicit format reward, teaching models to produce more concise and effective reasoning patterns that generalize across computational budgets and task difficulties.

\subsection{Decomposing the Sources of Improvement}

To understand whether our gains stem from improved reasoning or merely better formatting, we decompose performance improvements into their constituent components. Our analysis separates gains from better completion (avoiding truncation) versus fundamental reasoning improvements.

\begin{table}[t]
\centering
\caption{Decomposition of performance improvements into completion gains versus reasoning gains. ``From Completion" represents accuracy gained if all truncated base model responses had been allowed to complete. ``From Better Reasoning" represents the remaining improvement attributable to fundamental problem-solving enhancement. Percentages show the relative contribution of each component to total gains.}
\label{tab:improvement_source}

\begin{tabular}{ll|cc|ccc}
\toprule
Model & Dataset & Before & After & Total Gain & From Completion & From Better Reasoning \\
\midrule
Qwen2B & CSQA & 3.8 & 63.4 & $\DeltaUpGreen{59.6}$ & $\DeltaUpGreen{18.6}$ (31\%) & $\DeltaUpGreen{41.0}$ (69\%) \\
 & GPQA & 0.4 & 16.7 & $\DeltaUpGreen{16.3}$ & $\DeltaUpGreen{0.7}$ (4\%) & $\DeltaUpGreen{15.6}$ (96\%) \\
 & GSM8K & 24.6 & 54.8 & $\DeltaUpGreen{30.2}$ & $\DeltaUpGreen{0.0}$ (0\%) & $\DeltaUpGreen{30.2}$ (100\%) \\
 & MATH & 10.4 & 22.6 & $\DeltaUpGreen{12.2}$ & $\DeltaUpGreen{0.0}$ (0\%) & $\DeltaUpGreen{12.2}$ (100\%) \\
 & MathQA & 7.4 & 28.8 & $\DeltaUpGreen{21.4}$ & $\DeltaUpGreen{2.0}$ (9\%) & $\DeltaUpGreen{19.4}$ (91\%) \\
 & SVAMP & 48.3 & 76.0 & $\DeltaUpGreen{27.7}$ & $\DeltaUpGreen{0.4}$ (1\%) & $\DeltaUpGreen{27.3}$ (99\%) \\
\midrule
Llama3B & CSQA & 53.0 & 64.0 & $\DeltaUpGreen{11.0}$ & $\DeltaUpGreen{1.2}$ (11\%) & $\DeltaUpGreen{9.8}$ (89\%) \\
 & GPQA & 0.7 & 11.4 & $\DeltaUpGreen{10.7}$ & $\DeltaUpGreen{0.9}$ (8\%) & $\DeltaUpGreen{9.8}$ (92\%) \\
 & GSM8K & 69.6 & 80.4 & $\DeltaUpGreen{10.8}$ & $\DeltaUpGreen{0.2}$ (2\%) & $\DeltaUpGreen{10.6}$ (98\%) \\
 & MATH & 41.2 & 53.2 & $\DeltaUpGreen{12.0}$ & $\DeltaUpGreen{0.2}$ (2\%) & $\DeltaUpGreen{11.8}$ (98\%) \\
 & MathQA & 29.2 & 46.2 & $\DeltaUpGreen{17.0}$ & $\DeltaDownRed{0.2}$ (-1\%) & $\DeltaUpGreen{17.2}$ (101\%) \\
 & SVAMP & 71.3 & 75.7 & $\DeltaUpGreen{4.3}$ & $\DeltaUpGreen{0.0}$ (0\%) & $\DeltaUpGreen{4.3}$ (100\%) \\
\bottomrule
\end{tabular}
\end{table}

The results reveal that 69-100\% of improvements stem from better reasoning rather than just avoiding truncation. While base models exhibit high truncation rates (e.g., 74.8\% for Qwen2B on CSQA, 82.8\% on MATH), post-training reduces these dramatically. However, even after accounting for completion improvements, substantial reasoning gains remain, confirming that our method teaches fundamentally better problem-solving, not merely more efficient token usage.

\subsection{Formatting Improvements and Remaining Reasoning Gains}
To further isolate reasoning improvements from formatting effects, we analyze how post-training affects answer formatting compliance. Table~\ref{tab:formatting_fixes} shows the percentage of responses with correct reasoning but incorrect formatting, and calculates reasoning gains that persist beyond all formatting improvements.
\begin{table}[t]
\centering
\caption{Analysis of formatting improvements versus reasoning gains. ``Accuracy Lost to Format Errors" shows the percentage of responses with correct reasoning but incorrect formatting (e.g., writing ``The answer is A" instead of \texttt{\textbackslash boxed\{A\}}). ``Reasoning Gain Beyond Formatting" shows improvements that persist even after accounting for all formatting fixes, calculated as: Total Gain - (Format Loss Before - Format Loss After).}
\label{tab:formatting_fixes}

\begin{tabular}{ll|cc|c}
\toprule
Model & Dataset & \multicolumn{2}{c|}{Accuracy Lost to Format Errors (\%)} & Reasoning Gain \\
& & Before Training & After Training & Beyond Formatting (\%) \\
\midrule
Qwen2B & CSQA & 38.8 & 0.6 & $\DeltaUpGreen{21.4}$ \\
 & GPQA & 6.2 & 1.6 & $\DeltaUpGreen{11.7}$ \\
 & MathQA & 5.8 & 0.6 & $\DeltaUpGreen{16.2}$ \\
\midrule
Llama3B & CSQA & 7.0 & 1.4 & $\DeltaUpGreen{5.4}$ \\
 & GPQA & 7.6 & 3.8 & $\DeltaUpGreen{6.9}$ \\
 & MathQA & 5.8 & 2.8 & $\DeltaUpGreen{14.0}$ \\
\bottomrule
\end{tabular}
\end{table}
Post-training dramatically reduces formatting losses (e.g., from 38.8\% to 0.6\% on CSQA for Qwen2B). However, the ``Reasoning Gain Beyond Formatting" column reveals that substantial improvements remain even after perfect formatting is assumed. For instance, Qwen2B's 59.6 percentage point improvement on CSQA includes 38.2 points from better formatting; the remaining 21.4 points represent accuracy improvements distinct from formatting.

\subsection{Parser Implementation Details}
To quantify formatting improvements versus reasoning gains, we employed two parsers with different strictness levels. Our standard parser requires answers in \texttt{\textbackslash boxed\{\}} format and returns None for any deviation. The relaxed parser, used for impact analysis in Table~\ref{tab:formatting_fixes}, accepts common natural language patterns like ``The answer is A" or ``Answer: 42" by searching the entire response for valid answer formats. 

This relaxed parser first attempts the strict extraction, then falls back to regex patterns that capture answers expressed naturally in text. For multiple choice, it accepts patterns like ``[Tt]he answer is ([A-E])" or ``Answer: ([A-E])". For numerical answers, it extracts from patterns like ``= 42" at line endings or ``The final answer is 42". When multiple patterns match, it takes the last occurrence, mimicking how humans identify the final answer in a reasoning chain.

The gap between strict and relaxed parser accuracies precisely measures the ``Accuracy Lost to Format Errors" in Table~\ref{tab:formatting_fixes}. Strict formatting compliance is a key measure of a model's ability to follow instructions and is critical for downstream applications that rely on reliable parsing of LM outputs, while the relaxed parser accounts for human-interpretable correctness. This dual evaluation reveals that preference learning through consensus not only teaches proper formatting as an implicit reward but fundamentally improves problem-solving capabilities, with reasoning gains persisting even when formatting constraints are removed.

\section{Training Curves}
\label{app:training_curves}

\begin{figure}[t]
  \centering
  \begin{subfigure}{1.0\linewidth}
    \centering
    \includegraphics[width=\linewidth,height=0.43\textheight]{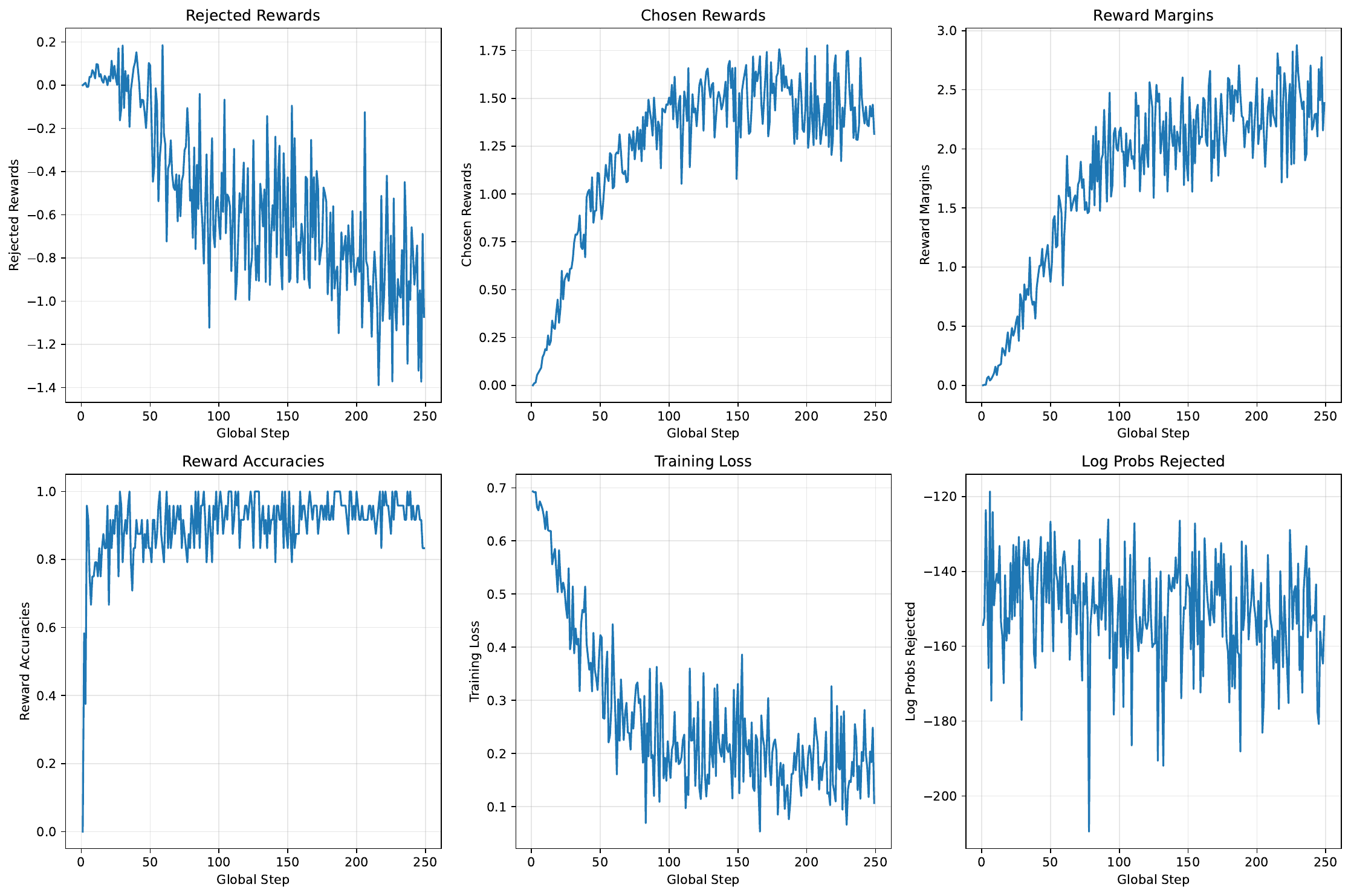}
    \caption{MV-DPO training curves}
    \label{fig:dpo_training_curves}
  \end{subfigure}
  \begin{subfigure}{1.0\linewidth}
    \centering
    \includegraphics[width=\linewidth,height=0.43\textheight]{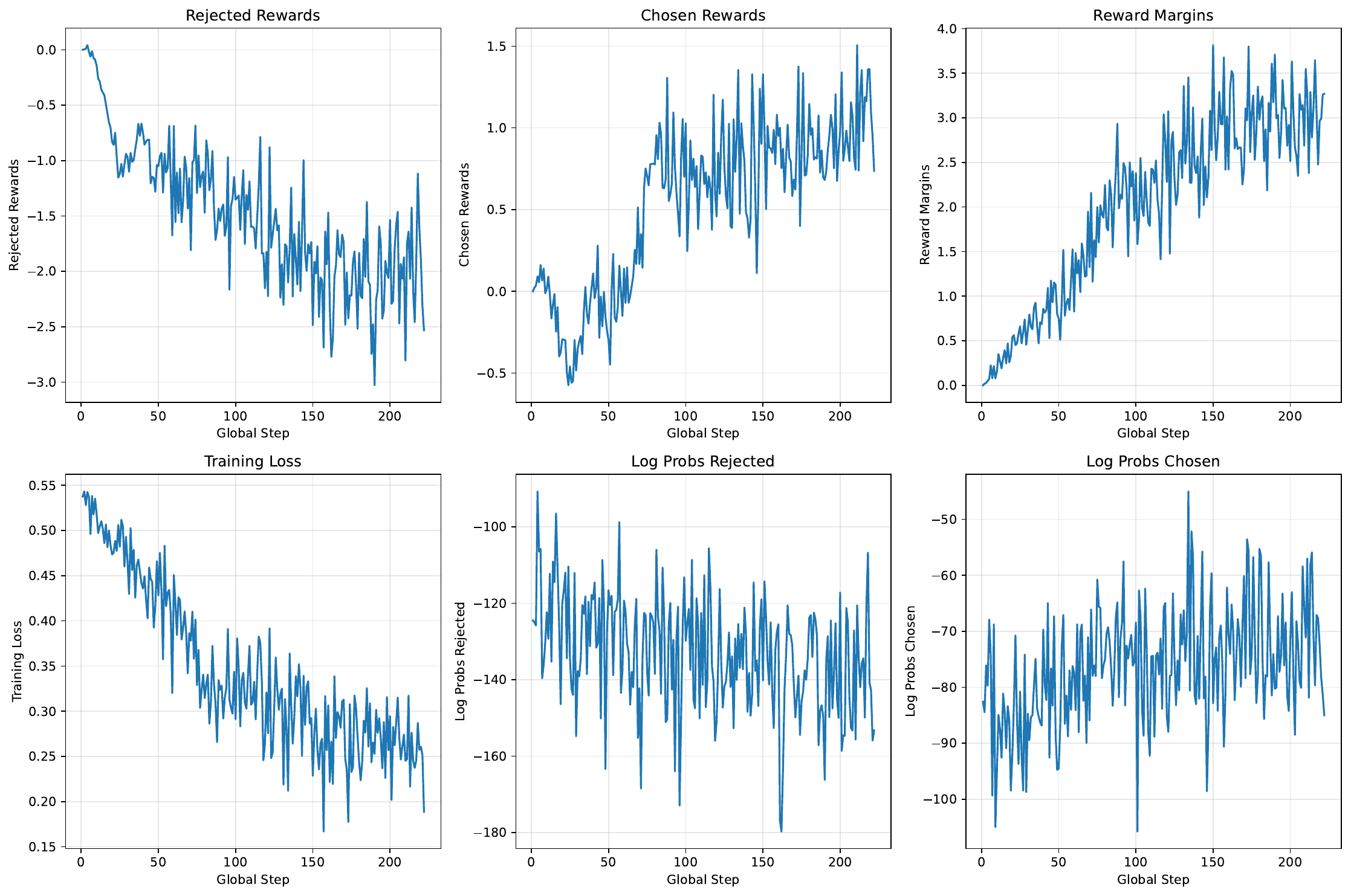}
    \caption{MV-KTO training curves}
    \label{fig:kto_training_curves}
  \end{subfigure}
  
    \caption{Training curves for MV-DPO and MV-KTO. (a) MV-DPO: increasing reward margin between majority (chosen) and minority (rejected) responses, with declining rejected log probabilities. (b) MV-KTO: similar reward margin increase using unpaired examples, with rejected log probabilities decreasing and chosen increasing.}
  \label{fig:dpo_kto_training}
\end{figure}

\begin{figure}[t]
  \centering
  \begin{subfigure}{1.0\linewidth}
    \centering
    \includegraphics[width=\linewidth,height=0.43\textheight,keepaspectratio]{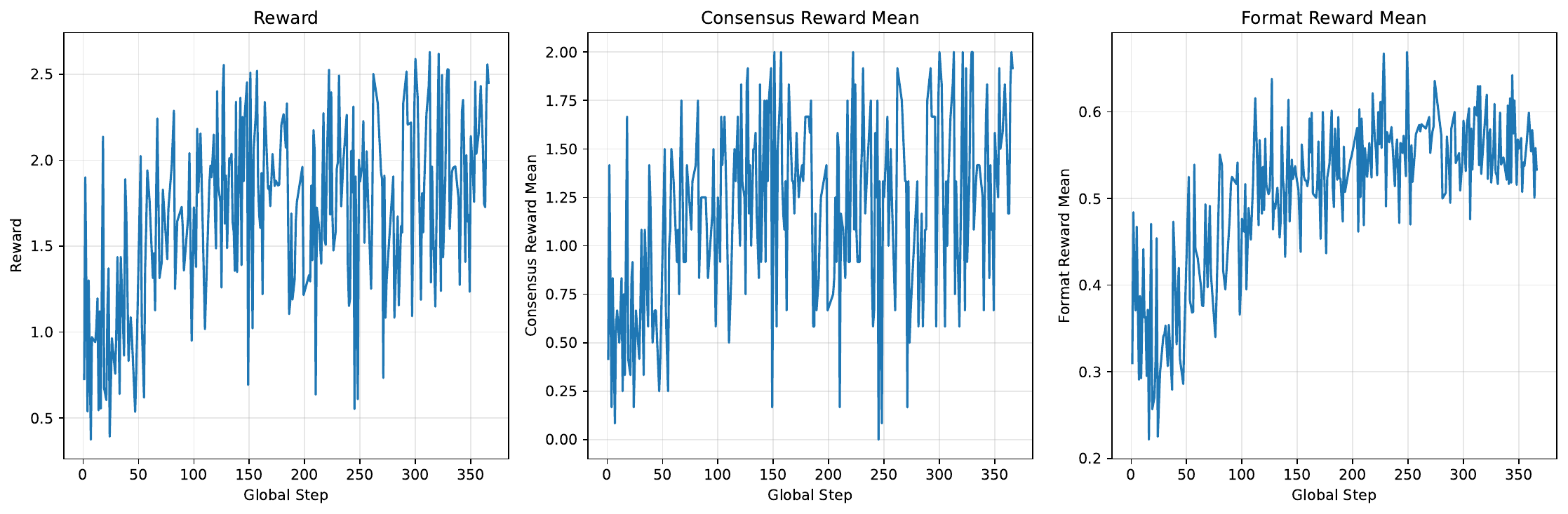}
    \caption{MV-GRPO training curves}
    \label{fig:grpo_training_curves}
  \end{subfigure}
  \begin{subfigure}{0.8\linewidth}
    \centering
    \includegraphics[width=\linewidth,height=0.43\textheight,keepaspectratio]{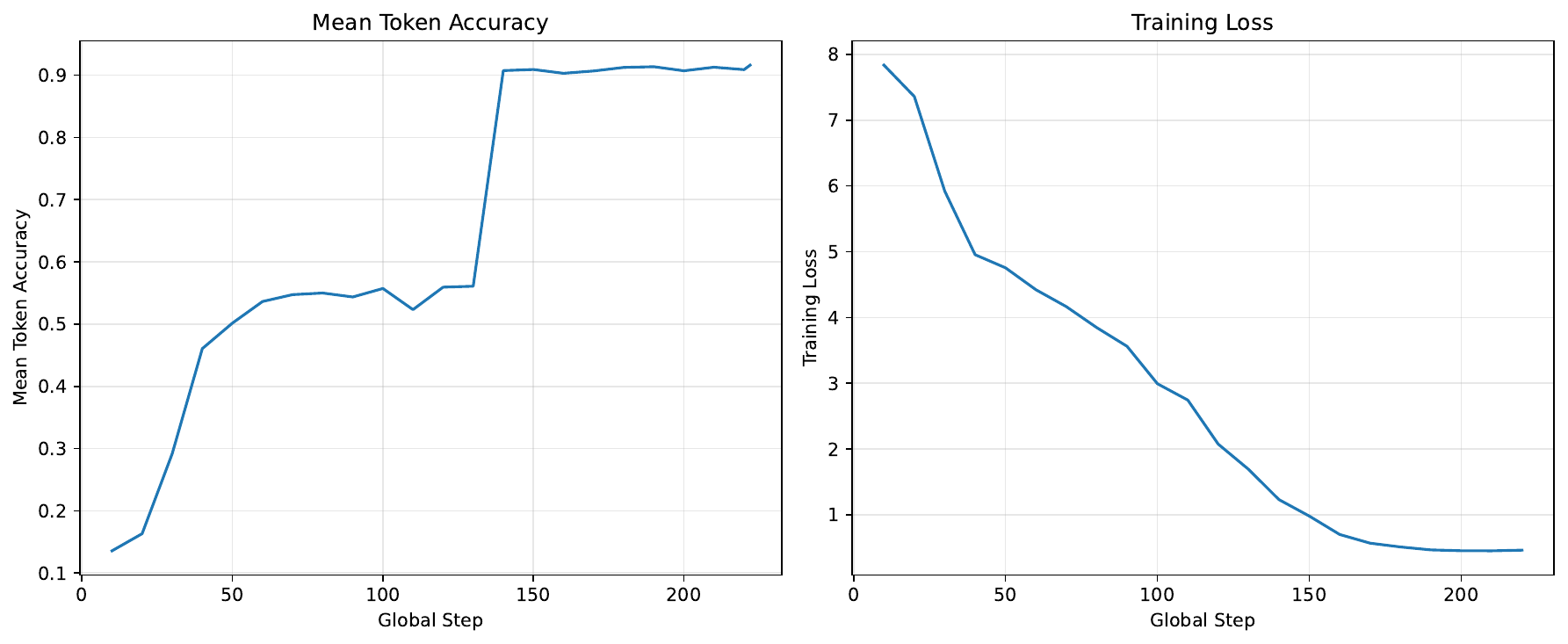}
    \caption{MV-SFT training curves}
    \label{fig:sft_training_curves}
  \end{subfigure}

    \caption{Training curves for MV-GRPO and MV-SFT. (a) MV-GRPO: consensus and format rewards both increase. (b) MV-SFT: token accuracy increases while loss decreases.}
  \label{fig:grpo_sft_training}
\end{figure}

The training dynamics of our post-training methods provide insight into how models learn to refine responses towards multi-agent consensus preferences. Figures~\ref{fig:dpo_training_curves}–\ref{fig:sft_training_curves} show example training curves across key metrics during post-training with MV-DPO, MV-KTO, MV-GRPO, and MV-SFT.

Across all methods, the reward margins between chosen (consensus) and rejected (non-consensus) responses increase consistently, indicating effective preference learning. MV-DPO and MV-KTO reveal this pattern strongly: margins start near zero and grow steadily as the models optimize towards favoring majority-preferred outputs. MV-KTO achieves similar reward improvements despite not requiring strict paired comparisons, demonstrating its robust learning dynamics.

Training accuracy converges for all methods, reaching high classification levels between consensus and non-consensus responses, showing that each approach successfully reinforces desired sampling behaviors. Correspondingly, losses decrease smoothly without signs of instability or collapse, indicating stable training processes.

Log probabilities of rejected responses decline across post-training methods, reflecting the models’ increasing tendency to assign lower likelihood to outputs outside the consensus. This is most pronounced in MV-DPO and MV-GRPO, where rejected rewards fall more steeply, delineating a clear separation between preferred and discouraged responses. MV-SFT, operating via imitation learning, exhibits strong improvements in token accuracy and loss while reinforcing consensus-aligned responses effectively.

Overall, the asymmetric reward trajectories, where chosen response rewards remain relatively stable or increase slightly while rejected response rewards decline sharply, suggest that post-training primarily discourages generation of minority or outlier outputs. This mechanism is a core driver behind the improved sampling consistency and reasoning quality observed in our experiments.

These training curves collectively support our hypothesis that post-training with majority vote preferences enables models to internalize collective agreement notions and reproduce responses better aligned with multi-agent consensus.

\section{Post-Training Self-Consistency Improvements Translate from 4-Bit Quantized Model to Full Model}
\label{app:boundscurves}

\begin{figure}[t]
  \centering
  \includegraphics[trim={0pt 0pt 0pt 25pt},clip,width=0.7\linewidth]{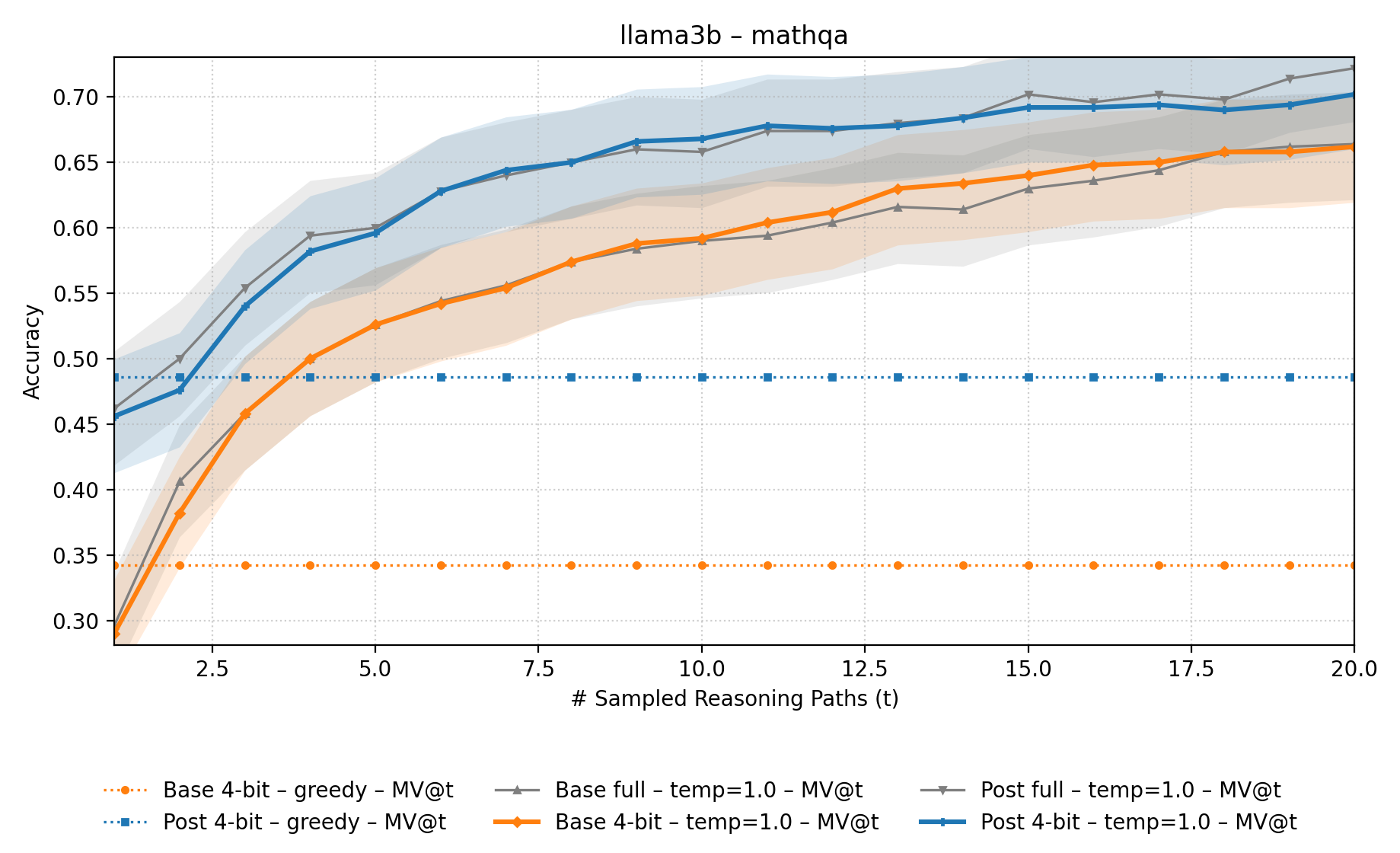}
  \caption{Llama-3B on MathQA (MV@t, t=20). \textcolor{blue}{Blue}: post-trained 4-bit model, \textcolor{orange}{Orange}: base 4-bit model, Grey: full-precision model.}
  \label{fig:Llama3B-MathQA-mv}
\end{figure}

\begin{figure}[t]
  \centering
  \includegraphics[trim={0pt 0pt 0pt 25pt},clip,width=0.7\linewidth]{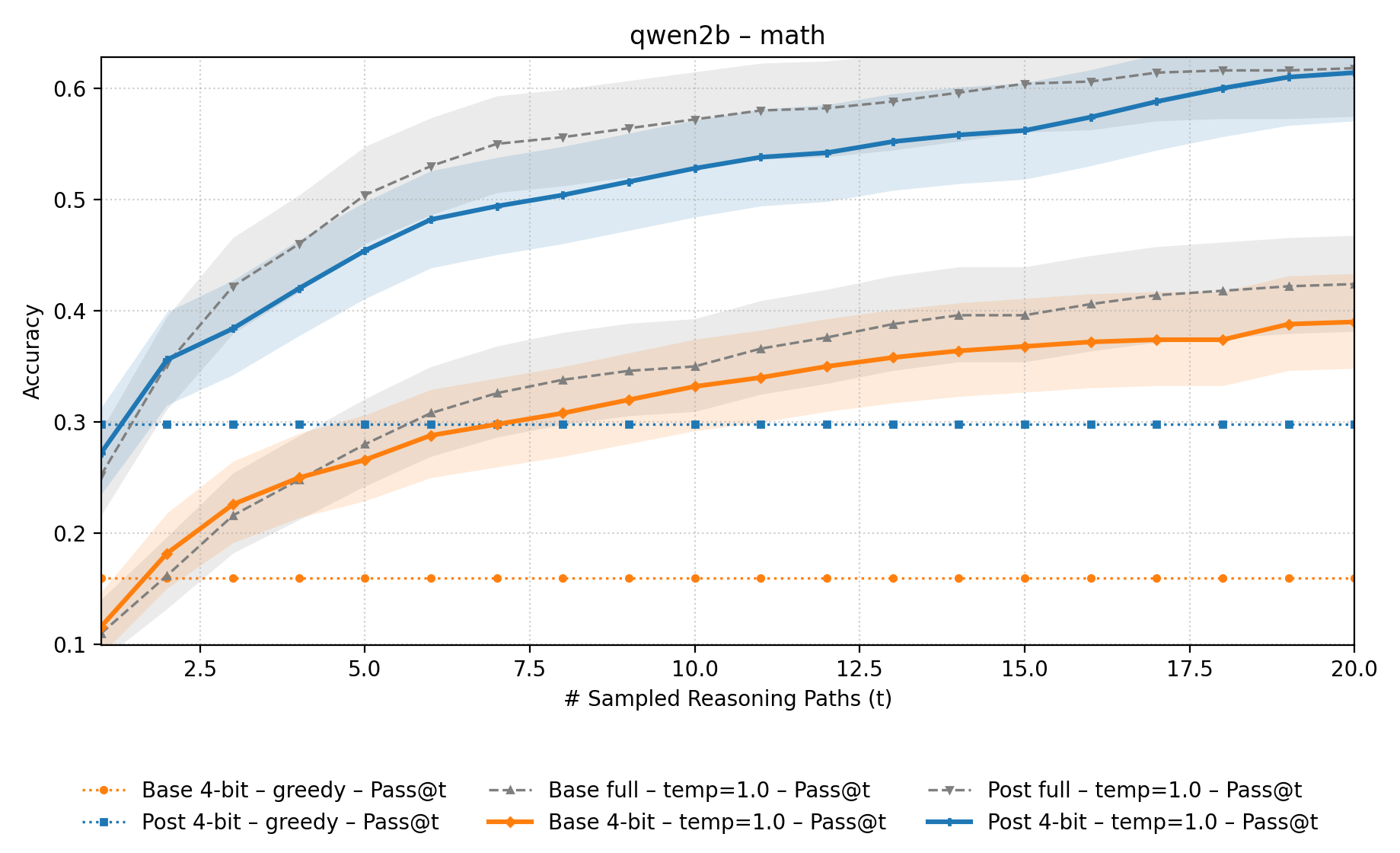}
  \caption{Qwen-2B on MATH (Pass@t, t=20). \textcolor{blue}{Blue}: post-trained 4-bit model, \textcolor{orange}{Orange}: base 4-bit model, Grey: full-precision model.}
  \label{fig:Qwen2B-math-pass}
\end{figure}

\begin{figure}[t]
  \centering
  \vspace{0.5em}
    \includegraphics[trim={0pt 25pt 0pt 310pt},clip,width=0.7\linewidth]{sampling_legend.pdf}
  \includegraphics[trim={0pt 50pt 0pt 25pt},clip,width=0.7\linewidth]{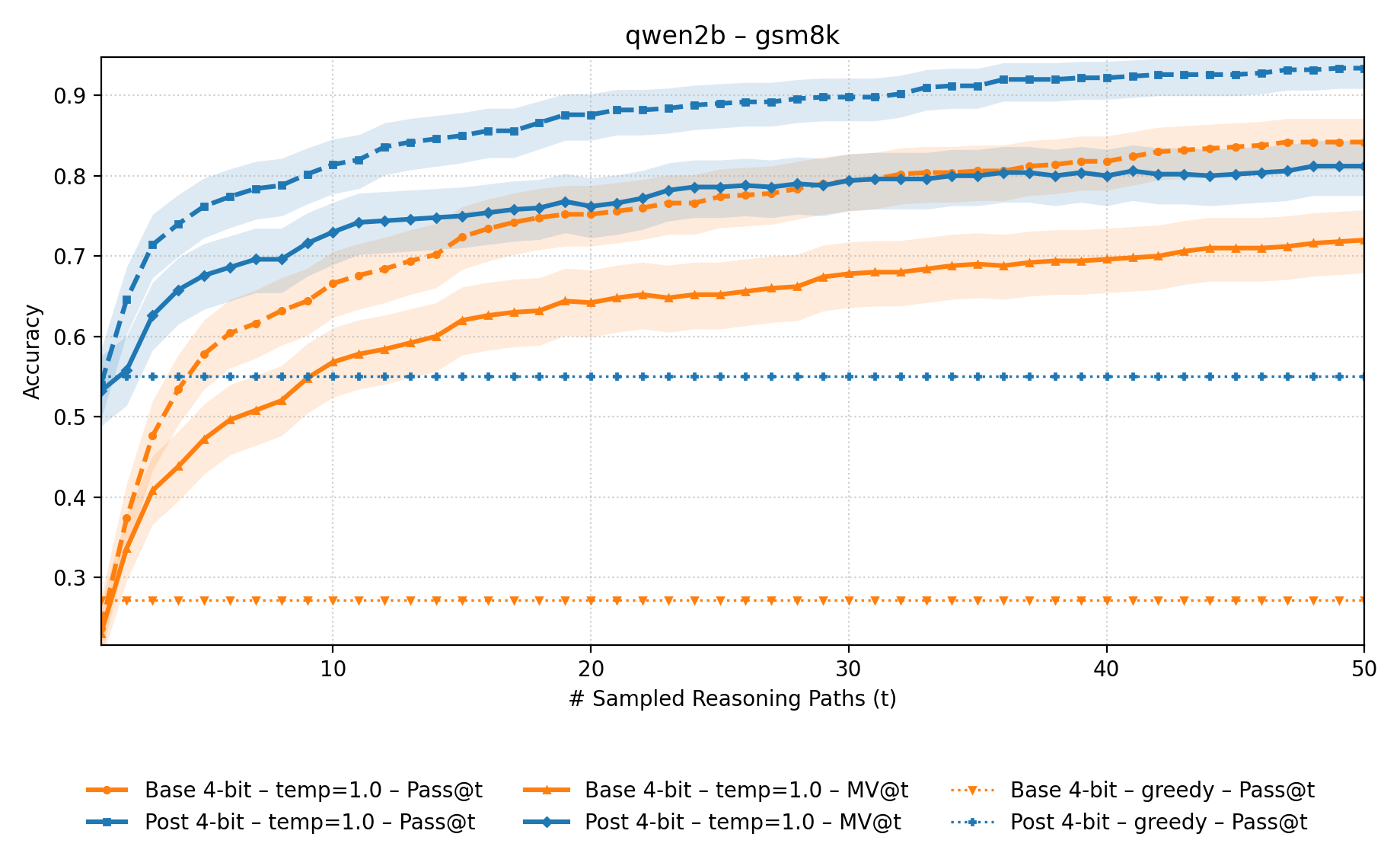}
  \caption{Qwen-2B on GSM8K (Pass@t, MV@t; t=50). \textcolor{blue}{Blue}: post-trained 4-bit model, \textcolor{orange}{Orange}: base 4-bit model.}
  \label{fig:Qwen2B-GSM8K-combined50}
\end{figure}

As shown in Figures \ref{fig:Llama3B-MathQA-mv}-\ref{fig:Qwen2B-GSM8K-combined50}, post-training improvements on 4-bit quantized models transfer to full-precision models, and the improvement margin persists with larger numbers of trajectories sampled ($t=50)$.

\section{Impact of Post-training on Debate Agreement Rates}
\label{app:agreement_rates}

\begin{figure}[h]
    \centering
    \includegraphics[width=1.0\linewidth]{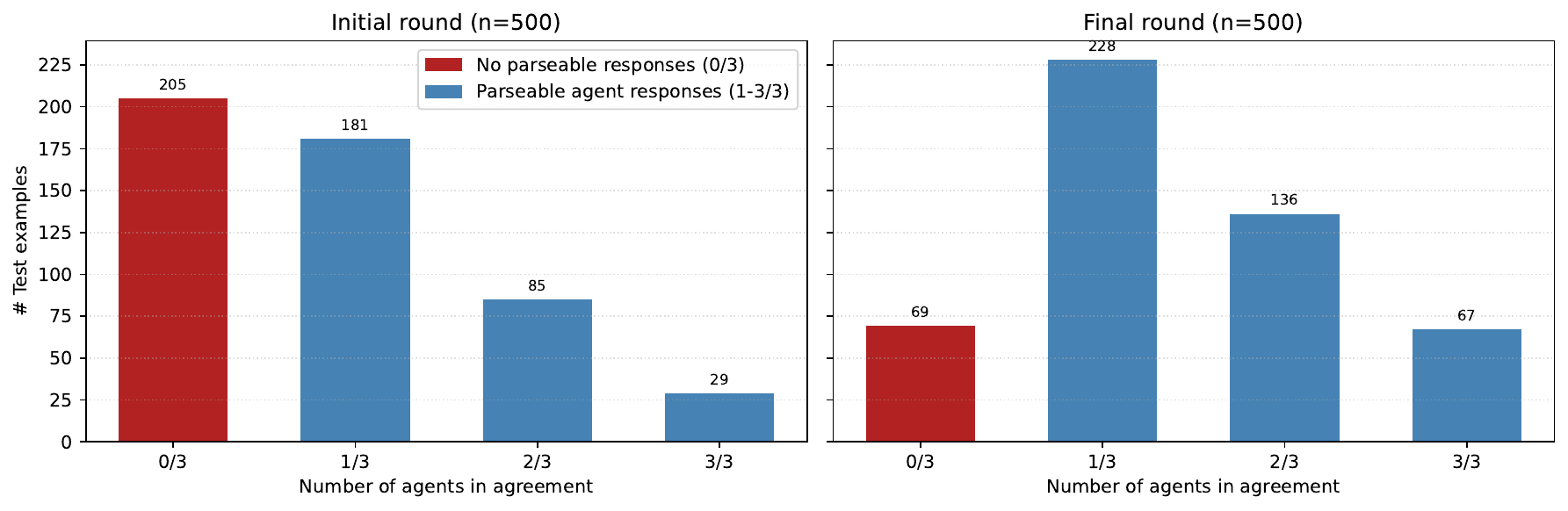}\\%
    \includegraphics[width=1.0\linewidth]{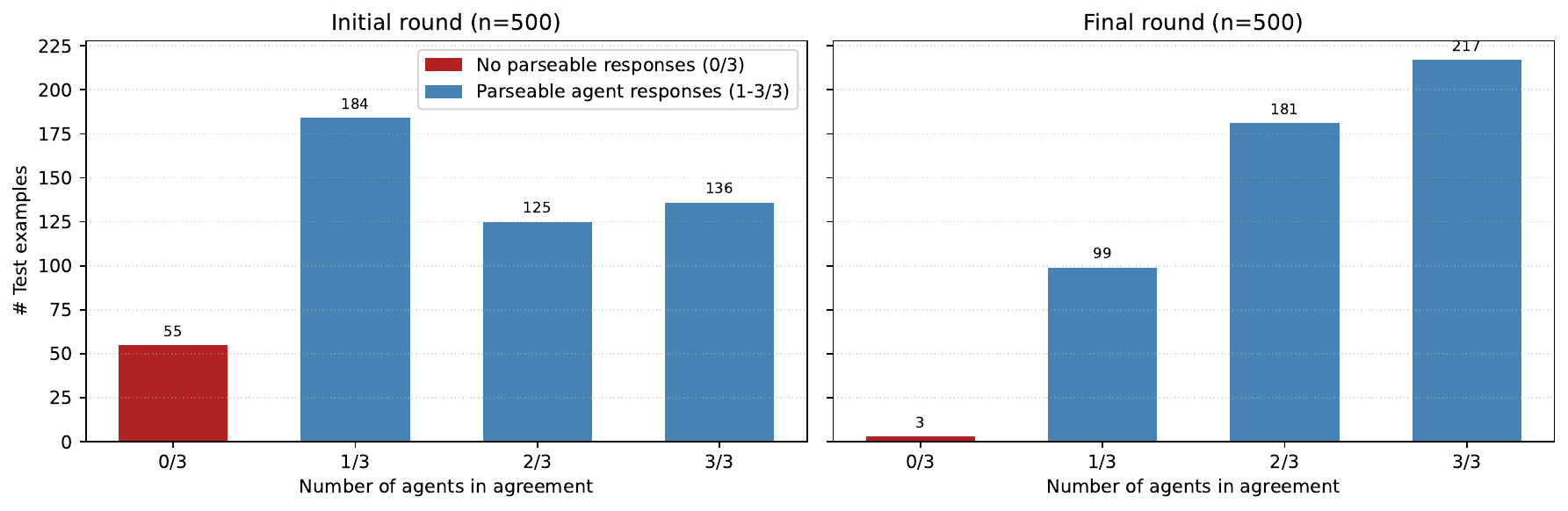}
    \caption{\MethodName\ drives meaningful improvements in both answer completeness and agent agreement, verifying the probability mass reallocation to the consensus set of reasoning trajectories (Ex: Qwen2B on GSM8K). Top: base model debate; bottom: post-trained model debate; left: initial debate round; right: final debate round.}
    \label{fig:agreement_rates}
\end{figure}

Figure~\ref{fig:agreement_rates} shows how agent agreement patterns evolve through debate, comparing base models (Iteration 0) against consensus-trained models (Iteration 1). These distributions directly measure whether our training successfully shifts probability mass toward consensus trajectories $\mathcal{G}_+(x)$.

Base models show relatively uniform agreement distributions, with only 13.4\% of examples achieving full consensus (3/3 agreement) in the final round. After consensus post-training with MV-DPO, this increases over three-fold to 43.4\%, with systematic improvements across all agreement levels: increased mass at 2/3 and 3/3 agreement, decreased mass at 1/3 agreement. This redistribution confirms that training drives the policy toward consensus-supporting trajectories.

Additionally, unparseable responses drop from 11\% to 0.6\% without explicit format rewards. Since our training favors consensus completions, which must finish within token limits to be comparable, the method implicitly rewards efficient, complete reasoning patterns. This suggests consensus alignment naturally encourages concise and coherent reasoning as a prerequisite for measurable agreement.

\section{Performance Comparison of RL on debate vs. single-round majority vote RL in single-agent setting}
\label{app:singleagentbenchmark}

Table ~\ref{tab:appendix_combined_results} reports the full results referenced in Table \ref{tab:debate_vs_mv_improvements}, including standard deviation over 3 seeds. We find that MACA with final round majority vote with debate context outperforms single-round majority-vote RL with DPO (akin to ScPO ~\citep{prasad2024self}) in all 8 configurations tested, and RL with GRPO (akin to TTRL ~\citep{Zuo2025TTRL}) in 6 out of 8 configurations tested, with the other 2 configs being within the standard deviation.

\begin{table}[t]
\caption{Single-agent: RL with debate outperforms RL with single-round majority vote. Benchmark against TTRL (single-round R0 MV-GRPO), ScPO (single-round R0 MV-DPO), MV-DPO (MACA: debate final round R1 MV-DPO with debate context after collaborative exchange). Accuracy averaged over 3 seeds with standard deviation reported in comparison to base(instruction-tuned) model.}
\centering
\small
\setlength{\tabcolsep}{6pt}
\begin{tabular}{lcccc}
\toprule
& Qwen-2B & Llama-3B & Phi-4B & Llama-8B \\
\midrule
\multicolumn{5}{c}{\textbf{MATH}} \\
\midrule
Base            & 9.4 & 40.2 & 39.8 & 36.0 \\
TTRL      & \textbf{27.4 $\pm$ 2.9} & 45.5 $\pm$ 5.7 & 45.9 $\pm$ 2.1 & 43.5 $\pm$ 0.2 \\
ScPO       & 11.7 $\pm$ 1.1 & 43.6 $\pm$ 0.2 & 39.9 $\pm$ 0.5 & 39.7 $\pm$ 0.6 \\
MV-DPO       & 26.1 $\pm$ 0.4 & \textbf{52.7 $\pm$ 0.7} & \textbf{46.7 $\pm$ 0.2} & \textbf{53.1 $\pm$ 0.8} \\
\midrule
\multicolumn{5}{c}{\textbf{GSM8K}} \\
\midrule
Base            & 26.4 & 72.8 & 76.4 & 67.8 \\
TTRL      & \textbf{50.9 $\pm$ 1.6} & 50.5 $\pm$ 8.3 & 81.2 $\pm$ 1.6 & 54.3 $\pm$ 5.7 \\
ScPO      & 42.3 $\pm$ 3.4 & 74.4 $\pm$ 0.8 & 82.3 $\pm$ 0.5 & 80.7 $\pm$ 1.2 \\
MV-DPO       & 50.2 $\pm$ 0.9 & \textbf{76.5 $\pm$ 0.3} & \textbf{83.7 $\pm$ 0.3} & \textbf{86.0 $\pm$ 0.2} \\
\bottomrule
\end{tabular}
\label{tab:appendix_combined_results}
\end{table}

\section{Efficiency}
\label{app:efficiency}

Here we detail the compute performance tradeoffs of MACA's MV-DPO with debate context against other single-agent benchmark methods. ScPO requires 0.075-0.58 GPU hours achieving -0.6\% to +13.6\% gains, TTRL requires 2.2-7.7 GPU hours achieving -26.4\% to +18\% changes, while MACA (MV-DPO with debate) requires 0.73-1.68 GPU hours achieving consistent positive gains. Our efficiency plots demonstrate that MV-DPO on debate provides the best performance-efficiency tradeoff—higher performance than ScPO/TTRL with comparable computational cost to ScPO and significantly lower cost than TTRL.

\begin{figure}
    \centering
    \includegraphics[width=1\linewidth]{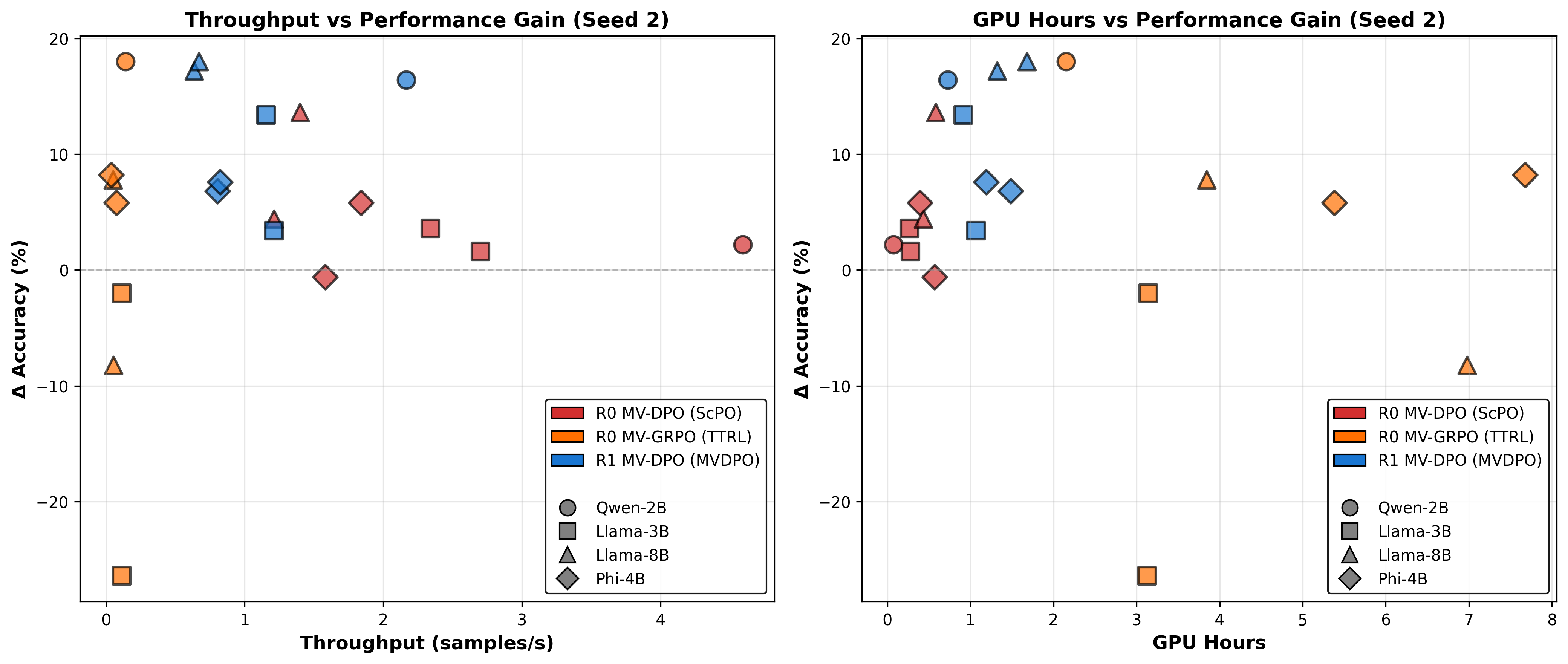}
    \caption{GPU throughput and training runtime compared with performance lift for single-agent post-training methods}
    \label{fig:efficiency}
\end{figure}

\end{document}